\documentclass[10pt,twocolumn,letterpaper]{article}
\usepackage[]{cvpr}

\definecolor{cvprblue}{rgb}{0.21,0.49,0.74}
\usepackage[pagebackref,breaklinks,colorlinks,allcolors=cvprblue]{hyperref}

\PassOptionsToPackage{table}{xcolor}
\usepackage{xcolor}
\usepackage{colortbl}

\usepackage{amsmath}
\usepackage{amssymb}
\usepackage{bm}
\usepackage{xspace,mathtools}

\usepackage{multirow}
\usepackage{multicol}
\usepackage{arydshln}  
\usepackage{booktabs}

\usepackage{graphicx}
\usepackage{epsfig}
\usepackage{wrapfig}

\usepackage{caption}
\usepackage{subcaption}

\usepackage{algorithm}
\usepackage{algpseudocode}

\usepackage{times}

\newcommand{\mjg}[1]{{{#1}}}

\definecolor{mlp}{gray}{0.95}
\definecolor{gs}{gray}{1.0}
\definecolor{row1}{HTML}{D8EFE7}
\definecolor{row2}{HTML}{EDF7F3}

\definecolor{cvprblue}{rgb}{0.21,0.49,0.74}
\newcommand{\cvprb}[1]{{\color{cvprblue}{#1}}}

\definecolor{l_gray}{gray}{0.95}
\definecolor{s_gray}{gray}{1.0}

\def\paperID{7866} 
\def\confName{CVPR}
\def\confYear{2026}

\title{GP-4DGS: Probabilistic 4D Gaussian Splatting from Monocular Video \\ via Variational Gaussian Processes}

\author{
Mijeong Kim\textsuperscript{\normalfont 1} 
\quad \quad \quad \quad
Jungtaek Kim\textsuperscript{\normalfont 3} 
\quad \quad \quad \quad
Bohyung Han\textsuperscript{\normalfont 1,2} 
\\
{\normalfont \textsuperscript{\normalfont 1}ECE and \textsuperscript{\normalfont 2}IPAI, Seoul National University, Korea \quad  \textsuperscript{\normalfont 3}University of Wisconsin--Madison, USA} \\
{\tt\small \{mijeong.kim, bhhan\}@snu.ac.kr}
\quad 
{\tt\small jungtaek.kim@wisc.edu} 
}

\begin{document}

\twocolumn[{
\maketitle
\centering
\includegraphics[width=0.96\linewidth]{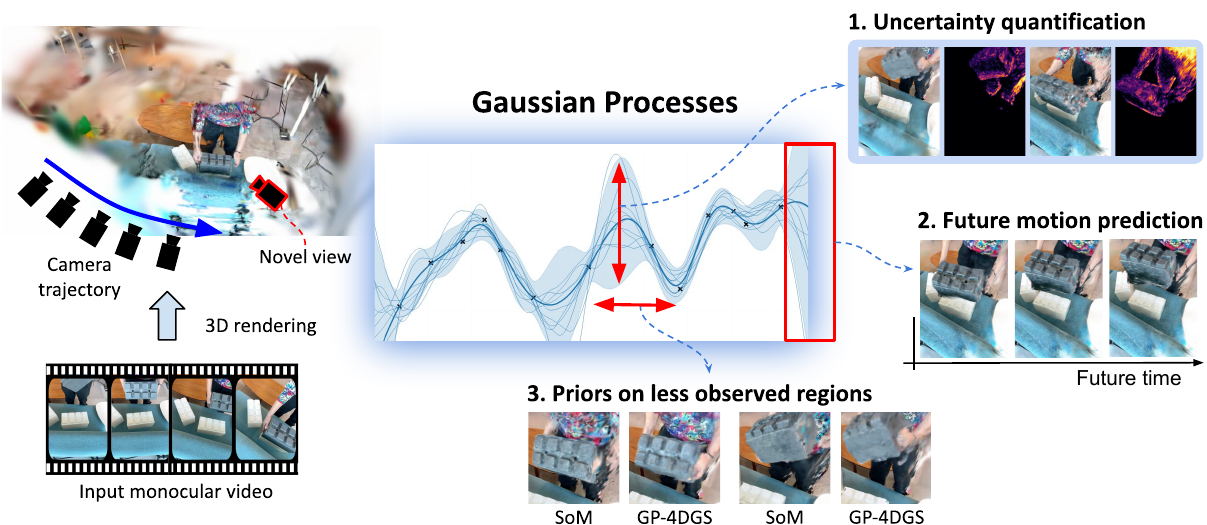}
\captionof{figure}
{We propose GP-4DGS, a novel integration of Gaussian Processes (GPs)~\cite{RasmussenCE2006book} into 4D Gaussian Splatting (4DGS).
Unlike existing deterministic approaches, this formulation enables robust uncertainty quantification, future motion prediction, and prior estimation for unobserved regions.}
\label{fig:teaser}
\vspace{5mm}
}]


\begin{abstract}
We present GP-4DGS, a novel framework that integrates Gaussian Processes (GPs) into 4D Gaussian Splatting (4DGS) for principled probabilistic modeling of dynamic scenes. 
While existing 4DGS methods focus on deterministic reconstruction, they are inherently limited in capturing motion ambiguity and lack mechanisms to assess prediction reliability. 
By leveraging the kernel-based probabilistic nature of GPs, our approach introduces three key capabilities: (i) uncertainty quantification for motion predictions, (ii) motion estimation for unobserved or sparsely sampled regions, and (iii) temporal extrapolation beyond observed training frames. 
To scale GPs to the large number of Gaussian primitives in 4DGS, we design spatio-temporal kernels that capture the correlation structure of deformation fields and adopt variational Gaussian Processes with inducing points for tractable inference. 
Our experiments show that GP-4DGS enhances reconstruction quality while providing reliable uncertainty estimates that effectively identify regions of high motion ambiguity. 
By addressing these challenges, our work takes a meaningful step toward bridging probabilistic modeling and neural graphics.

\end{abstract}
\section{Introduction}
\label{sec:intro}

Dynamic scene reconstruction from monocular videos has recently become an important problem due to its significant practical impact, enabling 3D capture in unconstrained environments for robotics~\cite{xin2023nerf,lu2023vl,yuan2023online}, autonomous systems~\cite{Min_2024_CVPR, Peng_2025_CVPR, Zhou_2024_CVPR, Yan_2025_CVPR}, and large reconstruction models~\cite{hong2024lrm,ZhangC2024geolrm}. 
Building on recent advances in static scene reconstruction~\cite{nerf,chen2022tensorf, garbin2021fastnerf, muller2022instant,3dgs,kim2022infonerf}, this field has rapidly evolved by extending static-scene base architectures with additional temporal modeling to handle scene dynamics.

The dominant paradigm in recent years is 4D Gaussian Splatting (4DGS)~\cite{wu2024deblur4dgs,rivero2024rig3dgs, wang2025gflow, kim20244d, wang2025shape, li2023spacetime, wu20234d}, which extends 3D Gaussian Splatting (3DGS) to dynamic scenes. 
These methods have achieved impressive visual quality in dynamic scene reconstruction task by representing scenes as collections of time-varying Gaussian primitives, optimized end-to-end through differentiable rendering.
However, these approaches treat motion as a deterministic optimization problem, imposing hand-crafted motion priors, \textit{e.g.}, polynomial deformations~\cite{li2023spacetime}, rigidity constraints~\cite{cai2024dynasurfgs}, or other fixed parametrizations~\cite{wang2025shape, kratimenos2024dynmf}, that are applied uniformly across all primitives without learning from data.
When primitives are poorly observed or occluded, these fixed priors become inappropriate or overly restrictive.
Moreover, existing methods lack principled mechanisms for uncertainty estimation and motion extrapolation beyond training frames.

In this work, we propose GP-4DGS, a novel integration of Gaussian Processes (GPs)~\cite{RasmussenCE2006book} into 4DGS that provides a principled probabilistic framework for dynamic scene reconstruction, as shown in Figure~\ref{fig:teaser}.
This simultaneously enables three capabilities absent in existing 4DGS algorithms: (i) uncertainty quantification for motion predictions, (ii) temporal extrapolation beyond observed frames, and (iii) learned motion priors that adapt to observation patterns.
Rather than imposing fixed motion constraints, we learn motion priors directly from well-observed primitives via kernel-based probabilistic modeling.
The key insight is to represent primitive deformations through GP posteriors, enabling the framework to automatically adjust regularization strength based on observation confidence and to naturally handle sparse or unobserved regions.
Notably, the first two capabilities emerge directly from the probabilistic formulation of GPs, without any additional modeling overhead.

This integration, however, is non-trivial and requires addressing fundamental modeling and computational challenges.
First, standard GP kernels assume isotropic correlations, fundamentally mismatched to spatio-temporal data where spatial dimensions $(x, y, z)$ and time $t$ have drastically different correlation structures.
To address this, we introduce a composite kernel design that explicitly separates spatial Mat\'ern kernels for geometric smoothness from periodic temporal kernels for motion periodicity.
Second, exact GP posterior computation has $\mathcal{O}(N^3)$ time complexity where $N$ is the number of Gaussian primitives, which is prohibitive as $N$ typically reaches tens of thousands.
We develop a scalable formulation combining variational inference with inducing points, reducing complexity to $\mathcal{O}(NM^2 + M^3)$, where $M$ is the number of inducing points with $M \ll N$.
Third, integrating probabilistic priors with vision-based optimization requires careful design. 
We propose a novel GP-GS optimization algorithm that forms synergy between the two by alternating between confidence-weighted GP training on well-observed data and GP-guided regularization during appearance optimization.
This creates a self-reinforcing loop where reliable observations progressively refine motion priors, which reciprocally stabilize reconstruction in poorly observed regions.
Our main contributions are summarized as follows:
\begin{itemize}
\item We introduce a probabilistic framework that integrates GPs with 4DGS, enabling uncertainty quantification for motion predictions, temporal extrapolation for future motion prediction, and observation-adaptive motion priors.

\item We develop a spatio-temporal GP kernel, a scalable variational inference scheme, and a GP-GS dual optimization strategy for principled probabilistic dynamic scene reconstruction.

\item We demonstrate that our method improves dynamic scene reconstruction, especially in occluded or sparsely observed regions, while providing reliable motion extrapolation and uncertainty estimates.
\end{itemize}

\section{Related Work}
\label{sec:related}

\subsection{Dynamic Novel View Synthesis}

Novel View Synthesis (NVS) is a form of inverse graphics that reconstructs a 3D scene from observed 2D images and renders novel images at arbitrary viewpoints.
Dynamic Novel View Synthesis (DyNVS) extends this paradigm to 4D dynamic scenes, where the input is a video containing motion rather than static 2D images.
Early progress was driven by variants of Neural Radiance Fields (NeRF)~\cite{nerf,gao2021dynamic,du2021neural,park2021nerfies,kplanes_2023,shao2023tensor4d}, which implicitly represent 4D scenes via neural networks optimized through differentiable rendering.
More recently, 4D Gaussian Splatting (4DGS) methods~\cite{huang2023sc, yang2023deformable, wu20234d, li2023spacetime, lin2024gaussian, lu20243d, guo2024motion, liang2025gaufre, lei2024mosca, duan20244d, waczynska2024d, liu2024modgs, stearns2024dynamic} have emerged as the state-of-the-art by explicitly deforming Gaussian primitives over time with a deformation field.
For instance, D-3DGS~\cite{yang2023deformable} employs an MLP, 4DGS~\cite{wu20234d} uses a HexPlane representation, and STG~\cite{li2023spacetime} utilizes polynomial 
functions to model such deformations.
Some works~\cite{lei2024mosca, stearns2024dynamic} further incorporate priors from pretrained models for depth estimation and 2D tracking, yet these priors are mostly deterministic and confined to observed regions. 
In contrast, our method introduces a probabilistic framework that learns data-adaptive priors, enabling robust generalization even to unobserved regions.

\subsection{Probabilistic Modeling in Gaussian Splattings}

Gaussian Splatting~\cite{3dgs} represents scenes as a collection of Gaussian primitives, which is inherently a deterministic representation.
However, recent works~\cite{savant2024modeling,kheradmand20243d,van2024variational,guo2025gp} have begun to reformulate Gaussian Splatting as a probabilistic framework to quantify uncertainty or improve optimization. 
For instance, Stochastic GS~\cite{savant2024modeling} models Gaussian attributes as random variables with learnable variances, enabling uncertainty quantification. 
Kheradmand et al.~\cite{kheradmand20243d} reformulate Gaussian primitives as samples drawn from a spatial distribution and adopt Stochastic Gradient Langevin 
Dynamics (SGLD) to refine densification and reduce sensitivity to initialization.
VBGS~\cite{van2024variational} adopts a Bayesian-based Gaussian mixture model to solve the challenge of catastrophic forgetting in continual learning settings.
GP-GS~\cite{guo2025gp} integrates GP priors to improve Structure-from-Motion initialization by inferring missing regions via GP posteriors.
However, these probabilistic approaches are confined to static scene representations.
In contrast, our work is the first to integrate GP priors into 4D Gaussian Splatting, enabling probabilistic motion modeling for robust dynamic scene reconstruction.

\section{Preliminaries}
\label{sec:preliminary}

We briefly review 4D Gaussian Splatting and Gaussian Processes, which form the foundation of our method.

\subsection{4D Gaussian Splatting}

\paragraph{Gaussian primitives}

3DGS achieves high-quality real-time rendering of static scenes by employing an explicit 3D scene representation.
This representation consists of a set of $N$ 3D Gaussian primitives, denoted by $\Gamma = \left\{\gamma_1, \gamma_2, \ldots, \gamma_N \right\}$.
Each Gaussian primitive $\gamma_k$ is represented by an unnormalized 3D Gaussian kernel $\mathcal{G}_k(\mathbf{x}_s)$ as
\begin{align}
\label{eq:gaussian}
\hspace{-2mm}\mathcal{G}_k(\mathbf{x}_s; \! \boldsymbol{p}_k, \! \boldsymbol{\Sigma}_k) \!= \exp\left(\!-\frac{1}{2}(\mathbf{x}_s \!-\! \boldsymbol{p}_k)^\top \! \boldsymbol{\Sigma}_k^{-1} \! (\mathbf{x}_s \!-\! \boldsymbol{p}_k)\!\right),
\end{align}
where $\boldsymbol{p}_k \in \mathbb{R}^3$ is a mean vector, $\boldsymbol{\Sigma}_k \in \mathbb{R}^{3 \times 3}$ is an anisotropic covariance matrix, and $\mathbf{x}_s \in \mathbb{R}^3$ is an arbitrary location in 3D space.
The covariance matrix $\boldsymbol{\Sigma}_k$ has to be positive semi-definite, which is challenging to hold during optimization.
Instead, to ensure this condition, we learn $\boldsymbol{\Sigma}_k$ by decomposing it into two learnable components, a rotation matrix $\mathbf{R}_k$ and a diagonal scaling matrix $\mathbf{S}_k$ as follows: 
\begin{align}
\label{formula:covariance decomposition}
    \boldsymbol{\Sigma}_k = \mathbf{R}_k \mathbf{S}_k \mathbf{S}_k^\top \mathbf{R}_k^\top.
\end{align}
In addition to $\boldsymbol{p}_k, \mathbf{R}_k$, and $\mathbf{S}_k$, each Gaussian primitive requires two additional learnable parameters for its opacity $\alpha_k \in [0,1]$ and feature $\boldsymbol{f}_k $.
The feature vector is typically represented by RGB colors or spherical harmonic coefficients for rendering view-dependent lighting and color effects.
Consequently, a single Gaussian primitive $\gamma_k$ is defined with its complete set of learnable parameters, $\{\boldsymbol{p}_k, \mathbf{R}_k, \mathbf{S}_k, \alpha_k, \boldsymbol{f}_k\}$.

\vspace{-2mm}
\paragraph{Deformation modeling}

Dynamic scene modeling requires extending the 3D formulation to capture temporal variations.
Most algorithms~\cite{yang2023deformable, huang2023sc, wu20234d, li2023spacetime, wang2025shape} deform the 3D Gaussian primitives from their canonical states to a target state over time. 
The transformed position $\boldsymbol{p}_{k, t}$ and rotation $\mathbf{R}_{k, t}$ at time $t$ are given by
\begin{align}
    (\boldsymbol{p}_{k, t}, \mathbf{R}_{k, t}) \! = \! \left( \boldsymbol{p}_k \! + \! \phi_p(\boldsymbol{p}_k, \mathbf{R}_k, t), \mathbf{R}_k \! + \! \phi_r(\mathbf{R}_k, t) \right),
\end{align}
where $\phi_p(\cdot)$ and $\phi_r(\cdot)$ are the deformation operations.

\vspace{-2mm}
\paragraph{Differentiable rasterization}

Before rendering with a set of deformed Gaussian primitives $\Gamma$ on an image space, each deformed Gaussian kernel $\mathcal{G}_{k,t}(\mathbf{x}_s; \boldsymbol{p}_{k,t}, \boldsymbol{\Sigma}_{k,t})$ is projected onto a 2D image space and forms a 2D Gaussian kernel $\mathcal{G}^\pi_{k,t}(\mathbf{r}; \boldsymbol{p}_{k,t}^\pi, \boldsymbol{\Sigma}_{k,t}^\pi)$, where $\pi: \mathbb{R}^3 \to \mathbb{R}^2$ denotes a projection from a world coordinate to an image space and $t$ is the target time.
In the projected Gaussian representation, $\mathbf{r} \in \mathbb{R}^2$ indicates a pixel location in an image, and the 2D mean $\boldsymbol{p}_{k,t}^\pi \in \mathbb{R}^2$ and covariance $\boldsymbol{\Sigma}_{k,t}^\pi \in \mathbb{R}^{2 \times 2}$ are given by
\begin{align}
   \boldsymbol{p}_{k,t}^\pi = \pi(\boldsymbol{p}_{k,t}) \quad \text{and} \quad \boldsymbol{\Sigma}_{k,t}^\pi = \mathbf{J} \mathbf{W} \boldsymbol{\Sigma}_{k,t} \mathbf{W}^\top \mathbf{J}^\top,
\end{align}
where $\mathbf{J}$ denotes the Jacobian of the affine approximation of the projective transformation, and $\mathbf{W}$ is the world-to-camera transform matrix.
When rendering the primitives in $\Gamma$ to a target camera, they are sorted by their depths with respect to the camera center. 
The color of a pixel $\mathbf{r}$ is then obtained by $\alpha$-blending, which is given by
\begin{align}
    \label{eq:color_render}
    \mathbf{\hat I}(\mathbf{r}) = \sum_{k = 1}^N  c_{k} \; \omega^\pi_{k,t}(\mathbf{r}),
\end{align}
where $\omega^\pi_{k,t}(\mathbf{r})$ represents a relative contribution of each Gaussian primitive to pixel $\mathbf{r}$ at time $t$ and $c_k$ is the color of the corresponding primitive.

\subsection{1D Gaussian Processes}

A Gaussian Process (GP) defines a probability distribution over functions, such that any finite collection of function values follows a joint Gaussian distribution as follows: 
\begin{align}
    {f(x) \sim \mathcal{GP}(m(x),\, k(x, x')),}
\end{align}
where $x \in \mathbb{R}$ and $f(x) \in \mathbb{R}$ are both scalar-valued in the simplest case, $m(x)$ and $k(x, x')$ are the mean function and kernel function, respectively.
Given observations $\mathcal{D} = \{(x_n, y_n)\}_{n=1}^N$ with $\mathbf{X} = \{x_n\}_{n=1}^N$ and $\mathbf{y} = \{y_n\}_{n=1}^N$, {the kernel hyperparameters are optimized by maximizing the marginal likelihood to capture the correlation structure of the observed data.}

For inference, the posterior distribution at a new query point $x^*$ is Gaussian,
$f(x^*) \sim \mathcal{N}(\bar{\mu}(x^*), \bar{\sigma}^2(x^*))$,
conditioned on the observed data $\mathcal{D}$.
The predictive mean and variance are respectively derived as 
\begin{align}
    \bar{\mu}(x^*) &= m(x^*) + \mathbf{k}_*^\top 
                           (\mathbf{K} + \sigma_n^2 \mathbf{I})^{-1}(\mathbf{y} - m(\mathbf{X})),  \\
    \bar{\sigma}^2(x^*) &= k(x^*, x^*) - \mathbf{k}_*^\top 
                           (\mathbf{K} + \sigma_n^2 \mathbf{I})^{-1} \mathbf{k}_*,
\end{align}
where $\mathbf{K} \in \mathbb{R}^{N \times N}$ is the covariance matrix with entries $[\mathbf{K}]_{ij} = k(x_i, x_{j})$, and $\mathbf{k}_* \in \mathbb{R}^N$ denotes the vector of covariances between $x^*$ and the training inputs. 
Intuitively, the predictive mean $\bar{\mu}(x^*)$ is a weighted combination of the observed residuals $(\mathbf{y} - m(\mathbf{X}))$ in output space, where the weights are determined by the kernel-defined similarity between the query point and the training data.

%
 
%
%

\section{Method}
\label{sec:method}

We present GP-4DGS, a principled integration of GPs into 4DGS for monocular video reconstruction.
Specifically, we design spatial and temporal kernels to capture geometric and motion correlations in primitive deformations in Section~\ref{subsec:gp_modeling}, adopt variational inference for scalable computation in Section~\ref{subsec:VGP}, and introduce a GP-GS optimization strategy for joint training of GP and 4DGS in Section~\ref{subsec:gp_training}.
This naturally provides uncertainty quantification and temporal extrapolation, capabilities absent in existing 4DGS frameworks, as described in Sections~\ref{subsec:uncertainty} and~\ref{subsec:extrapolation}.

\subsection{Motion Modeling with Gaussian Processes}
\label{subsec:gp_modeling}

We model the temporal deformation of Gaussian primitives with a multi-input multi-output GP, using a composite kernel with spatial and temporal components for canonical geometry and motion periodicity, respectively.

\subsubsection{Probabilistic Deformation}

Given a 4D input $\mathbf{x} = (\boldsymbol{p}, t) \in \mathbb{R}^4$, where $\boldsymbol{p} = (p_x, p_y, p_z)$ denotes the canonical 3D position of an arbitrary primitive and $t$ represents the target time, our GP outputs a $d$-dimensional deformation vector $\mathbf{y} = (f_1(\mathbf{x}), \ldots, f_{d}(\mathbf{x})) \in \mathbb{R}^d$.
We set $d = 9$ with three dimensions for translational deformation and six for the 6D continuous rotation representation~\cite{zhou2019continuity} of Gaussian primitive orientations.
Each output $f_i(\mathbf{x})$ is modeled as an independent GP with mean function $m_i(\mathbf{x})$ and kernel function $k_i(\mathbf{x}, \mathbf{x}')$ as follows:
\begin{align}
f_i(\mathbf{x}) \sim \mathcal{GP}(m_i(\mathbf{x}), k_i(\mathbf{x}, \mathbf{x}')).
\end{align}
This defines a probabilistic distribution over functions, where a prediction at any input $\mathbf{x}$ is Gaussian-distributed, $f_i(\mathbf{x}) \sim \mathcal{N}(\bar{\mu}_i(\mathbf{x}), \bar{\sigma}_i^2(\mathbf{x}))$, with both mean and variance derived from the GP posterior.

\subsubsection{Kernel Modeling}

A kernel defines the correlation structure between GP inputs, determining how the model generalizes from observed to unobserved regions.
While standard GP kernels assume isotropic correlations across all input dimensions, our input space exhibits fundamentally different correlation characteristics between the spatial dimensions $\boldsymbol{p}$ and the temporal dimension $t$.
To capture this heterogeneity, we propose a composite kernel that separates spatial and temporal components as follows:
\begin{align}
k_i(\mathbf{x}, \mathbf{x}') = k_{i}^{\text{spatial}}(\boldsymbol{p}, \boldsymbol{p}') + k_{i}^{\text{temporal}}(\mathbf{x}, \mathbf{x}'),
\end{align}
where $\boldsymbol{p} = (p_{x}, p_{y}, p_{z})$ denotes the spatial canonical component in 3D, and $i \in \{1, \ldots, d\}$ is the index of the  function.

\vspace{-2mm}
\paragraph{Spatial correlations}
To capture smooth geometric correlations in canonical space, we adopt the Mat\'ern kernel:
\begin{align}
k_{i}^{\text{spatial}}(\boldsymbol{p}, \boldsymbol{p}') = \sigma_{s,i}^2 
\frac{2^{1-\nu_i}}{\Gamma(\nu_i)} r_{s,i}^{\nu_i} K_{\nu_i}(r_{s,i}),
\end{align}
where $r_{s,i} = \sqrt{2\nu_i \sum_{j\in\{x, y, z\}} (p_{j} -  p'_{j})^2 / \ell_{s,i,j}^2}$ is the anisotropic scaled distance, $K_{\nu_i}$ is the modified Bessel function, $\sigma_{s,i}^2$ is the signal variance, and $\nu_i$ controls smoothness.
This encodes the prior that nearby primitives in canonical space exhibit similar deformations.
Note that we choose the Mat\'ern over the RBF kernel for its ability to handle discontinuities, which is essential for modeling spatially disconnected objects in dynamic scenes.

\vspace{-2mm}
\paragraph{Temporal correlations}
To capture motion patterns along time, we model temporal correlations as a sum of per-axis Mat\'ern kernels weighted by a periodic kernel as follows:
\begin{align}
\hspace{-2mm} k_{i}^{\text{temporal}}(\mathbf{x}, \mathbf{x}') = \hspace{-3mm}
\sum_{j \in \{x, y, z \}} \hspace{-2mm} k_{i,j}({p}_j, p'_{j}) \ k_{i,j}^{\text{periodic}}(t, t'),
\end{align}
where $k_{i,j}$ is a one-dimensional Mat\'ern kernel over the corresponding axis, and $k_{i,j}^{\text{periodic}}$ is a periodic kernel over time $t$, defined as follows:
\begin{align}
k_{i,j}^{\text{periodic}}(t, t') \!= \! \sigma_{i,j}^2 
\exp\left(-\frac{2\sin^2(\pi|t-t'|/\tau_{i,j})}{\ell_{i,j}^2}\right),
\end{align}
with period $\tau_{i,j}$, length scale $\ell_{i,j}$, and signal variance $\sigma_{i,j}^2$.
The periodic kernel provides strong inductive bias for temporal extrapolation by capturing patterns in motion, enabling robust predictions beyond the observed time range.

\subsection{Variational Gaussian Processes for 4DGS}
\label{subsec:VGP}

Standard GP posterior computation requires constructing an $N \times N$ kernel matrix and $\mathcal{O}(N^3)$ matrix inversion, where $N$ is the number of data points.
In our setting, $N$ corresponds to the number of Gaussian primitives, which typically reaches tens of thousands, making exact inference computationally intractable.
To address this, we adopt variational GPs~\cite{titsias2009variational}.

\vspace{-2mm}
\paragraph{Inducing points}
We approximate the full GP with $M$ inducing points $\mathbf{Z} = \{\mathbf{z}_m\}_{m=1}^M$, $\mathbf{z}_m \in \mathbb{R}^4$, $M \ll N$, which serve as a learned summary of the deformation field in the 4D input space $\mathbf{x} \! = \! (\mathbf{x}_{s}, t)$.
With this approximation, we only compute two kernel matrices: $\mathbf{K}_{ZZ}^{(i)} = \mathbf{K}_i(\mathbf{Z}, \mathbf{Z}) \in \mathbb{R}^{M \times M}$ among inducing points and $\mathbf{K}_{XZ}^{(i)} = \mathbf{K}_i(\mathbf{X}, \mathbf{Z}) \in \mathbb{R}^{N \times M}$ between primitives and inducing points.
This reduces complexity from $\mathcal{O}(N^3)$ to $\mathcal{O}(NM^2 + M^3)$, making inference tractable even with tens of thousands of Gaussian primitives.

For initialization of $\mathbf{Z}$, spatial locations $\mathbf{x}_{s}$ are selected by extracting time-series features from primitive trajectories using Chronos~\cite{ansari2024chronos} and clustering them via $k$-means to select $M$ representative canonical positions.
Temporal locations $t$ are sampled uniformly over the observed time range.

\vspace{-2mm}
\paragraph{Training}
In addition to the inducing point locations $\mathbf{Z}$, we parameterize a variational posterior $q(\mathbf{u}_i) = \mathcal{N}(\mathbf{m}_i, \mathbf{S}_i)$ over the function values at all inducing points, where $\mathbf{u}_i \in \mathbb{R}^M$ collects the values for output dimension $i \in \{1, \ldots, d\}$.  
The kernel hyperparameters (\emph{e.g.}, length scales, signal variances, and periods), inducing point locations  $\mathbf{Z}$, and variational parameters $\{(\mathbf{m}_i, \mathbf{S}_i)\}_{i=1}^d$ are jointly optimized via the ELBO as follows:
\begin{align}
\label{eq:elbo}
\mathcal{L}_{\text{ELBO}} \!=\! \sum_{i=1}^{d} \left[ \mathbb{E}_{q} [\log p(\mathbf{y}_i|\mathbf{u}_i)] - \text{KL}[q(\mathbf{u}_i) \| p(\mathbf{u}_i)] \right],
\end{align}
where $p(\mathbf{u}_i) = \mathcal{N}(\mathbf{0}, \mathbf{K}_{ZZ}^{(i)})$ is the GP prior, the expectation term encourages fitting the observed deformations, and the KL term regularizes the posterior toward the prior.

\vspace{-2mm}
\paragraph{Inference}

After training, given an arbitrary query point $\mathbf{x}^*$, we compute the mean and variance of the deformation from the variational posterior as follows:
\begin{align}
\label{eq:inference}
\bar{\mu}^*_i & = \mathbf{k}_*^{(i)\top} \! (\mathbf{K}_{ZZ}^{(i)})^{-1} {\mathbf{m}_i}, \\
\bar{\sigma}^{* 2}_i & = k_i^* \!-\! \mathbf{k}_*^{(i)\top} \boldsymbol{\Sigma}_i \mathbf{k}_*^{(i)},
\end{align}
where $\mathbf{k}_*^{(i)} = \mathbf{k}_i(\mathbf{Z}, \mathbf{x}^*) \in \mathbb{R}^M$ is the cross-covariance vector between the inducing points and the query, $k_i^* = k_i(\mathbf{x}^*, \mathbf{x}^*)$ is the kernel variance at $\mathbf{x}^*$, and $\boldsymbol{\Sigma}_i = (\mathbf{K}_{ZZ}^{(i)})^{-1} - (\mathbf{K}_{ZZ}^{(i)})^{-1}\mathbf{S}_i(\mathbf{K}_{ZZ}^{(i)})^{-1}$.
This formulation scales as $\mathcal{O}(M)$ per query $\mathbf{x}^*$, enabling efficient inference without accessing all training primitives.
The predicted mean and variance vectors over all output dimensions are $\boldsymbol{\bar{\mu}} = [\bar{\mu}_1, \ldots, \bar{\mu}_{d}]^\top $ and $\boldsymbol{\bar{\sigma}}^2 = [\bar{\sigma}^2_1, \ldots, \bar{\sigma}^2_{d}]^\top $.

\subsection{GP-GS Optimization}
\label{subsec:gp_training}

This section describes our GP-GS optimization strategy, as outlined in Algorithm~\ref{alg:gp_gs}, which alternates between GP training on confident observations and 4DGS optimization guided by GP predictions, creating a feedback loop that progressively refines both representations.
\begin{algorithm}[t]
\caption{GP-GS Optimization Strategy}
\label{alg:gp_gs}
\begin{algorithmic}[1]
    \Require Training images $\mathcal{T}$, Gaussian primitives $\{\gamma_k\}_{k = 1}^N$
    \Ensure Optimized GS model with GP regularization
    \While{not converged}
        \State \textbf{Stage 1: GP Training} \hspace{3mm} // {For every $N_{\text{GP}}$ iterations}
        \State Compute confidence: $C_k = \sum_{\mathbf{I} \in \mathcal{T}} \sum_{\mathbf{r} \in \mathbf{I}} \omega^\pi_k(\mathbf{r})$
        \State Select confident data: $\mathcal{D}_C = \{(\mathbf{x}_i, \mathbf{y}_i) : C_k > \tau_C\}$
        \State Train variational GPs on $\mathcal{D}_C$ with Eq.~\eqref{eq:elbo}
        \State \textbf{Stage 2: GS Optimization}
        \State Infer variational GP posteriors and cache $\bar{\boldsymbol{\mu}}_{(k,t)}(\mathbf{x})$
        \State Compute loss: $\mathcal{L}_{\text{GP}} \! = \! \mathbb{E}[ \delta_k \cdot \|\mathbf{y}_{(k,t)} - \bar{\boldsymbol{\mu}}_{(k,t)}\|^2]$
        \State Update GS: $\mathcal{L}_{\text{total}} = \mathcal{L}_{\text{recon}} + \lambda_{\text{GP}} \mathcal{L}_{\text{GP}}$
    \EndWhile
    \State \Return optimized GS and GP models
\end{algorithmic}
\end{algorithm}

\subsubsection{Stage 1: GP Optimization}

\paragraph{Response-aware data sampling}
We measure the informativeness of each Gaussian primitive via its cumulative contribution to pixel rendering as follows:
\begin{align} \label{eq:visibility}
    C_k = \sum_{\mathbf{I} \in \mathcal{T}} \sum_{\mathbf{r} \in \mathbf{I}} \omega^\pi_{k,t}(\mathbf{r}),
\end{align}
where $\mathbf{r}$ is a pixel in a training image $\mathbf{I} \in \mathcal{T}$, $\omega^\pi_{k,t}(\mathbf{r})$ is the $\alpha$-blending weight of primitive $\gamma_k$ at pixel $\mathbf{r}$, and $t$ is captured time of image $\mathbf{I}$.
We select primitives with $C_k > \tau_C$ to form a confident subset $\mathcal{D}_C \subset \mathcal{D}$.
The GP is then trained on $\mathcal{D}_C$ by maximizing the ELBO in Eq.~\ref{eq:elbo}, optimizing the kernel hyperparameters, inducing point locations $\mathbf{Z}$, and variational parameters $\{(\mathbf{m}_i, \mathbf{S}_i)\}_{i=1}^d$.

\vspace{-2mm}
\paragraph{Handling noisy inputs}
Since the spatial coordinates  $\boldsymbol{p} = (p_x, p_y, p_z)$ are also optimized parameters rather than exact positions, we inject Gaussian noise into each spatial dimension during 
GP training as follows:
\begin{align}
\tilde{\mathbf{x}} = (p_x + \epsilon_x,\; p_y + \epsilon_y,\; p_z + \epsilon_z,\; t),
\end{align}
where $\epsilon_x, \epsilon_y, \epsilon_z \sim \mathcal{N}(0, 0.02)$.
This acts as regularization, enabling the GP to handle uncertainty in canonical positions and produce more robust predictions.

\subsubsection{Stage 2: GS Optimization}

\paragraph{GP inference}
After training on $\mathcal{D}_C$, we run GP inference on all primitives.
For each primitive $k$ at time $t$ with input $\mathbf{x}^*_{(k,t)} = (\boldsymbol{p}_k, t)$, we compute the posterior mean $\boldsymbol{\bar{\mu}}_{(k,t)}^*$, which serves as a pseudo-guidance signal.
To balance computational efficiency and guidance freshness, we cache GP predictions every $N_{GP} = 2000$ iterations during GS optimization. Between updates, cached predictions are reused across optimization steps.

\vspace{-2mm}
\paragraph{GP guidance loss}
We regularize GS optimization with a GP guidance loss as follows:
\begin{align}
\mathcal{L}_{\text{GP}} = \frac{1}{NT} \sum_{k=1}^{N} \sum_{t=1}^{T} \delta_{(k,t)} \cdot \left\| \mathbf{y}_{(k,t)} - \boldsymbol{\bar{\mu}}_{(k,t)}^*\right\|^2,
\end{align}
where $\delta_{(k,t)} = \mathbf{1}(\|\mathbf{y}_{(k,t)} - \boldsymbol{\bar{\mu}}_{(k,t)}^*\| > \tau_{\delta})$ selects primitives that deviate from GP motion predictions $\boldsymbol{\bar{\mu}}_{(k,t)}^*$.
The threshold $\tau_{\delta}$ is annealed during optimization from $\tau_{\delta, \text{start}}$ to $\tau_{\delta, \text{end}}$, progressively tightening constraints as the two representations converge.
The total training loss is $\mathcal{L}_{\text{total}} = \mathcal{L}_{\text{recon}} + \lambda_{\text{GP}}\mathcal{L}_{\text{GP}}$, where $\mathcal{L}_{\text{recon}}$ follows SoM~\cite{wang2025shape}, consisting of photometric, D-SSIM, flow, and smoothness losses.
The balancing weight $\lambda_{\text{GP}}$ is set to 0.1.

\subsection{Interpretability via Uncertainty Quantification}
\label{subsec:uncertainty}

\begin{figure}[t]
     \centering
          \begin{subfigure}[b]{0.24\linewidth}
         \centering
        \includegraphics[width=\linewidth]{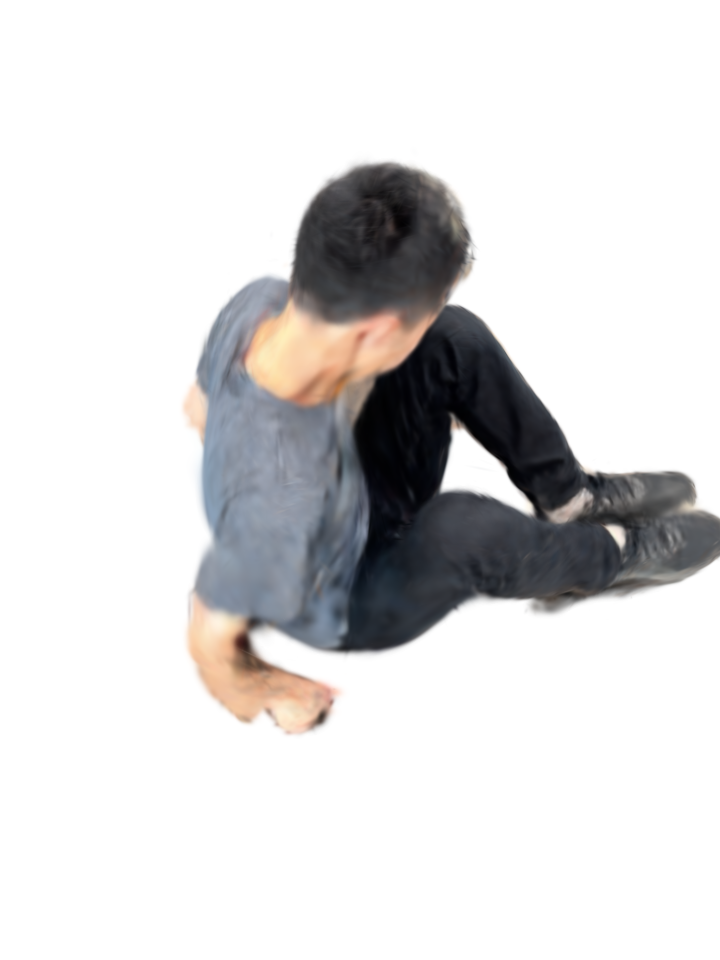}
        \includegraphics[width=\linewidth]{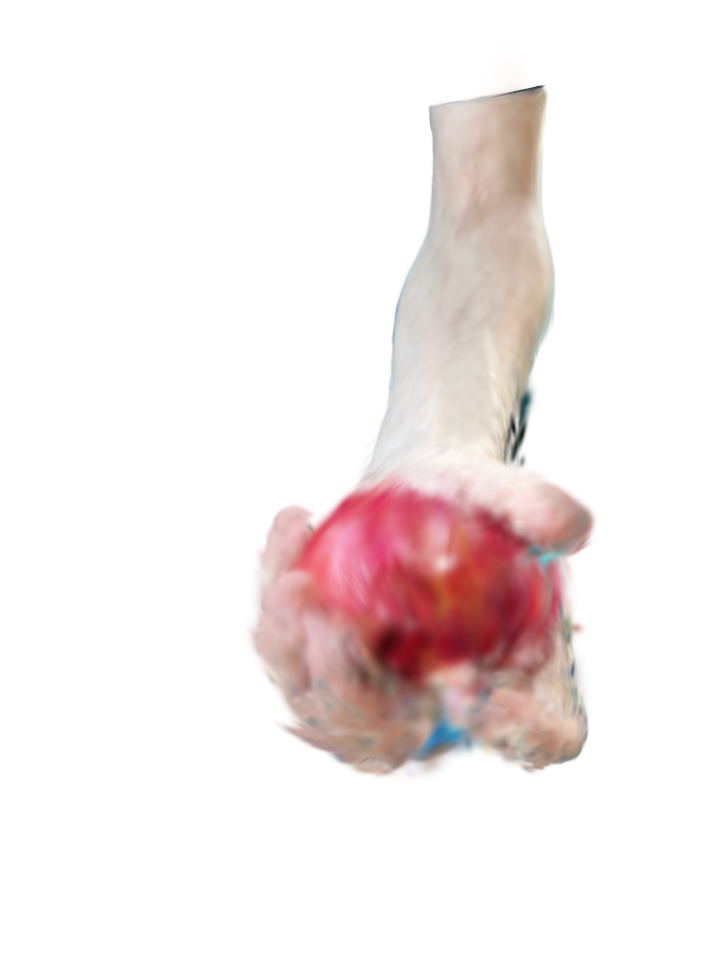}
 \caption*{Rendering}
     \end{subfigure}
          \begin{subfigure}[b]{0.24\linewidth}
         \centering
        \includegraphics[width=\linewidth]{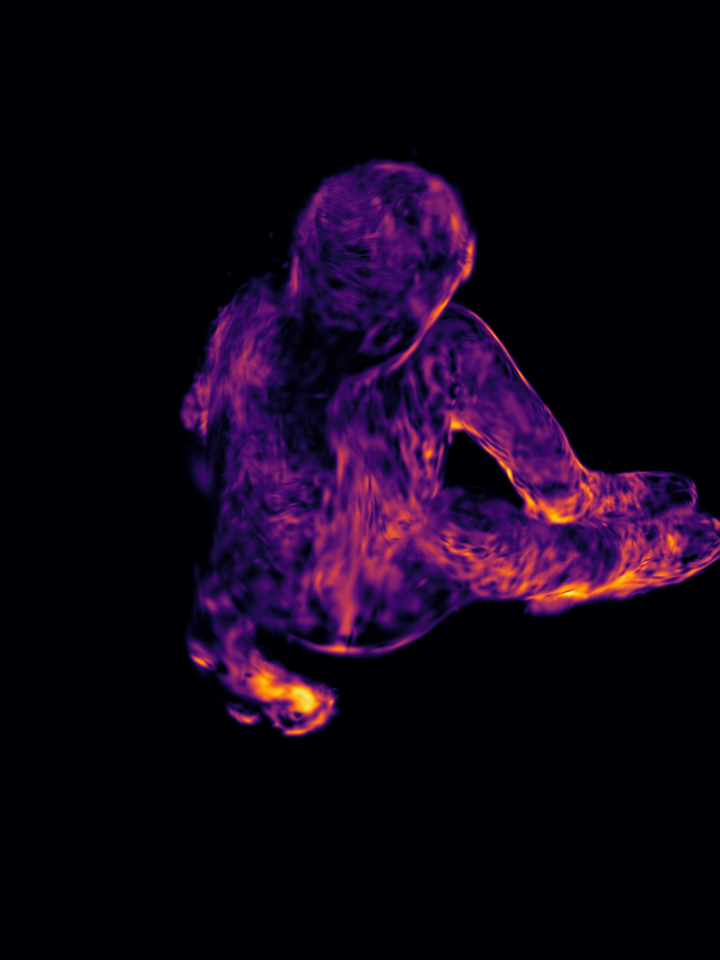}
        \includegraphics[width=\linewidth]{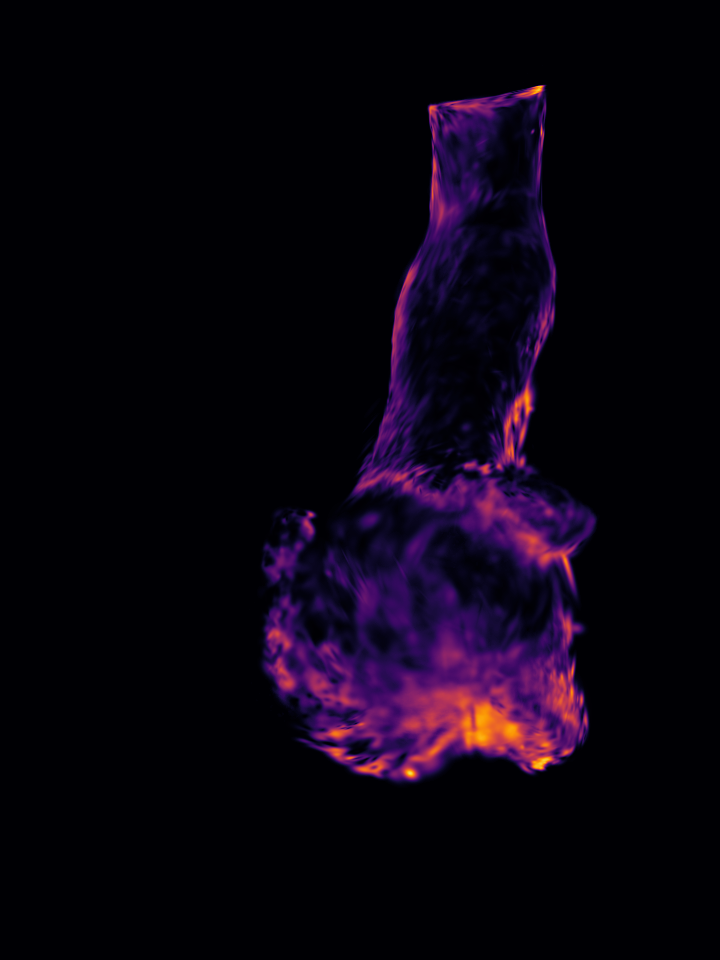}
 \caption*{Uncertainty}
     \end{subfigure}
          \begin{subfigure}[b]{0.24\linewidth}
         \centering
        \includegraphics[width=\linewidth]{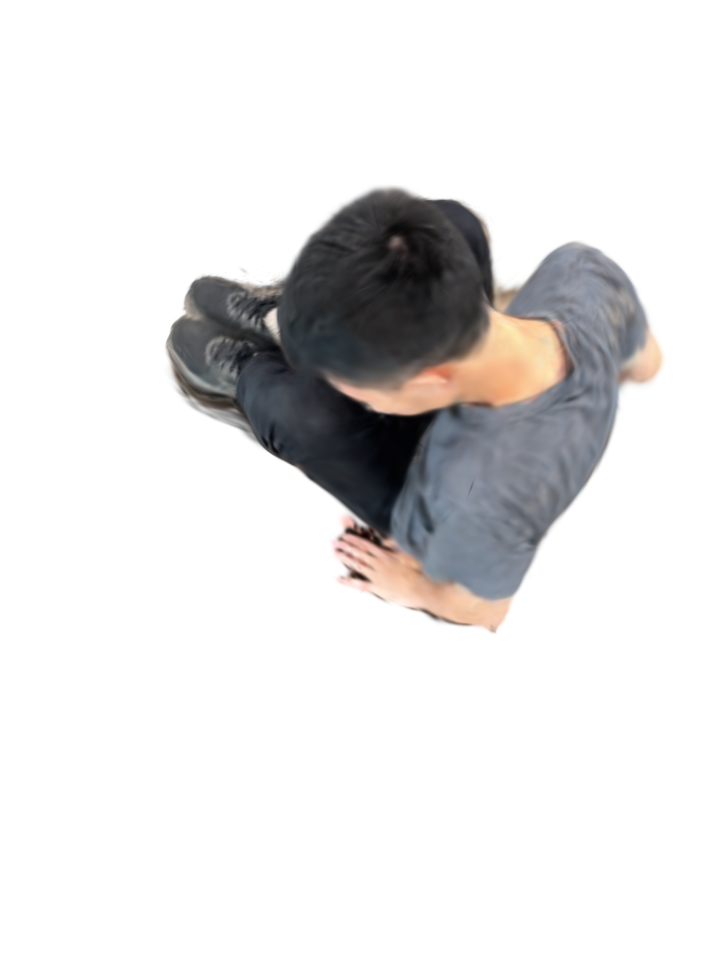}
        \includegraphics[width=\linewidth]{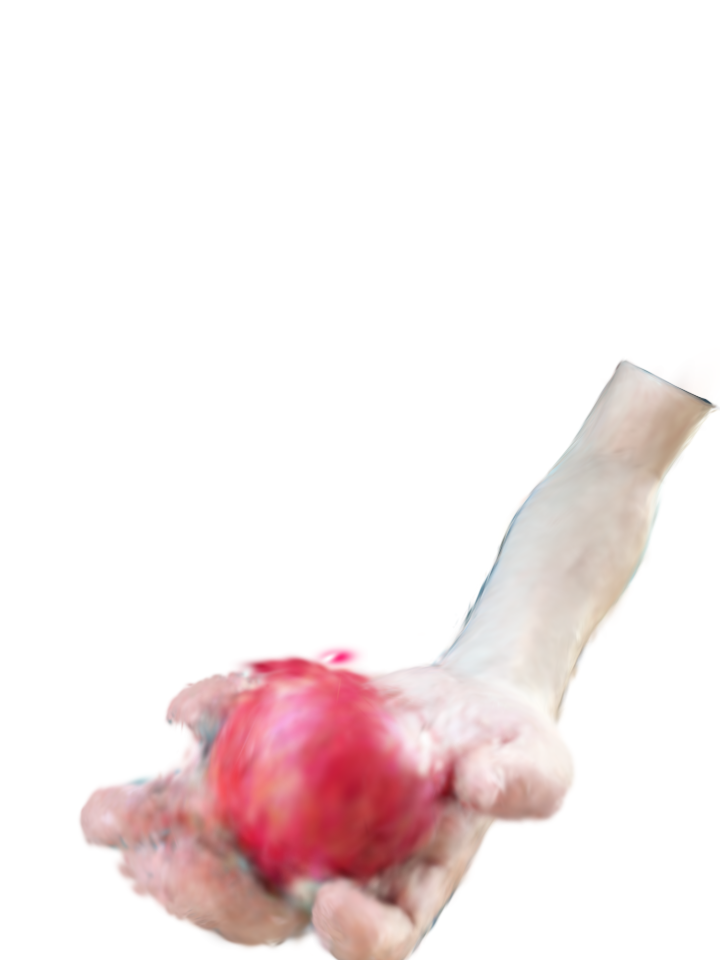}
 \caption*{Rendering}
     \end{subfigure}
     \begin{subfigure}[b]{0.24\linewidth}
         \centering
        \includegraphics[width=\linewidth]{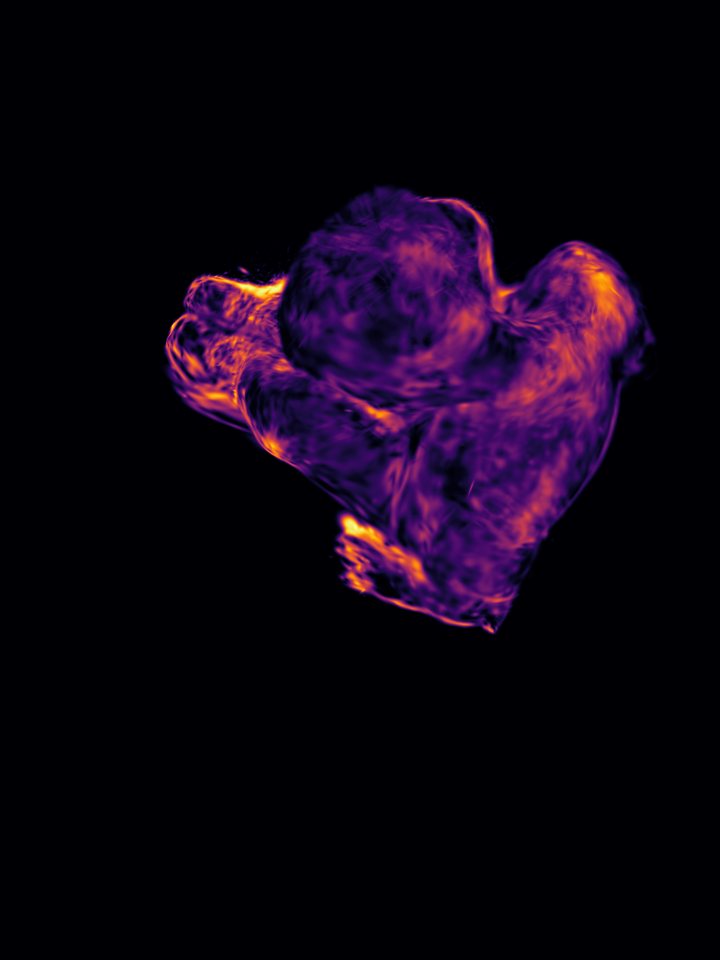}
        \includegraphics[width=\linewidth]{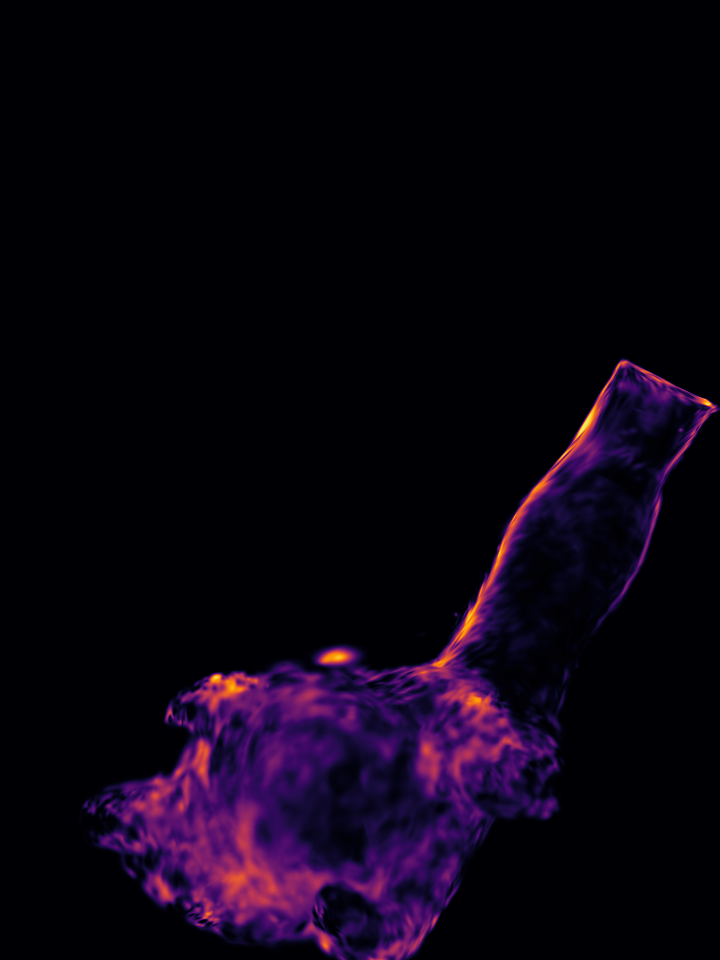}
 \caption*{Uncertainty}
     \end{subfigure}
     \vspace{-1mm}
        \caption{Uncertainty quantification. GP-4DGS provides principled uncertainty estimates for motion, a capability inherently lacking in existing 4DGS methods.
        }
        \label{fig:uncertainty}
\end{figure}

As GPs define a distribution over functions, they naturally provide variance estimates alongside mean predictions, which we use to quantify the uncertainty of primitive motions.
For translation, uncertainty is directly obtained from the first three dimensions of $\boldsymbol{\bar{\sigma}}^*$.
For rotation, the 6D-to-matrix conversion via Gram-Schmidt orthogonalization is non-linear, precluding direct variance propagation.
Instead, we use Monte Carlo sampling to approximate uncertainty in the final positions as follows:
\begin{align}
U_{k,t} = \text{Var}(\{   \boldsymbol{p}^{(s)}_{k, t}   \}_{s=1}^{S}),
\end{align}
where $\boldsymbol{p}^{(s)}_{k, t}$ is the $s^\text{th}$ deformed position of primitive $k$ across $S$ sampled deformations at time $t$.
We then render a motion uncertainty map by projecting $U_{k,t}$ to a target view as follows:
\begin{align}
\label{eq:uncertainty_map}
\mathbf{\hat{U}}(\mathbf{r}) = \sum_{k=1}^{N} U_{k,t} \; \omega^\pi_{k,t}(\mathbf{r}),
\end{align}
which is analogous to color rendering in Eq.~\eqref{eq:color_render}, and examples are shown in Figure~\ref{fig:uncertainty}.

\subsection{Extrapolation for Future Motion Prediction}
\label{subsec:extrapolation}

A key advantage of GP-based motion reasoning is the ability to predict future motion beyond the training sequence.
Given a canonical position and a future timestamp $t_f$, we directly query the trained GP in Eq.~\eqref{eq:inference} with $\mathbf{x}_{f}^{*} = (\boldsymbol{p}, t_f)$ to obtain the deformation at that time.
This enables motion forecasting without additional training or architectural modifications.

\section{Experiments}
\label{sec:experiments}
This section evaluates the efficacy of GP-4DGS, demonstrating that the GP integration provides superior structural priors for sparse-view dynamic reconstruction while achieving reliable uncertainty estimation and motion prediction.

\subsection{Experimental Settings}
We utilize DyCheck~\cite{gao2022monocular}, a benchmark comprising handheld monocular videos with rapid motion that presents challenging realistic scenarios for dynamic scene reconstruction. 
To evaluate robustness under extreme viewpoint shifts, we qualitatively assess our method on DAVIS~\cite{pont20172017}, where camera poses are estimated using Mega-SAM~\cite{li2025megasam}. 
Our variational GP framework is implemented using GPyTorch~\cite{gardner2018gpytorch}.

\subsection{Performance of Dynamic Novel View Synthesis}

\paragraph{Results on DyCheck}
\begin{table}[t]
\centering
\caption{Quantitative results on the DyCheck dataset. 
We evaluate performance on all seven scenes (All), the five scenes used in SoM~\cite{wang2025shape} (SoM 5), and a challenging subset with reduced viewpoint overlap. 
GP-4DGS consistently achieves superior results, particularly under sparse observations.}
\label{table:quantitative_dycheck}
\vspace{-1mm}
\hspace{-3mm}
\scalebox{0.8}{
\setlength\tabcolsep{3pt} 
\begin{tabular}{ c  l  c  c  c }
\toprule
{Data} & \multicolumn{1}{c}{{Method}} & {mPSNR} $\uparrow$ & {mSSIM} $\uparrow$ & {mLPIPS} $\downarrow$ \\
\midrule
\multirow{3}{*}{All} 
& Gaussian Marbles~\cite{stearns2024dynamic}  & 15.84 &0.54 &0.57 \\ 
& SoM~\cite{wang2025shape} & 17.09 & 0.65 & 0.39 \\
& GP-4DGS (ours) &\textbf{17.38} & \textbf{0.65} & \textbf{0.37} \\ 
\midrule
\multirow{9}{*}{SoM 5} 
& SC-GS~\cite{li2023spacetime} & 14.13 & 0.48 & 0.49\\
& D-3DGS~\cite{yang2023deformable} &11.92 & 0.49 & 0.66\\
& 4DGS~\cite{wu20234d} &13.42 &0.49 &0.56\\
& T-NeRF~\cite{wang2025shape}~ & 15.60 & 0.55 & 0.55 \\
& HyperNeRF~\cite{park2021hypernerf} & 15.99 & 0.59 &0.51 \\
& DynIBaR~\cite{li2023dynibar} & 13.41 &0.48& 0.55 \\
& Gaussian Marbles~\cite{stearns2024dynamic} & 16.03 & 0.54 & 0.58 \\
& SoM~\cite{wang2025shape} & 16.73 & 0.64 & 0.43 \\
 & GP-4DGS (ours) & \textbf{16.92} & \textbf{0.66} & \textbf{0.41} \\
\midrule
\multirow{3}{*}{\shortstack{Challenging \\ subset~\cite{huang2025vivid4d}}} 
& Gaussian Marbles~\cite{stearns2024dynamic} & 14.05 & 0.40 & 0.61 \\
& SoM~\cite{wang2025shape} & 14.56 & \textbf{0.46} & 0.53 \\
 & GP-4DGS (ours) & \textbf{15.02} & \textbf{0.46} & \textbf{0.51} \\
\bottomrule
\end{tabular}
}
\vspace{-2mm}
\end{table}
\begin{figure}[t]
    \centering
    \noindent
    \hspace{-2mm}
    \begin{minipage}{0.04\linewidth}
\rotatebox[origin=c]{90}{\raggedleft \small  \hspace{0.9cm}  wheel \hspace{1.7cm} spin \hspace{1.1cm} paper-windmill}
\end{minipage}\hspace{-1.7mm}%
    \begin{minipage}{0.98\linewidth}
        \centering
        \begin{subfigure}[b]{0.23\linewidth}
            \centering
            \includegraphics[width=\linewidth]{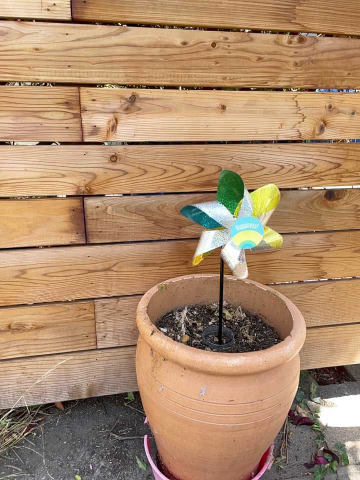}
            \includegraphics[width=\linewidth]{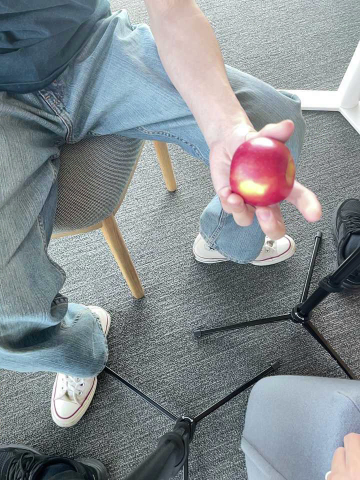}
            \includegraphics[width=\linewidth]{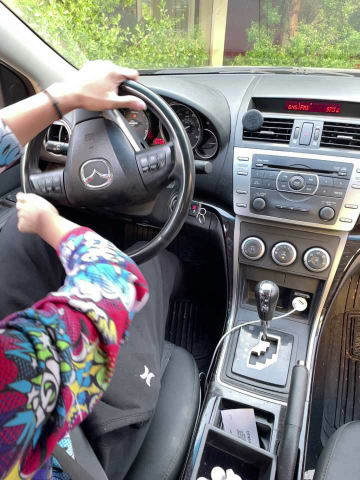}
                      \caption*{Ground Truth}
        \end{subfigure}\hspace{0.07cm}%
        \begin{subfigure}[b]{0.23\linewidth}
            \centering
           \includegraphics[width=\linewidth]{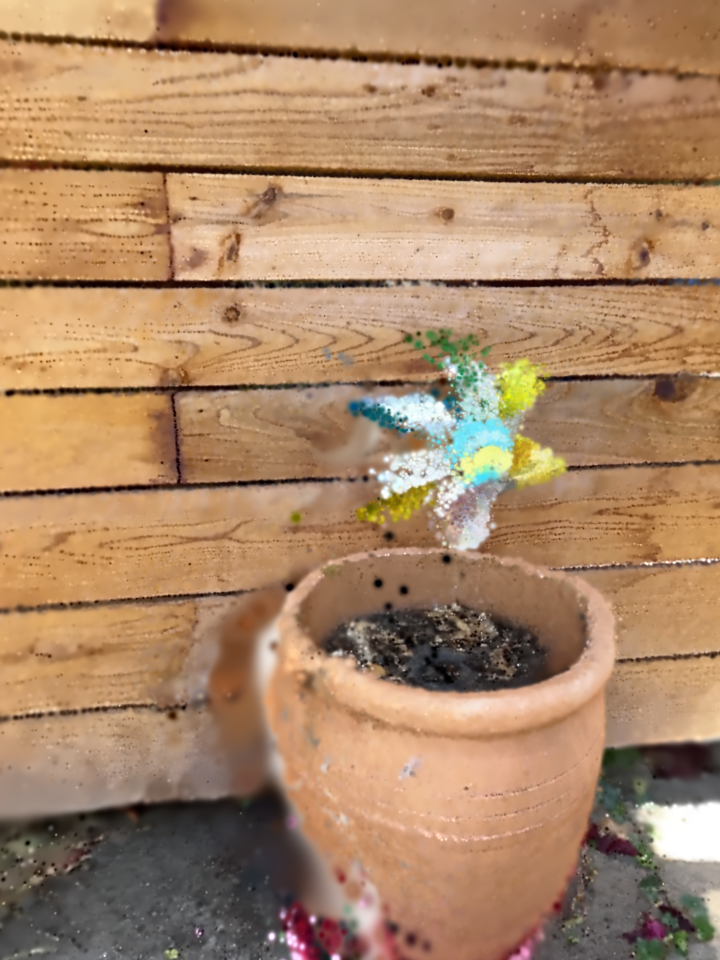}
                       \includegraphics[width=\linewidth]{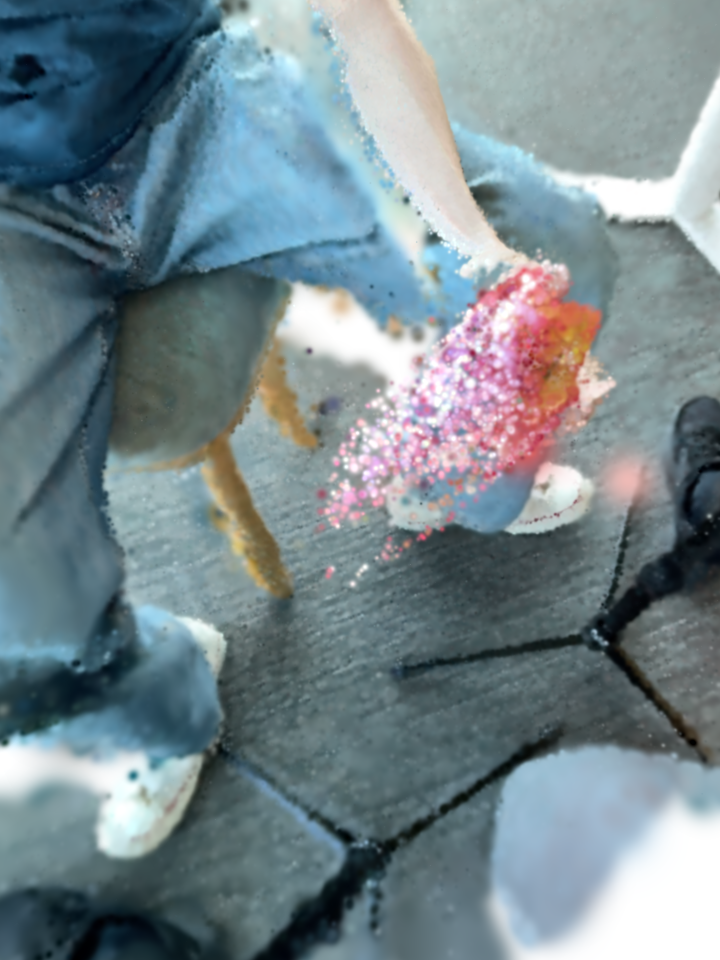}
             \includegraphics[width=\linewidth]{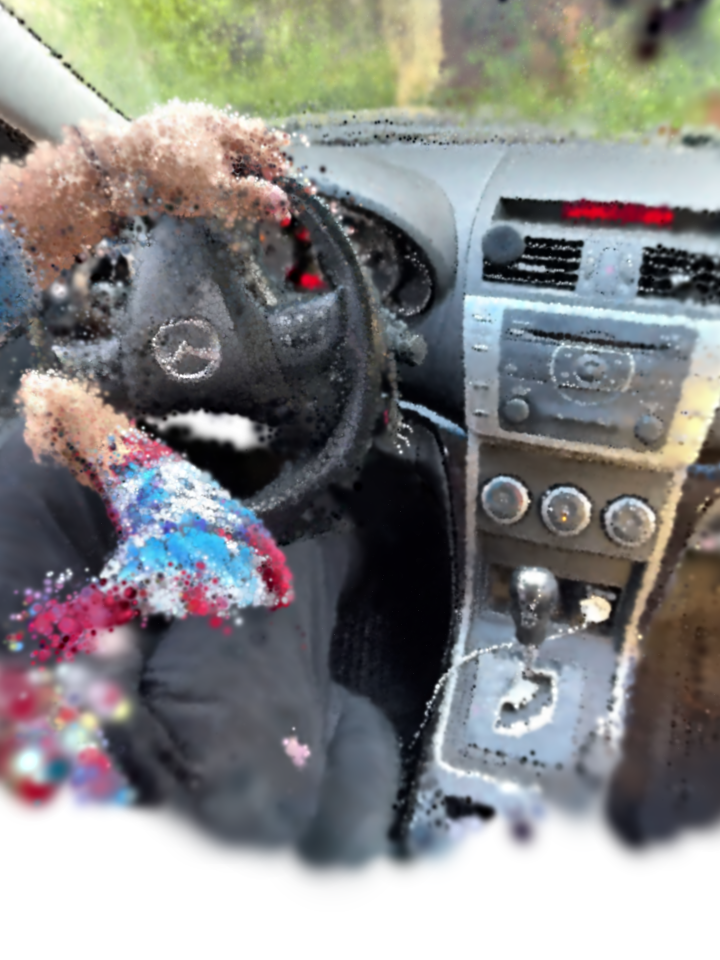}
            \caption*{GM~\cite{stearns2024dynamic}}
        \end{subfigure}\hspace{0.07cm}%
        \begin{subfigure}[b]{0.23\linewidth}
            \centering
                \includegraphics[width=\linewidth]{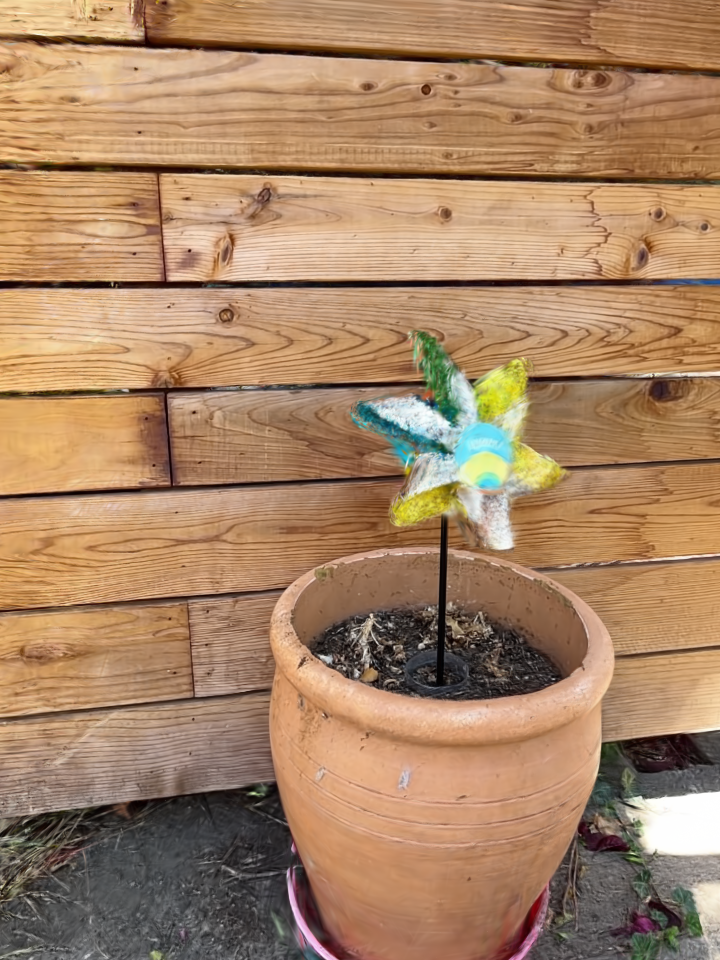}
             \includegraphics[width=\linewidth]{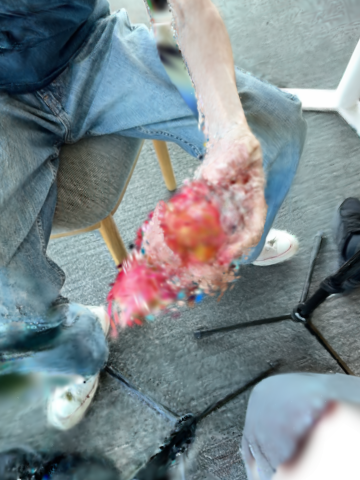}
                \includegraphics[width=\linewidth]{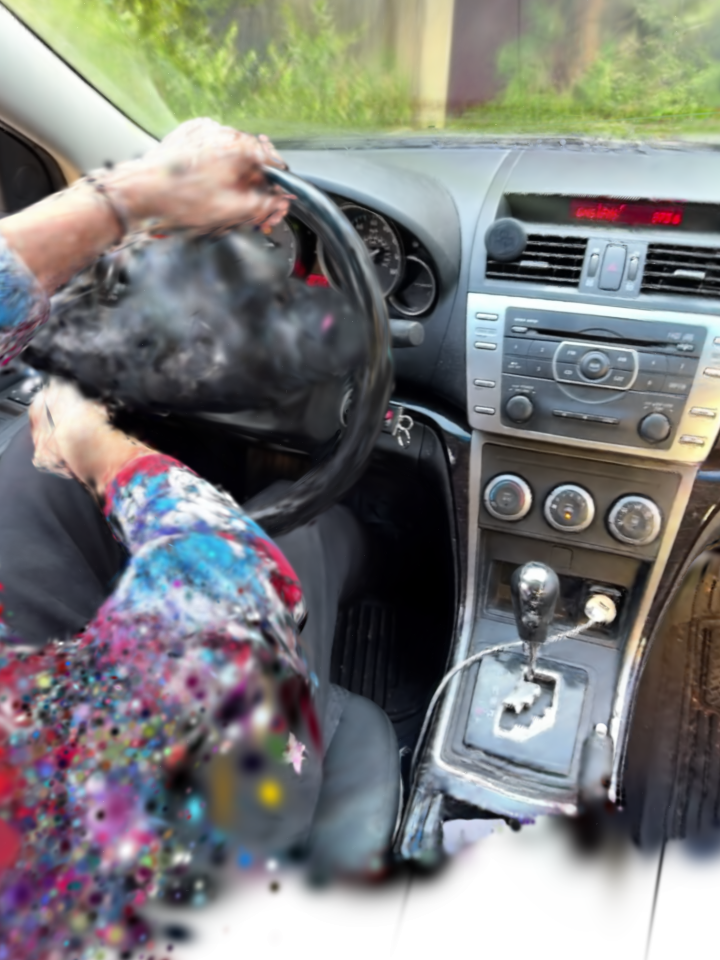}
            \caption*{SoM~\cite{wang2025shape}}
        \end{subfigure}\hspace{0.07cm}%
        \begin{subfigure}[b]{0.23\linewidth}
            \centering
                \includegraphics[width=\linewidth]{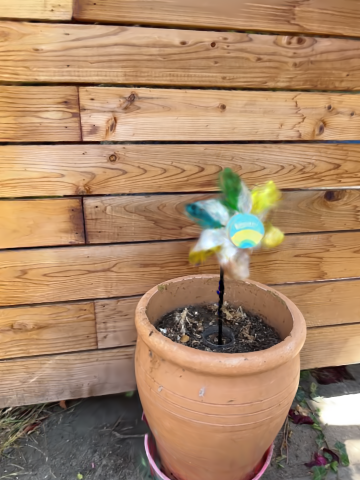}
            \includegraphics[width=\linewidth]{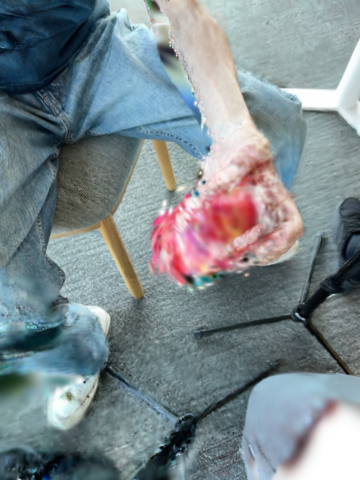}
                 \includegraphics[width=\linewidth]{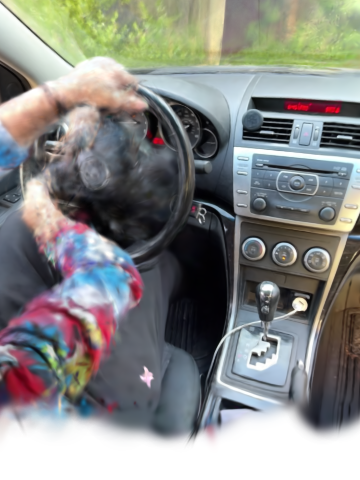}
            \caption*{GP-4DGS}
        \end{subfigure}
    \end{minipage}
    \vspace{-2mm}
    \caption{
    \mjg{Qualitative comparison of novel view synthesis on the DyCheck dataset. GP-4DGS shows more accurate geometry compared to baselines, particularly in regions with less observation. }}
    \label{fig:NVS_dycheck}
    \vspace{-3mm}
\end{figure}
Following the evaluation protocol of DyCheck~\cite{gao2022monocular}, we assess novel view synthesis quality using masked versions of standard metrics---mPSNR, mSSIM, and mLPIPS---to focus specifically on co-visible regions.
Table~\ref{table:quantitative_dycheck} presents that GP-4DGS consistently surpasses all baselines, with the performance gap in mPSNR and mLPIPS widening on the \textit{Challenging subset}\footnote{The challenging subset follows the evaluation protocol of~\cite{huang2025vivid4d}, where training and test viewpoint overlap is significantly reduced.}.
This demonstrates that our GP-based optimization effectively mitigates artifacts from sparse observations by propagating spatiotemporal correlations from confident regions to unobserved ones.
Qualitative results in Figure~\ref{fig:NVS_dycheck} confirm these gains; our method recovers sharper textures and more accurate geometry compared to Gaussian Marbles~\cite{stearns2024dynamic} and SoM~\cite{wang2025shape}, which often suffer from floater artifacts or geometric blurring under rapid motion.

\vspace{-2mm}
\paragraph{Extreme novel view synthesis on DAVIS}
\begin{figure}[t]
     \centering
     \includegraphics[width=\linewidth, trim={0 1mm 0 0}, clip]{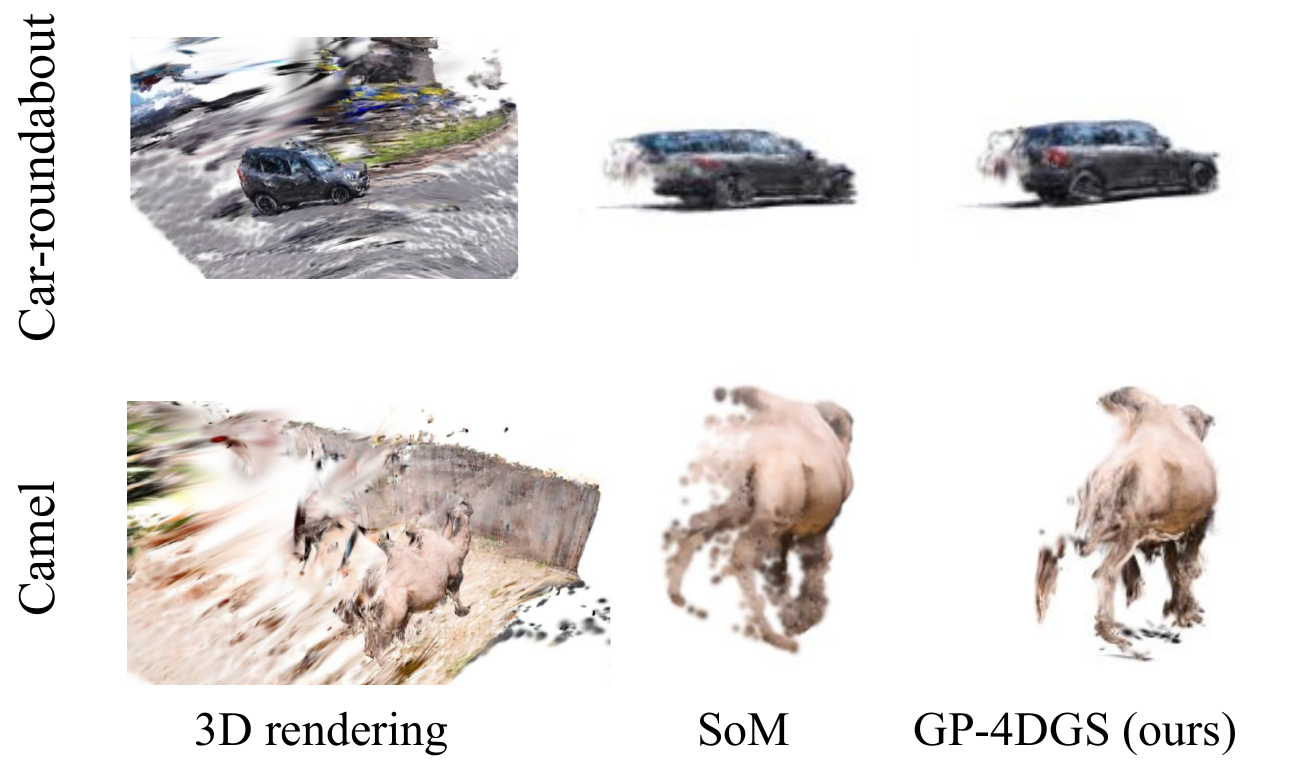}
     \vspace{-6mm}
     \caption{
     Qualitative comparison on the DAVIS dataset under extreme viewpoint shifts from training view. 
     Unlike the baseline, our spatiotemporal GP prior effectively regularizes the scene by faithfully propagating motion constraints into poorly observed regions.%
}
     \label{fig:davis}
     \vspace{-2mm}
\end{figure}

To assess robustness under extreme viewpoint shifts, we evaluate GP-4DGS on DAVIS  by rendering views significantly outside the training distribution. 
As illustrated in Figure~\ref{fig:davis}, our method preserves geometry and sharp edges more faithfully than the baseline~\cite{wang2025shape}, validating the reliability of GP-GS optimization in maintaining structural integrity, even in regions with sparse observations.

\subsection{Probabilistic Interpretability of 4DGS}

\subsubsection{Future Motion Extrapolation}
\begin{figure}[t]
    \centering
    \begin{subfigure}[b]{\linewidth}
        \centering
        \includegraphics[width=\linewidth]{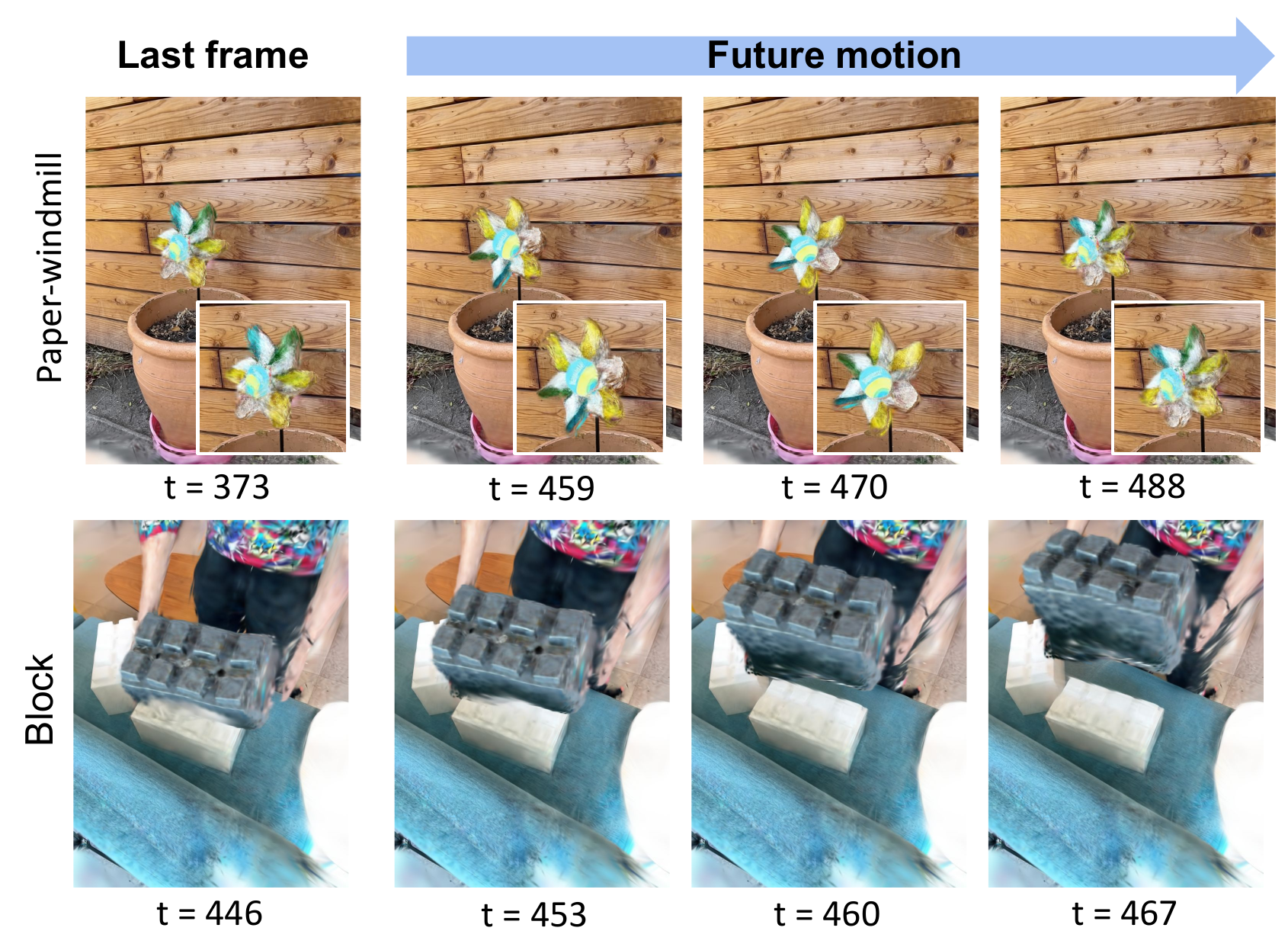}
    \end{subfigure}
    \vspace{-7mm}     
    \caption{
    Motion extrapolation results from GP-4DGS.
    Our GP-based approach naturally predicts future motion by querying the model at timesteps beyond the training range. }
    \label{fig:extrapolation}
    \vspace{-2mm}
\end{figure}

GP-4DGS enables motion extrapolation by querying the model at future timesteps. 
To demonstrate its effectiveness, we withhold the last 5 or 15 frames of each sequence during training for evaluation.
As shown in Table~\ref{tab:extrapolation}, our method significantly outperforms naive linear extrapolation, particularly in periodic scenes, where the temporal kernel captures cyclic dynamics. 
Figure~\ref{fig:extrapolation} further confirms that GP-4DGS predicts physically plausible and temporally coherent motion for these unobserved timesteps, demonstrating its ability to learn underlying dynamics rather than merely interpolating training frames.

\begin{table}[t]
        	\centering
	\caption{
	Future motion extrapolation performance in terms of PSNR ($\uparrow$). 
	We evaluate performance on the last 5 and 15 frames excluded for training. GP-4DGS outperforms na\"ive linear extrapolation, especially in periodic scenes where the temporal kernel effectively captures cyclic structures.
	}
	\vspace{-2mm}
	\scalebox{0.8}{
   	\setlength\tabcolsep{6pt} 
    	\begin{tabular}{l|cc|cc}
    	\toprule
	\multirow{2}{*}{{ Method}}  & \multicolumn{2}{c}{{Periodic motion}}  & \multicolumn{2}{c}{{Non-periodic motion}} \\
	& 5 frames   & 15 frames & 5 frames   & 15 frames\\
	\midrule 
	Linear extrapolation & 11.55 & \ \ 8.11 & 15.02 & 11.92\\ 
	GP-4DGS (ours)  & \textbf{17.62} & \textbf{16.65} & \textbf{15.27} & \textbf{13.22}\\ 
	\bottomrule
    	\end{tabular}
    	}
\vspace{-4mm}
\label{tab:extrapolation}
\end{table}

\subsubsection{Uncertainty Quantification}
GP-4DGS enables principled uncertainty quantification through its probabilistic formulation. 
To evaluate its reliability, we adopt the Area Under the Sparsification Error (AUSE), which measures the alignment between estimated uncertainty and actual reconstruction error. 
As shown in Table~\ref{tab:uncertainty}, GP-4DGS consistently outperforms both Random and UA-4DGS~\cite{kim20244d} baselines. 
Notably, the performance gap becomes larger when evaluating AUSE on high-quality frames (e.g., top 20 and 40 frames). 
This indicates that by accounting for both spatial and temporal correlations, our model effectively identifies subtle residual errors in well-reconstructed regions, where baselines fail to produce well-calibrated uncertainty.

\begin{table}[t]
        	\centering
	\caption{
	Uncertainty quantification results in terms of AUSE-MSE ($\downarrow$), measured as the area gap between the reconstruction error- and predicted uncertainty-based sparsification curves.
	Top 20 and 40 denote the frames with the lowest MSEs. GP-4DGS achieves the lowest AUSE across all settings.
    	}
    	\vspace{-2mm}
	\scalebox{0.8}{
   	\setlength\tabcolsep{5pt} 
    	\begin{tabular}{l|lllccc}
    	\toprule
	\multirow{2}{*}{{ Method }}  & \multicolumn{3}{c}{{AUSE-MSE~\cite{savant2024modeling} ($\times 10^{-2}$) $\downarrow$}} \\
	  & Top 20 frames   & Top 40 frames & All frames\\
	\midrule 
	Random   &  ~~~~~~9.76&  ~~~~~~9.30&  ~~~~10.98 \\
	UA-4DGS~\cite{kim20244d} 	& ~~~~~~7.60 &  ~~~~~~8.11 &  ~~~~~~8.62\\
	\hdashline 
	GP-4DGS (ours) & ~~~~~ \textbf{7.22}  \textcolor{blue}{(-0.38)}&  ~~~~~ \textbf{8.00}  \textcolor{blue}{(-0.11)}  &  ~~~~~ \textbf{8.49}  \textcolor{blue}{(-0.13)} \\
	\bottomrule
    	\end{tabular}
    	}
\vspace{-1mm}
\label{tab:uncertainty}
\end{table}

\subsection{Additional Analysis}

\paragraph{GP guidance as a motion prior}

\begin{figure}[t]
    \centering
    \includegraphics[width=\linewidth]{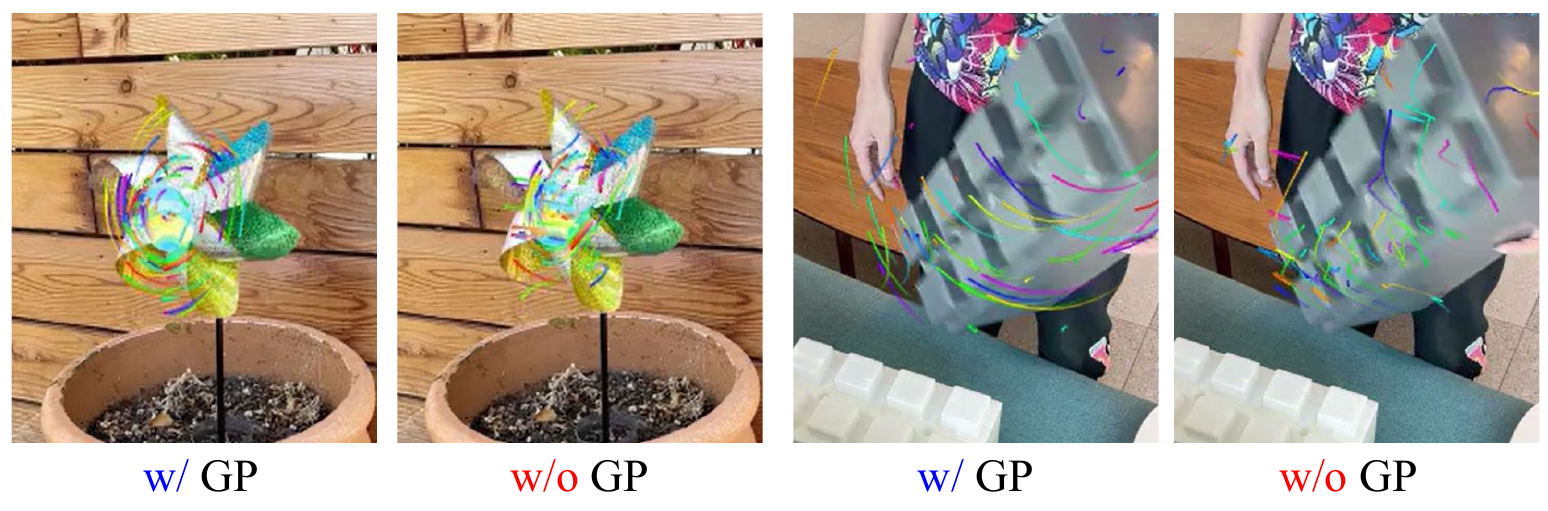}
       \vspace{-7mm}
    \caption{Trajectory comparison on the (left) \textit{paper-windmill} and (right) \textit{block} scene. GP guidance effectively regularizes motion trajectories, reducing noise and producing physically plausible motion patterns, compared to the baseline approach.}
    \label{fig:trajectory}
\end{figure}

\begin{figure}[t]
    \centering
    \begin{subfigure}[b]{0.49\linewidth}
        \centering
        \includegraphics[width=\linewidth]{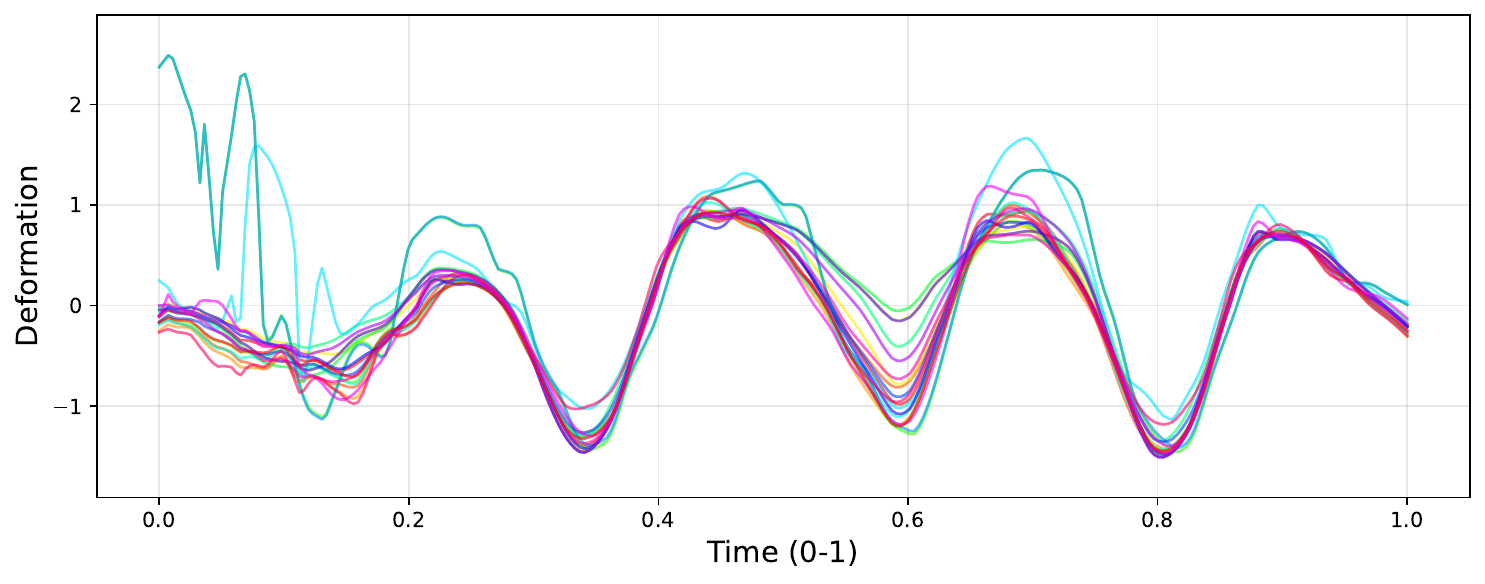}
        \caption{Trajectory w/o GP}
    \end{subfigure}
    \begin{subfigure}[b]{0.49\linewidth}
        \centering
        \includegraphics[width=\linewidth]{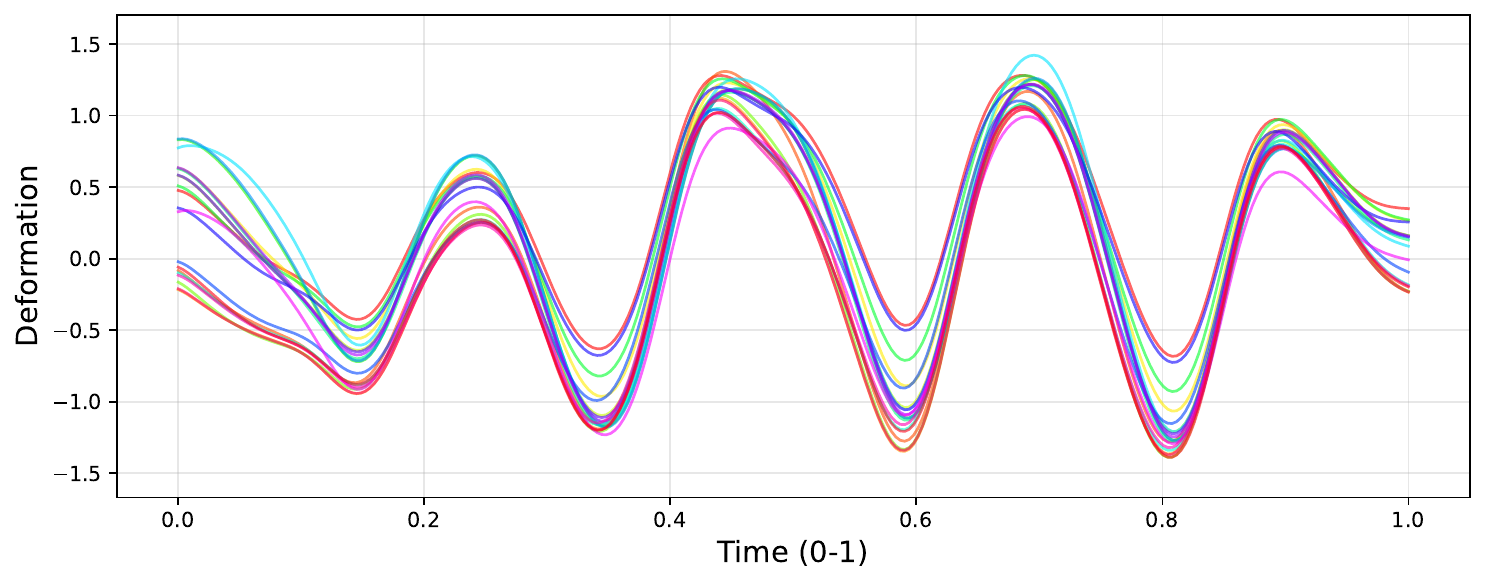}
        \caption{Trajectory from GP}
    \end{subfigure}
           \vspace{-2mm}
    \caption{Trajectory comparison on the \textit{spin} scene between the initial GS reconstruction and the GP guidance in the GP-GS optimization. These graphs correspond to the first dimension in 6D rotation.  
    The GP provides accurate and stable motion priors.}
    \label{fig:trajectory_graph}
    \vspace{-4mm}
\end{figure}
We present the regularizing effect of GP guidance on Gaussian trajectories in Figures~\ref{fig:trajectory} and~\ref{fig:trajectory_graph}. 
Without GP guidance, reconstructions suffer from noise and fluctuations, especially in sparsely observed regions. 
In contrast, GP-GS optimization leverages learned correlation structures to propagate motion priors from confident observations to uncertain regions. 
This process effectively stabilizes the trajectories, ensuring temporal smoothness and structural consistency throughout the sequence.

\begin{figure}[t]
	\vspace{-0.5mm}
    \centering
        \begin{subfigure}[b]{0.31\linewidth}
        \centering
        \includegraphics[width=\linewidth]{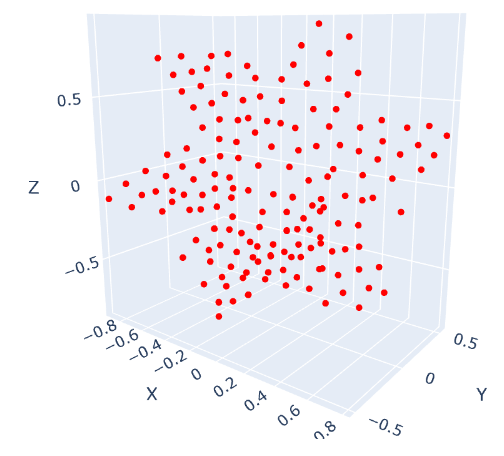}
        \caption{Canonical space}
        \vspace{-2mm}
    \end{subfigure}
            \begin{subfigure}[b]{0.68\linewidth}
        \centering
        \includegraphics[width=\linewidth, trim={0 -10mm 0 0}, clip]{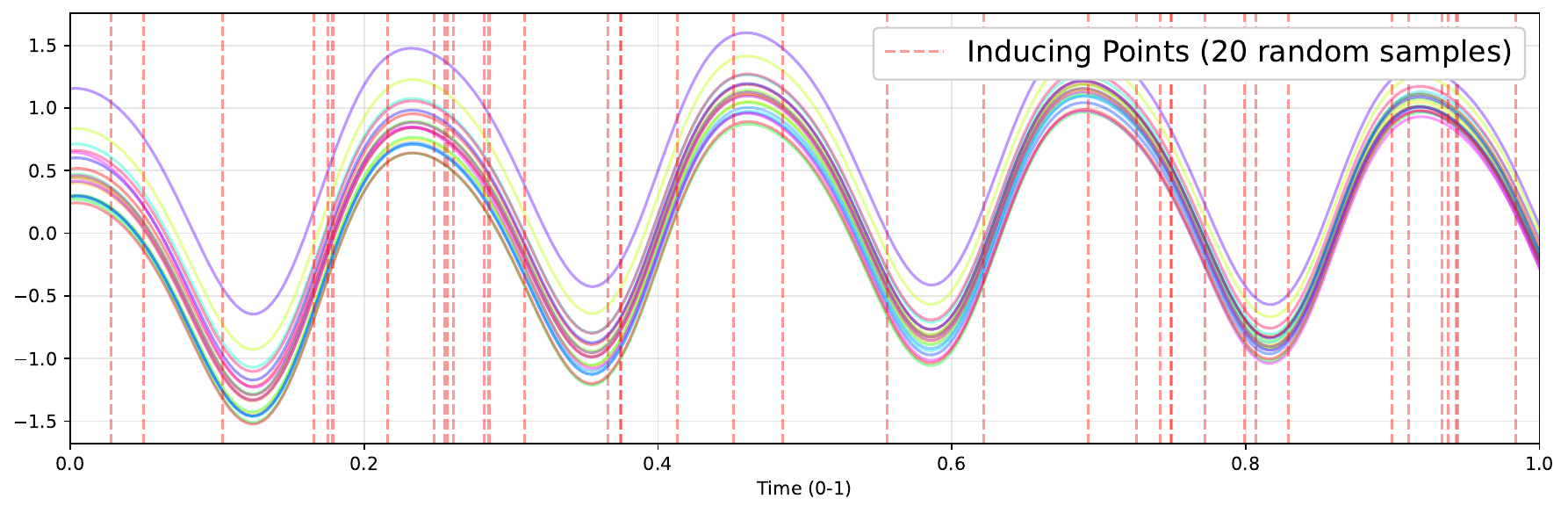}
           \caption{time axis in $[0. 1]$}
                   \vspace{-2mm}
    \end{subfigure}
    \caption{Visualization of inducing points (\textit{paper-windmill}). 
 {Inducing points (red) are well-distributed across the canonical space and temporal axis to ensure comprehensive coverage of the scene.}
   }
     \label{fig:inducing_points}
    \vspace{-1mm}
\end{figure}

\vspace{-2mm}
\paragraph{Inducing point coverage}
To show that the GP faithfully captures scene deformation, we visualize the inducing point distribution in Figure~\ref{fig:inducing_points}. 
The inducing points are well-distributed in both the canonical space and temporal axis, ensuring comprehensive coverage of motion dynamics over the full sequence. 
Such an arrangement prevents the GP from biasing toward specific clusters, enabling consistent and stable global deformation.

\vspace{-3mm}
\paragraph{Inducing point initialization}
We compare our time-series feature-based selection with random and velocity-based counterpart for inducing point initialization.
To isolate the impact of initialization, we evaluate ELBO under a single offline GP optimization without the full GP-GS loop. 
As shown in Table~\ref{tab:ablation_inducing_ELBO}, our method consistently achieves higher ELBO, demonstrating superior convergence and representation of motion dynamics. 
See Section~A of the supplementary document for detailed initialization strategies.%
%
%
\begin{table}[t]
\centering
\caption{{Comparison of inducing point initialization methods in terms of ELBO ($\uparrow$).
Our time-series-based selection achieves superior convergence with higher ELBO than baselines.}}\label{tab:ablation_inducing_ELBO}
\vspace{-1mm}
\scalebox{0.8}{
\setlength\tabcolsep{5pt} 
\begin{tabular}{lccc}
\toprule
{Scene}  & {Random Init.} & {Velocity KNN} & {Time-series (ours)} \\
\midrule
 \textit{paper-windmill }&0.85 &1.12 &\textbf{1.28}\\
 \textit{apple }&1.12& \textbf{1.21} &1.19\\
 \textit{spin }&0.92 &1.15 &\textbf{1.35}\\
 \textit{teddy }&1.45 &1.52 &\textbf{1.68}\\
 \textit{block }&1.08 &1.31 &\textbf{1.48}\\
 \textit{space-out }&0.78 &1.18 &\textbf{1.52}\\
 \textit{wheel }&1.35 &1.68 &\textbf{1.80}\\
\midrule
 \textit{Average }&1.10 &1.37 &\textbf{1.53}\\
\bottomrule
\end{tabular}
}
\vspace{-4mm}
\end{table}
%
%
%
%
%

\section{Conclusion}
\label{sec:conclusion}

We proposed a novel probabilistic framework that integrates variational Gaussian Processes into the 4D Gaussian Splatting pipeline. 
By leveraging GP-based motion priors, our method effectively capture complex spatiotemporal correlations and propagates learned dynamics to uncertain regions, resulting in physically plausible motion trajectories. 
Furthermore, our approach provides principled uncertainty quantification and enables reliable future motion prediction. 
We believe our framework represents a significant step toward integrating principled probabilistic modeling with high-fidelity neural graphics.

\paragraph{Acknowledgements}




This work was partly supported by 
the National Research Foundation of Korea (NRF) grant [RS-2022-NR070855, Trustworthy Artificial Intelligence]
and the Institute of Information \& communications Technology Planning \& Evaluation (IITP) grants 
[RS-2025-25442338, AI star Fellowship Support Program (Seoul National University);  
RS-2022-II220959 (No.2022-0-00959), (Part 2) Few-Shot Learning of Causal Inference in Vision and Language for Decision Making; 
No.RS-2021-II211343, Artificial Intelligence Graduate School Program (Seoul National University)] 
funded by the Korea government (MSIT).
We also thank Eduardo P\'{e}rez-Pellitero from Huawei Noah's Ark Lab for his valuable discussions and support.

{
    \small
    \bibliographystyle{ieeenat_fullname}
    \bibliography{references}
}


\clearpage
\setcounter{section}{0}
\setcounter{table}{0}
\setcounter{figure}{0}
\setcounter{equation}{0}
\setcounter{algorithm}{0}
\renewcommand\thesection{\Alph{section}}
\renewcommand\thetable{\Alph{table}}
\renewcommand\thefigure{\Alph{figure}}
\renewcommand\theequation{\Alph{equation}}
\renewcommand\thealgorithm{\Alph{algorithm}}

\section{Effect of Time-series Feature Extractor}

\paragraph{Motivation}
In Variational Gaussian Processes (VGPs), the initialization of inducing points $\mathcal{Z}$ is a key factor in summarizing the 4D deformation field. 
To represent complex motion from a large candidate set of trajectories, it is essential to select points that capture the most informative dynamics.

Standard approaches often use velocity differences as a heuristic for $k$-nearest neighbor (KNN) selection. 
While efficient, velocity-based features only reflect instantaneous change, failing to account for long-term temporal patterns or higher-order dependencies within a trajectory. 
To bridge this gap, we employ Chronos~\cite{ansari2024chronos}, a pre-trained time-series foundation model, to generate embeddings that encode the global temporal deformation of each Gaussian primitive.

\paragraph{Time-series feature extraction}
For a trajectory $\boldsymbol{\xi}_k \in \mathbb{R}^{T \times 3}$ of the $k$-th primitive, we extract features for each spatial dimension $i \in \{x, y, z\}$ independently. 
This preserves axis-specific semantics while capturing temporal dynamics within each dimension. 
Chronos produces a 256-dimensional embedding $\mathbf{f}_{k,i}$ per axis, which we concatenate into a single trajectory descriptor as follows:
\begin{align}
\mathbf{F_k} = [\mathbf{f}_{k,x} \parallel \mathbf{f}_{k,y} \parallel \mathbf{f}_{k,z}] \in \mathbb{R}^{768},
\end{align}
By using learned representations, $\mathbf{F}_k$ captures complex motion cues such as periodic patterns and non-linear shifts that simple velocity vectors overlook.

\paragraph{Inducing point selection}
We construct the final set of inducing points $\mathcal{Z}$ by combining these spatial features with temporal sampling. 
First, we apply KNN in the feature space $\mathbf{F}$ to identify $M_{\text{spatial}}$ representative trajectory landmarks. 
These are then combined with $M_{\text{time}}$ points sampled from the normalized time interval $[0, 1]$ via a Cartesian product:

\begin{align}
\mathcal{M} = \mathcal{M}_{\text{spatial}} \times \mathcal{M}_{\text{time}}, \quad |\mathcal{M}| = M_{\text{spatial}} \times M_{\text{time}}.
\end{align}
This strategy ensures that the selected inducing points are placed on semantically significant trajectories, leading to more accurate reconstruction and fast convergence in VGPs.

\section{Effect of Periodic Component in GPs}

The choice of mean function and kernel design is critical for enabling long-term extrapolation and short-term motion prediction for future frames. 
We systematically compare constant versus periodic components to understand their complementary roles in this selection.

\subsection{Periodic Mean}

Priors on the output means also can effect on extrapolation, where they have minimal effect on the observed data region, but their strength is manifested in unobserved regions as inductive bias.
The prior over inducing point values is defined as follows~\footnote{This generalizes the standard form $m_i(\mathbf{x}) =  0 $ commonly used in Variational Gaussian Processes by introducing a learned mean function}: 
\begin{align}
f_i(\mathbf{x}) \sim \mathcal{GP}(m_i(\mathbf{x}), k_i(\mathbf{x}, \mathbf{x}')).
\end{align}
where $m_i$ is the mean function and the most common choice in standard a constant mean function as follows:
\begin{align}
m_i (\mathbf{x}) = c_i,
\end{align}
where $c$ is a learnable mean offset.
To enable temporal extrapolation on periodic motions, a \textbf{periodic mean} serves as an effective solution:
\begin{align}
m_i (\mathbf{x})= c_i + A_i\sin\left(\frac{2\pi t}{T_i} + \phi_i\right),
\end{align}
where $\mathbf{x} = (\boldsymbol{p} , t)$,  $c$ is the mean offset, $A_i$ is the amplitude, $T_i$ is the period, and $\phi_i$ is the phase shift. 
All parameters are learned jointly during training along with parameters in kernel.

\subsection{Periodic Kernel}

The temporal component of the kernel is critical for capturing motion patterns over time. 
We also compared two designs for the temporal kernel structure: constant kernel and periodic kernel. 
The \textbf{constant kernel} employs a stationary Matérn kernel over spatial and temporal dimensions jointly as follows:
\begin{align}
k_{i}^{\text{temporal}}(\mathbf{x}, \mathbf{x}') = \sum_{j \in \{ x, y, z\}} k_{i,j}^{\text{matérn}}(( p_j, t), ( p'_j, t')),
\end{align}
The \textbf{periodic kernel} factorizes spatial and temporal correlations as follows;
\begin{align}
k_{i}^{\text{temporal}}(\mathbf{x}, \mathbf{x}') = \sum_{j \in \{x, y, z\}} k_{i,j}^{\text{matérn}}(p_j, p'_j) \cdot k_{i,j}^{\text{periodic}}(t, t'),
\end{align}
where $k^{\text{periodic}}(t, t')$ captures recurring temporal patterns.

\subsection{Long-term \& Short-term Extrapolation}

For time-wise extrapolation, we compare the effects of periodic mean and periodic kernel. 
As summarized in Table~\ref{tab:prior_kernel_ablation}, the periodic mean is effective for long-term extrapolation on periodic motions, while the periodic kernel provides more general performance for both periodic and non-periodic motions, enabling short-term extrapolation.
Constant kernels without periodic structure fail entirely, halting predictions at the trajectory boundary

\begin{table}[t]
\centering
\caption{Comparison prior and kernel design for extrapolation.}
\label{tab:prior_kernel_ablation}
\scalebox{0.8}{
\setlength\tabcolsep{3pt} 
\begin{tabular}{ccccl}
\toprule
\textbf{Prior} & \textbf{Kernel} & \textbf{Long-term} & \textbf{Short-term} & \textbf{Recommendation} \\
\midrule
Constant & Constant &  &  & No extrapolation \\
Constant & Periodic &  & $\checkmark$ & Non-periodic motion \\
Periodic & Constant & Limited &  & Artifacts on non-periodic \\
Periodic & Periodic & $\checkmark$ & $\checkmark$ & Periodic motion \\
\bottomrule
\end{tabular}
}
\end{table}

\begin{figure*}[t]
     \centering
     \vspace{-1mm}
     \includegraphics[width=0.99\linewidth]{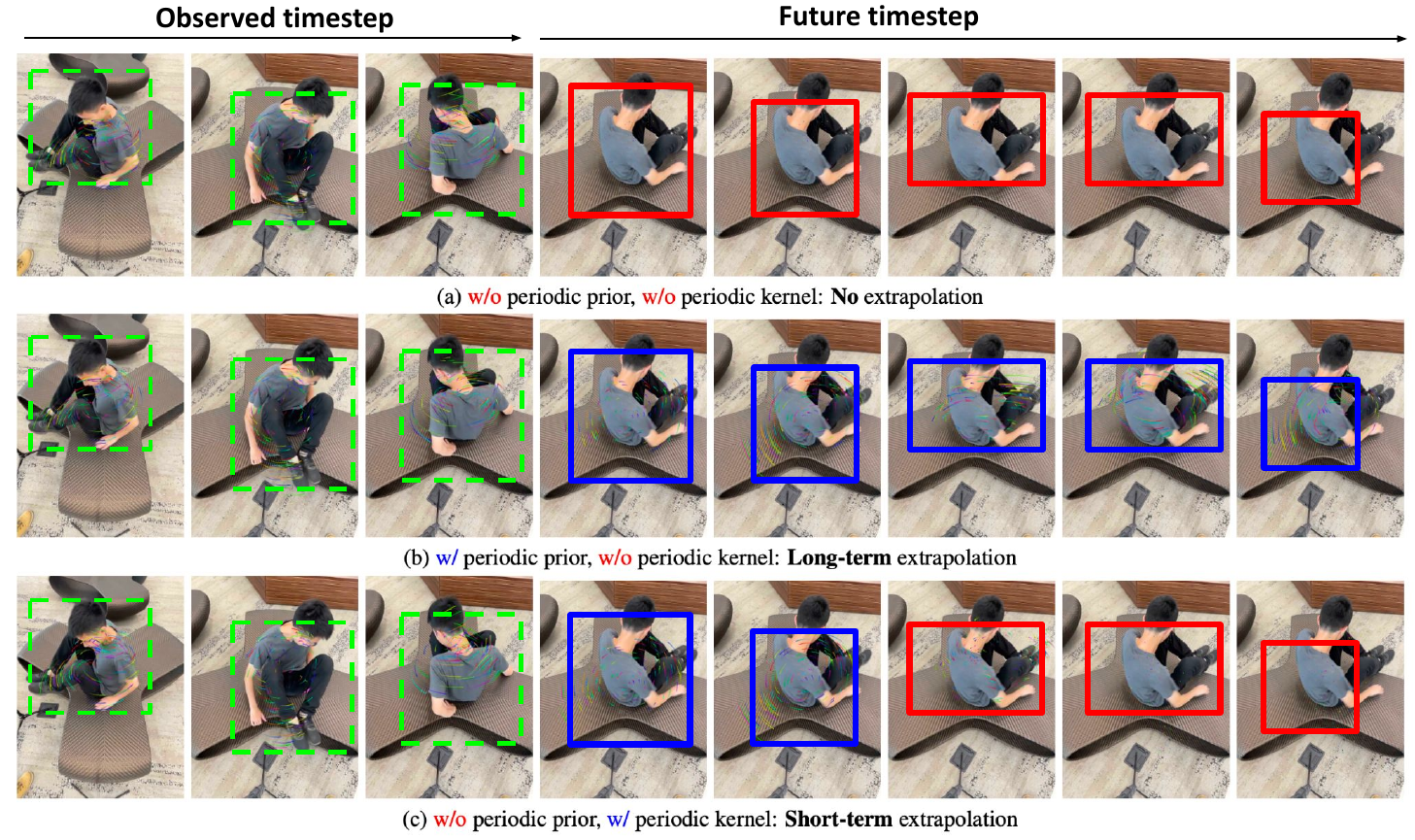}
     \vspace{-2mm}
  \caption{\textbf{Ablation of periodic component for extrapolation} with \textbf{\textcolor{blue}{predicted motion trajectories via GPs}}. 
Periodic priors enable long-term extrapolation, while periodic kernels enable short-term extrapolation. 
The first three columns show the seen frame period, columns four through five show short-term motion estimation, and columns six through eight show long-term motion estimation. 
The background for future frames is set to the last frame of the training sequence.}
     \label{fig:periodic_motion}
     \vspace{-2mm}
\end{figure*}

\begin{figure}[t]
     \centering
    \begin{subfigure}[b]{0.49\linewidth}
    \centering
    \includegraphics[width=0.49\linewidth]{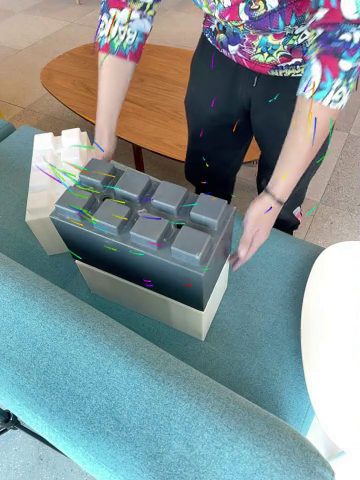}    \includegraphics[width=0.49\linewidth]{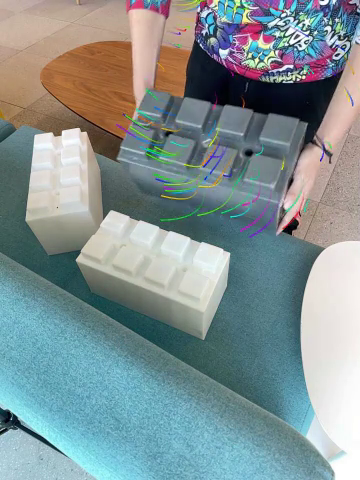}
    \caption{local-fluctuation}
    \vspace{-2mm}
    \end{subfigure}
    \begin{subfigure}[b]{0.49\linewidth}
    \centering
    \includegraphics[width=0.49\linewidth]{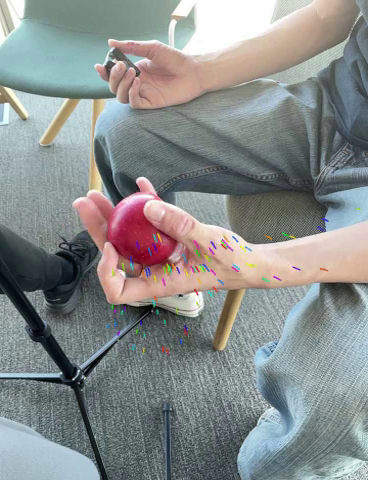}
    \includegraphics[width=0.49\linewidth]{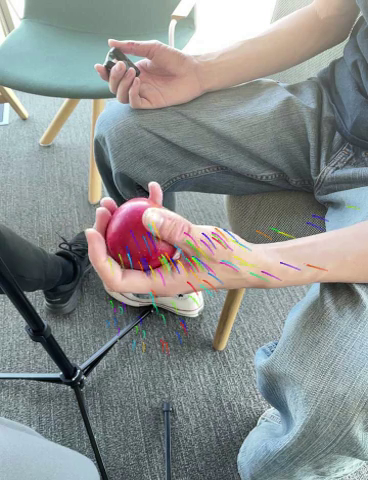}
    \caption{short-term extrapolation}
    \vspace{-2mm}
    \end{subfigure}
     \caption{Examples of local fluctuation in non-periodic motion with periodic prior. 
     Thus, we do not use a periodic prior for non-periodic motion and a periodic kernel is enough to predict for short-term extrapolation.
     }
     \label{fig:non-periodic_motion}
     \vspace{-4mm}
\end{figure}

Figure~\ref{fig:periodic_motion} demonstrates the effectiveness of periodic components on periodic motions.
Without any periodic structure, GPs fail to extrapolate beyond the observed time range. 
The periodic mean significantly improves long-term extrapolation, capturing the cyclical patterns well into the future.
For non-periodic motions, Figure~\ref{fig:non-periodic_motion} shows the periodic mean introduces spurious local oscillations and underfits the observed data on non-periodic sequences.

\begin{figure}[t]
    \centering
    \begin{subfigure}[b]{0.493\linewidth}
        \centering
        \includegraphics[width=\linewidth]{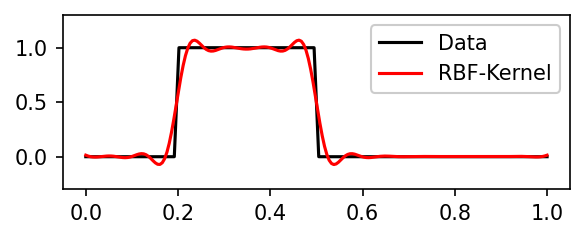}
    \end{subfigure}
    \begin{subfigure}[b]{0.493\linewidth}
        \centering
        \includegraphics[width=\linewidth]{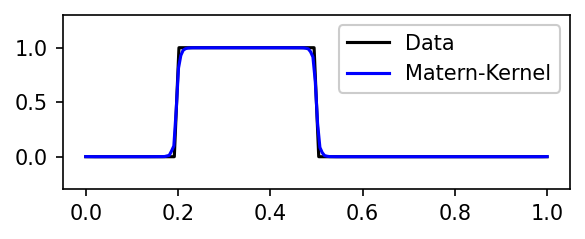}
    \end{subfigure}
    \vspace{-6.5mm}
   \caption{Kernel comparison on discontinuous data. RBF kernel (left) fails, but Mat\'{e}rn kernel (right) accurately fits step functions. }
          \label{fig:kernel_comparison}
    \vspace{-2mm}
\end{figure}

\section{Additional Analysis and Comparison}

\paragraph{Visualization of GP-GS trajectory}

To supplement Figure~\cvprb{5} in the main paper, we provide detailed visualizations of Gaussian primitive's trajectories across all dimensions (3D translation and 6D rotation representation) to show the impact of our GP algorithm, in Figure~\cvprb{C}.
This demonstrates the quality of our variational Gaussian process predictions for both translation (X, Y, Z coordinates) and rotation components.
Figure~\ref{fig:trajectory_comparison_block} shows trajectories with non-periodic motions, demonstrating that our method also handles these challenging scenarios effectively.

\begin{figure}[h]
\centering
\label{fig:trajectory_comparison_spin}
\begin{subfigure}[b]{0.49\linewidth}
\centering
\includegraphics[width=0.99\linewidth]{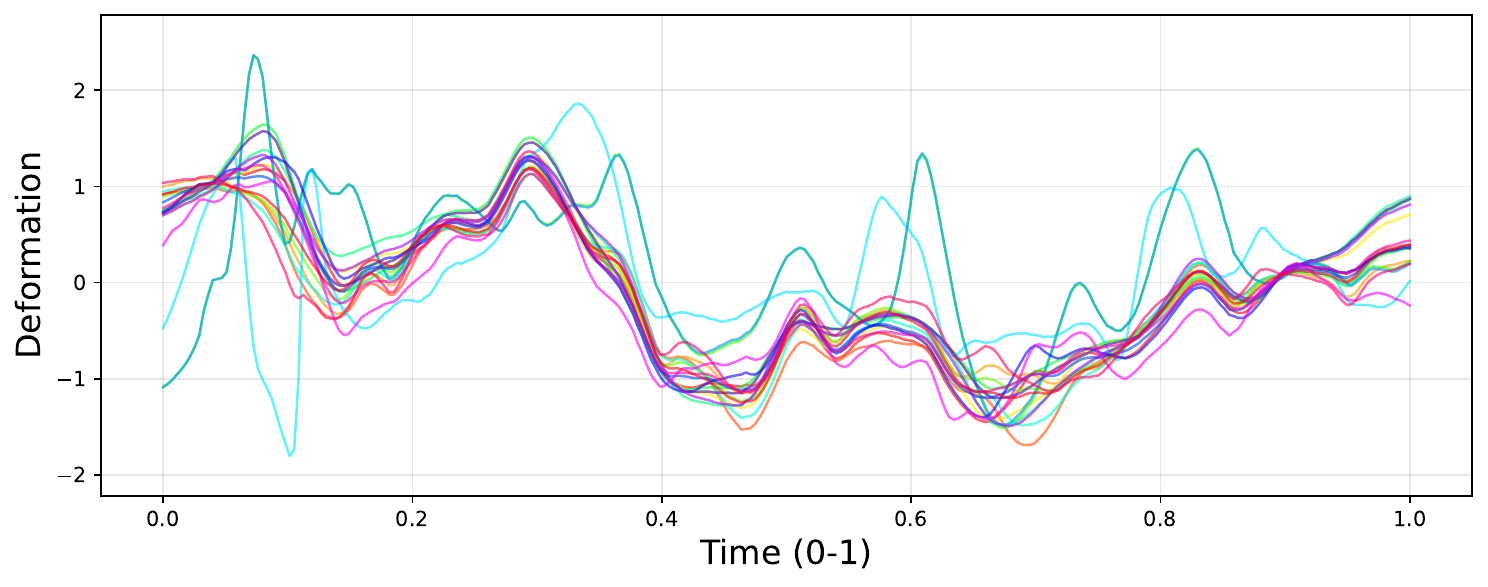}
\includegraphics[width=0.99\linewidth]{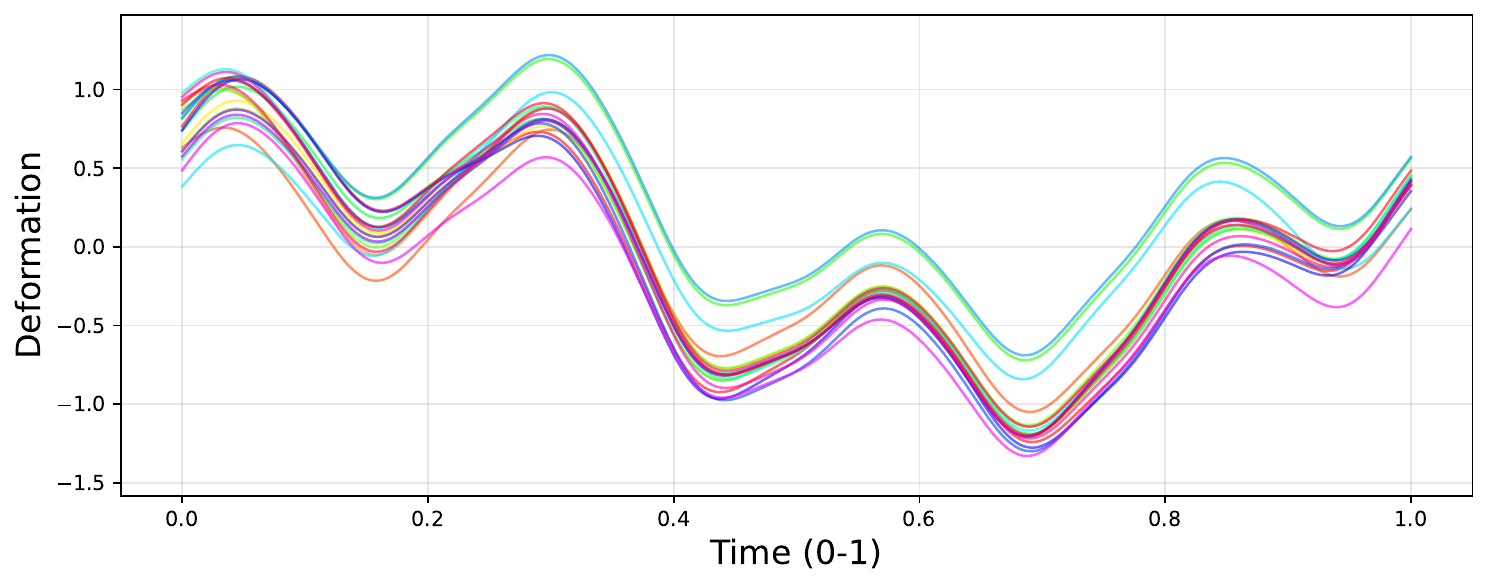}
\caption{X Translation}
\vspace{1mm}
\end{subfigure}
\begin{subfigure}[b]{0.49\linewidth}
\centering
\includegraphics[width=0.99\linewidth]{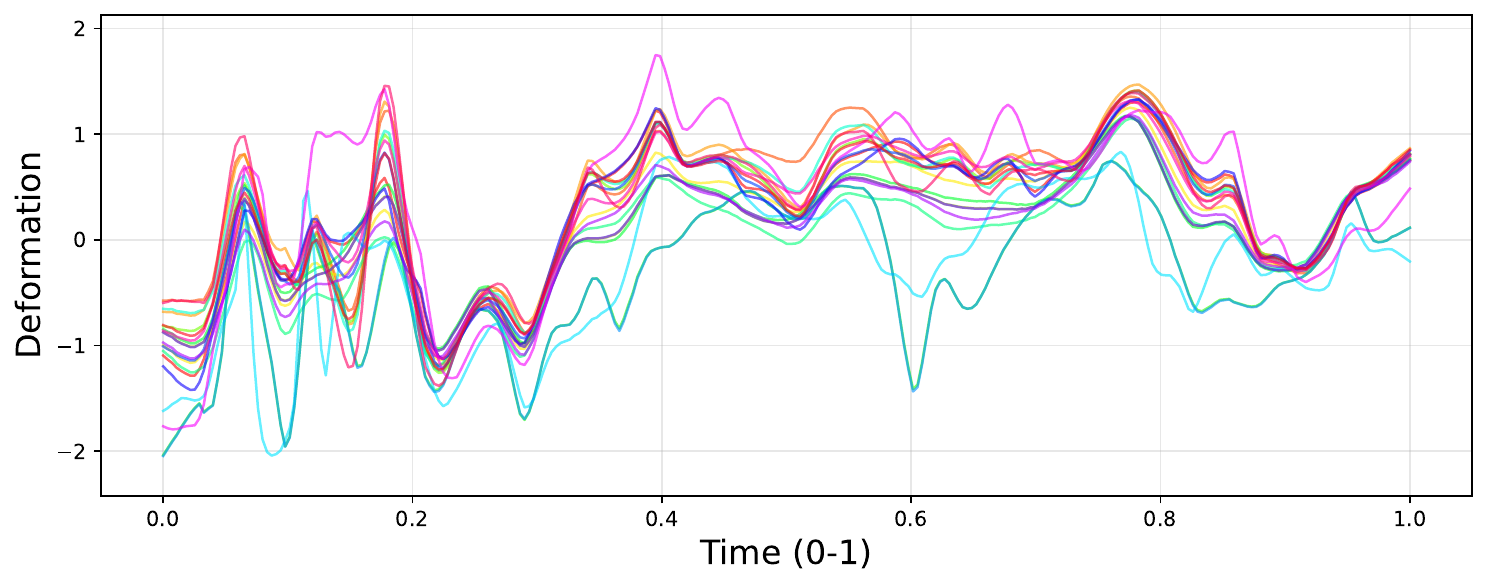}
\includegraphics[width=0.99\linewidth]{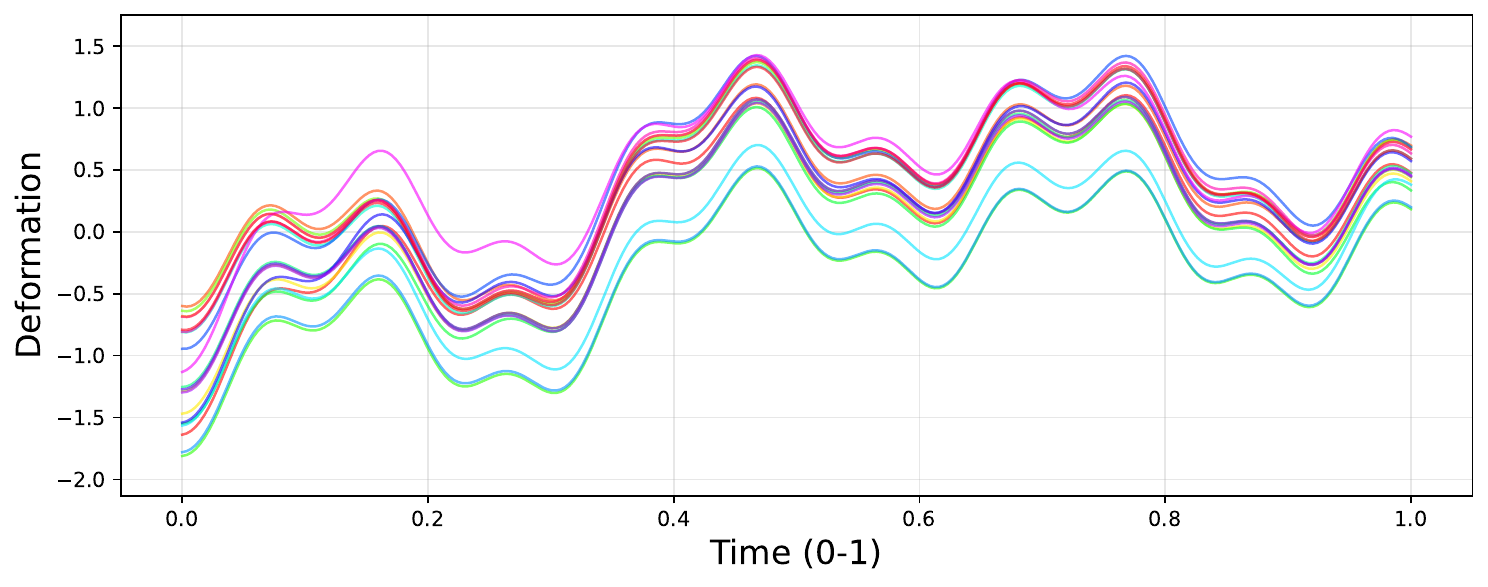}
\caption{Y Translation}
\vspace{1mm}
\end{subfigure}
\begin{subfigure}[b]{0.49\linewidth}
\centering
\includegraphics[width=0.99\linewidth]{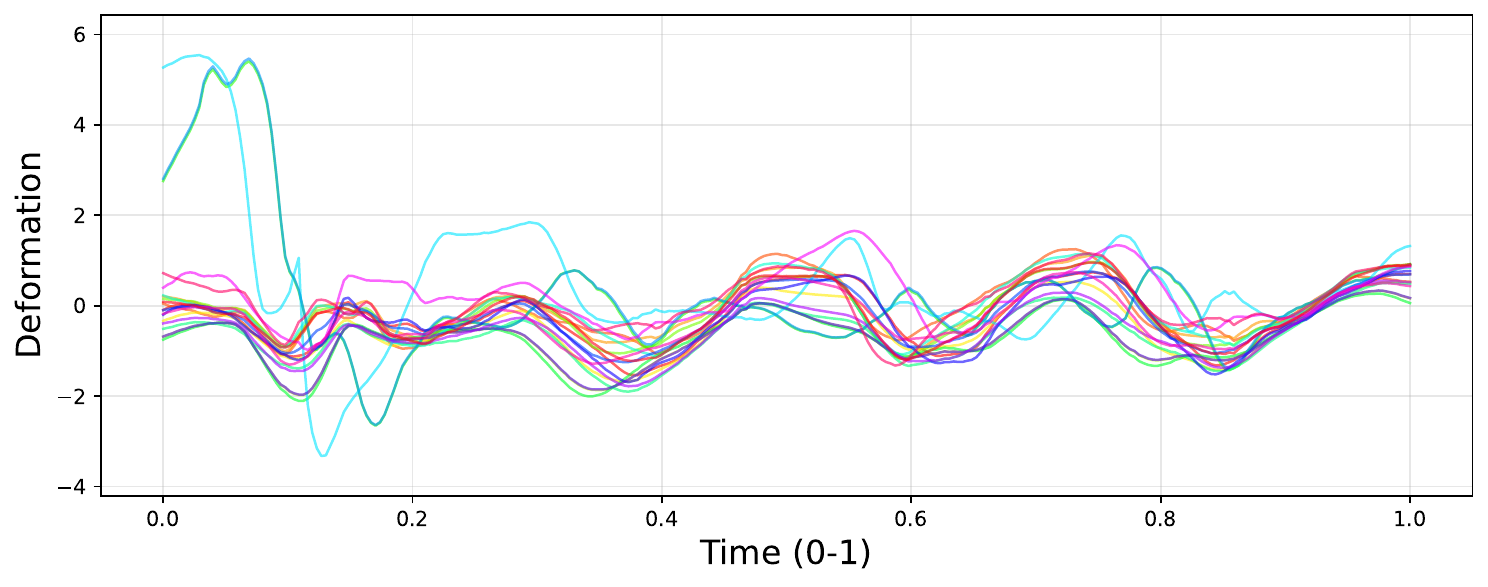}
\includegraphics[width=0.99\linewidth]{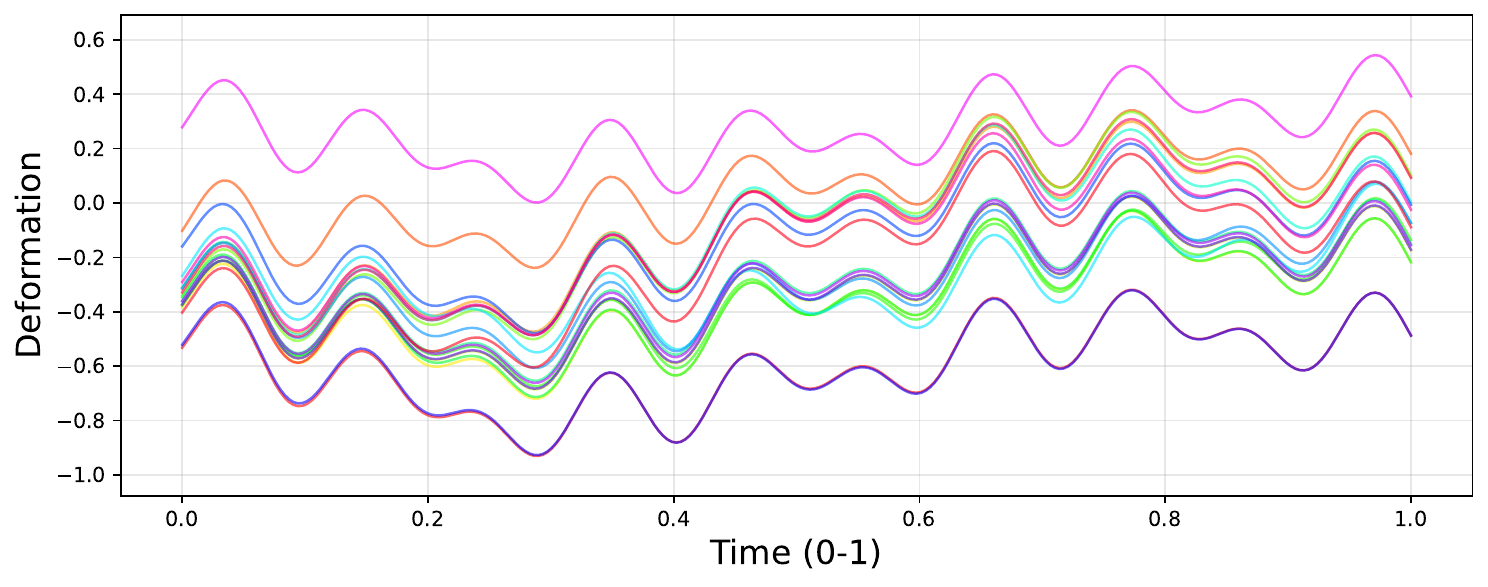}
\caption{Z Translation}
\vspace{1mm}
\end{subfigure}
\begin{subfigure}[b]{0.49\linewidth}
\centering
\includegraphics[width=0.99\linewidth]{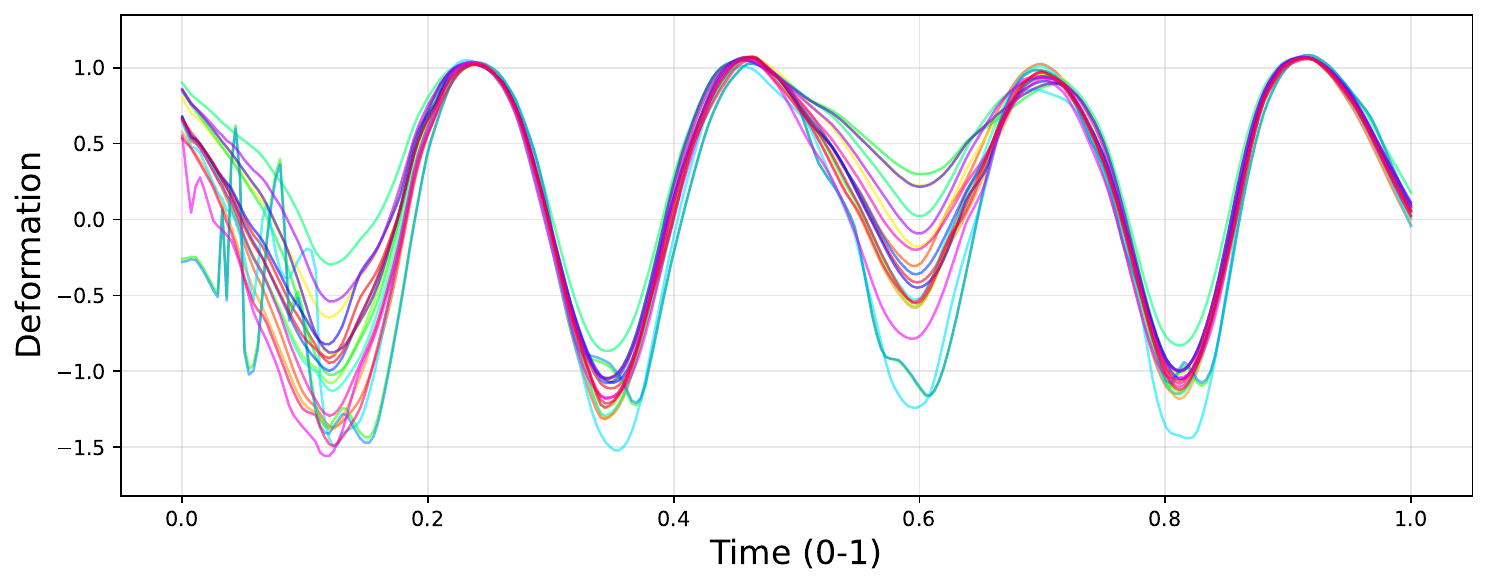}
\includegraphics[width=0.99\linewidth]{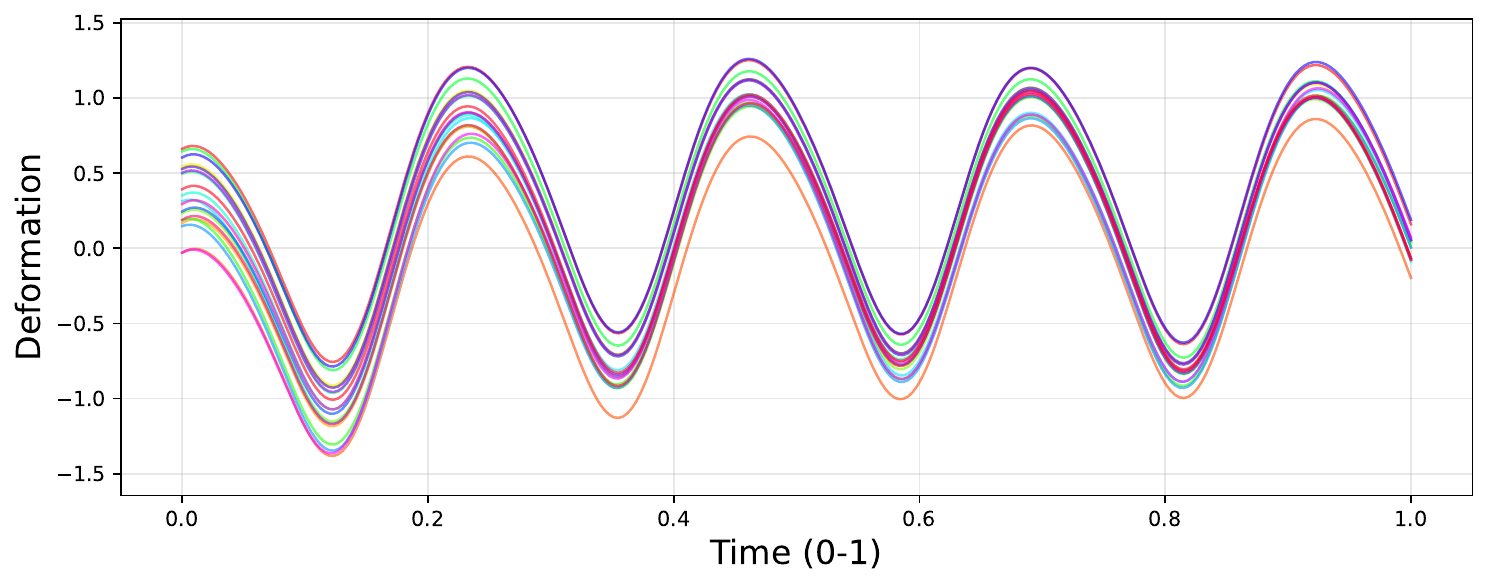}
\caption{6D Rotation Dim. 1}
\vspace{1mm}
\end{subfigure}
\begin{subfigure}[b]{0.49\linewidth}
\centering
\includegraphics[width=0.99\linewidth]{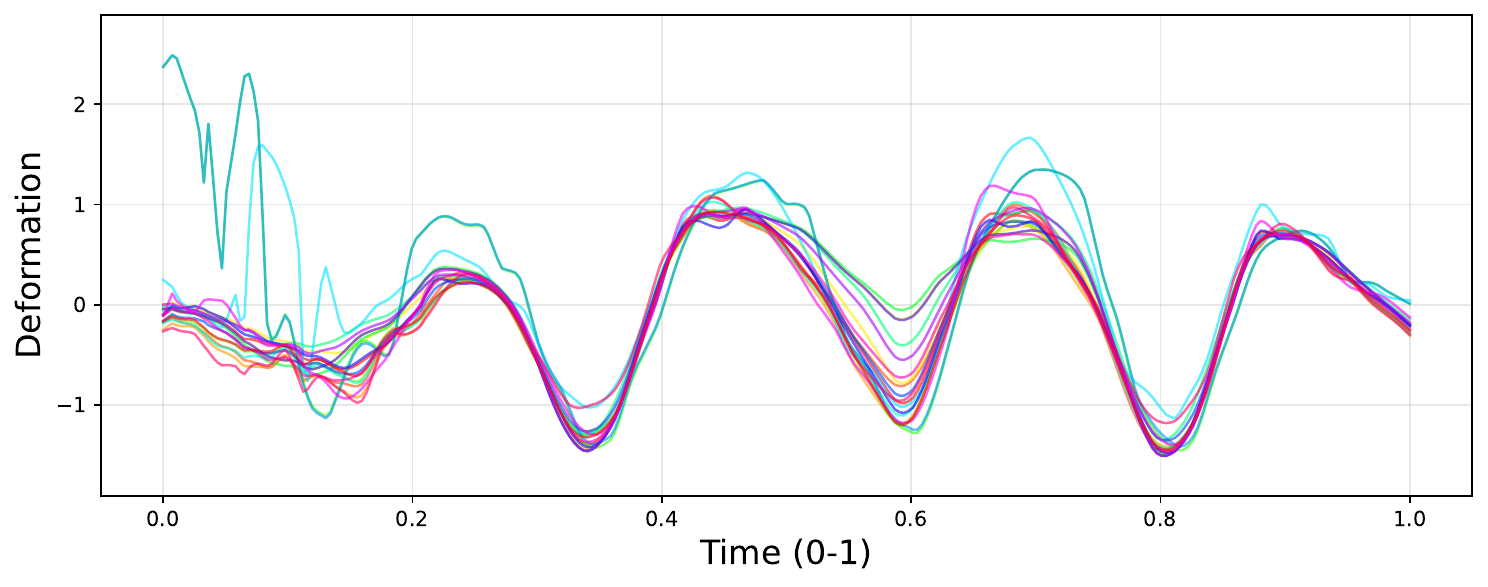}
\includegraphics[width=0.99\linewidth]{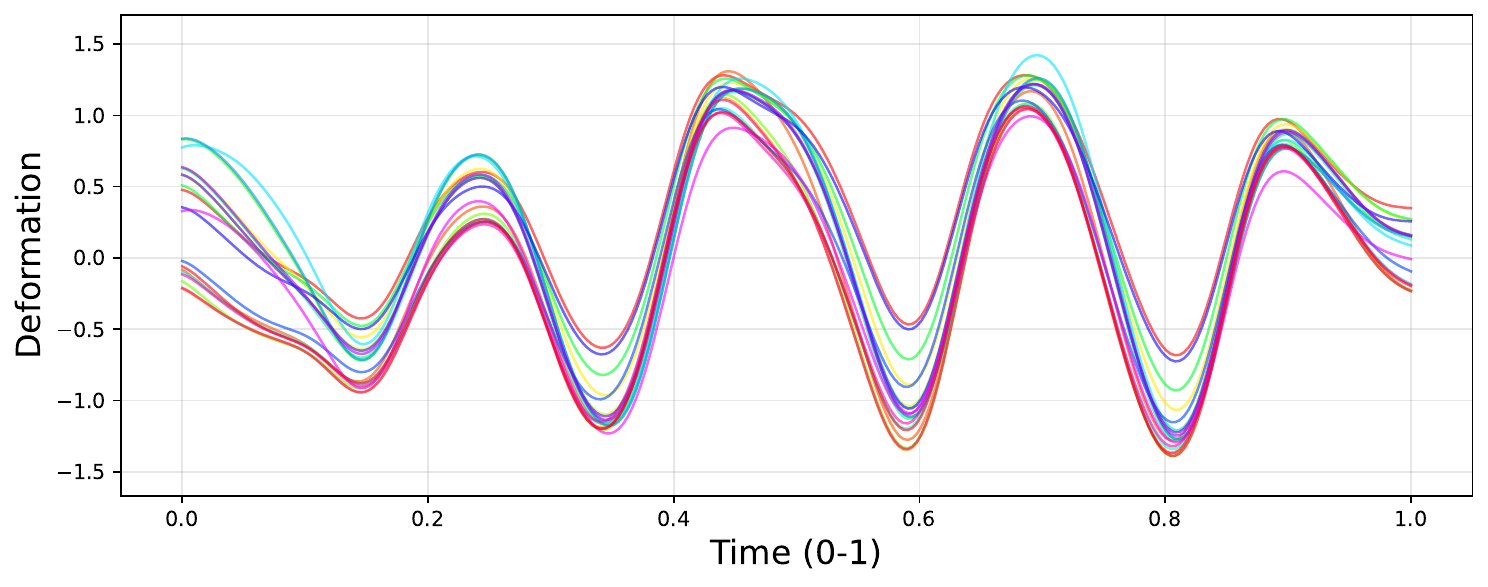}
\caption{6D Rotation Dim. 2}
\vspace{1mm}
\end{subfigure}
\begin{subfigure}[b]{0.49\linewidth}
\centering
\includegraphics[width=0.99\linewidth]{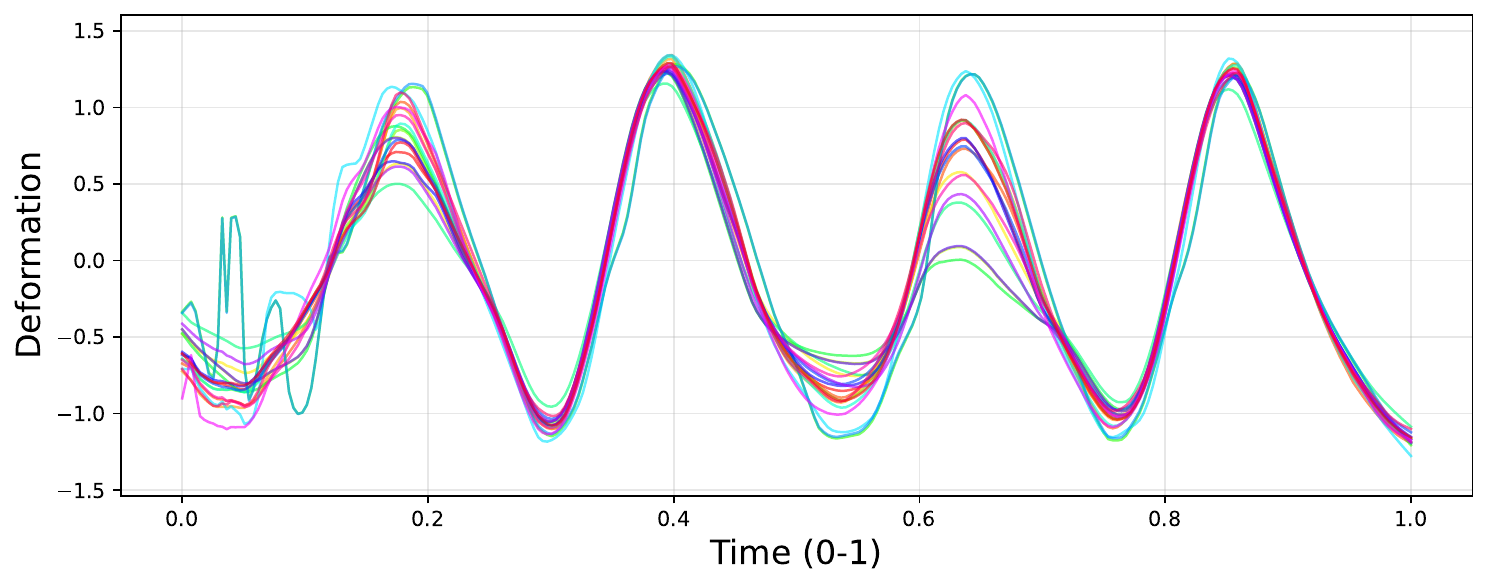}
\includegraphics[width=0.99\linewidth]{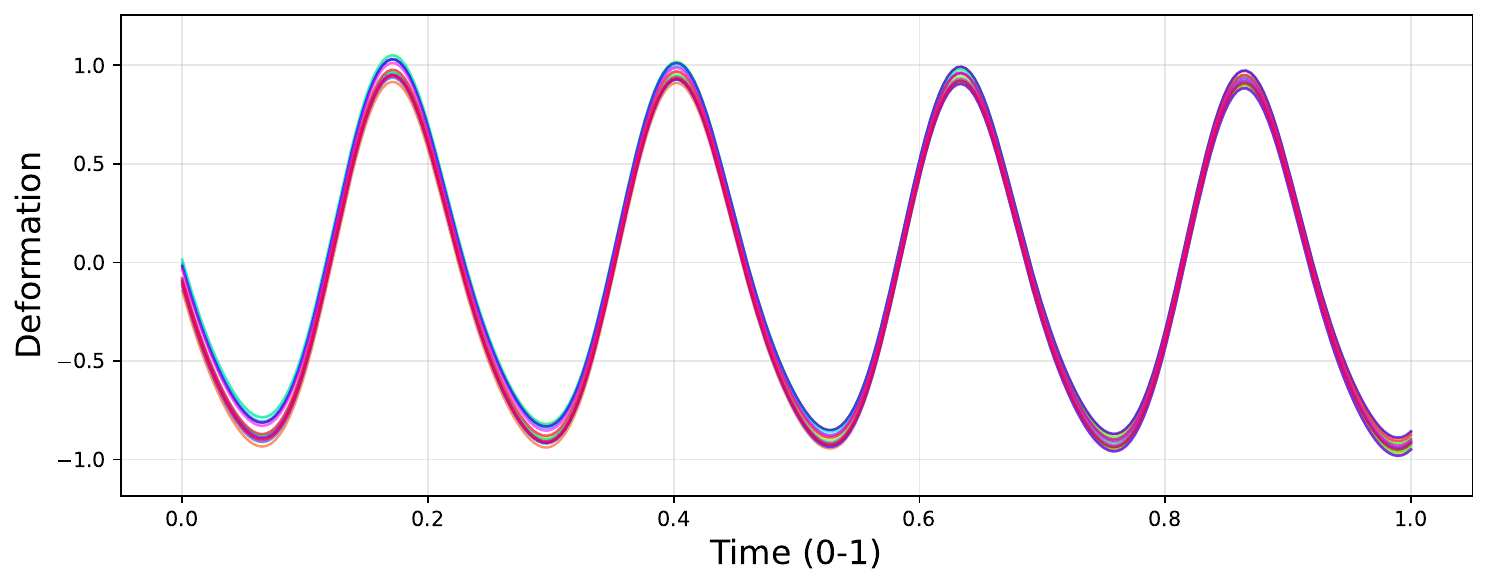}
\caption{6D Rotation Dim. 3}
\vspace{1mm}
\end{subfigure}
\begin{subfigure}[b]{0.49\linewidth}
\centering
\includegraphics[width=0.99\linewidth]{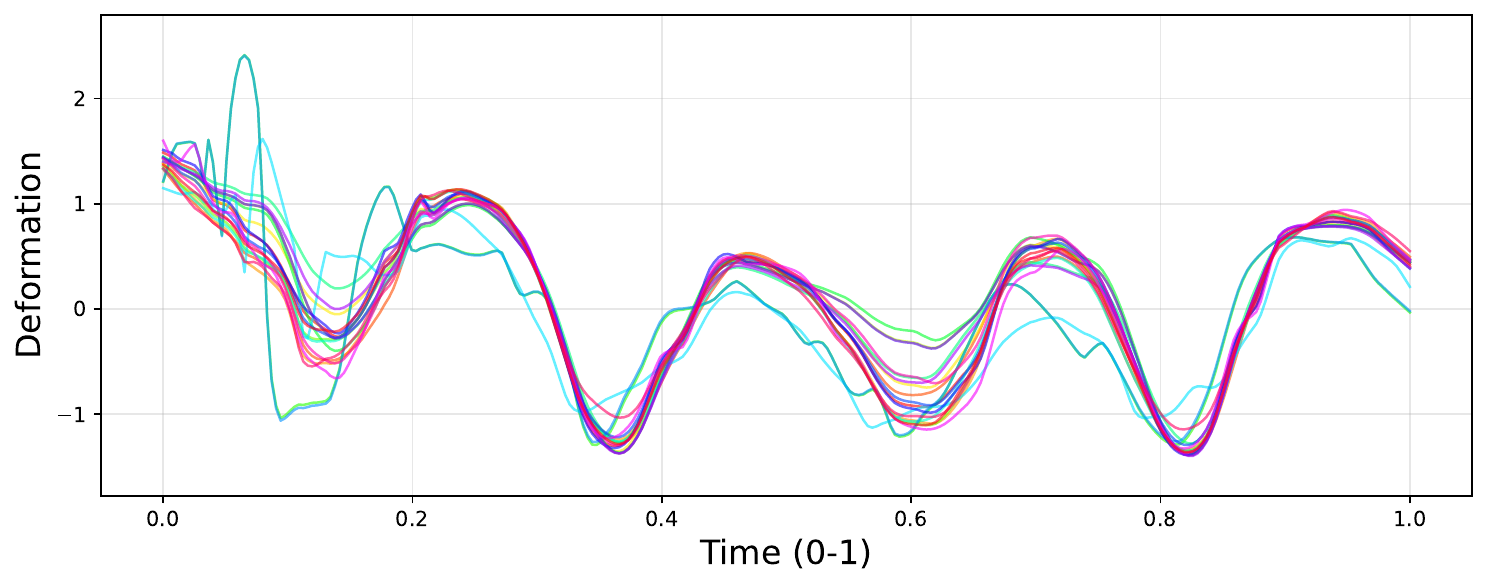}
\includegraphics[width=0.99\linewidth]{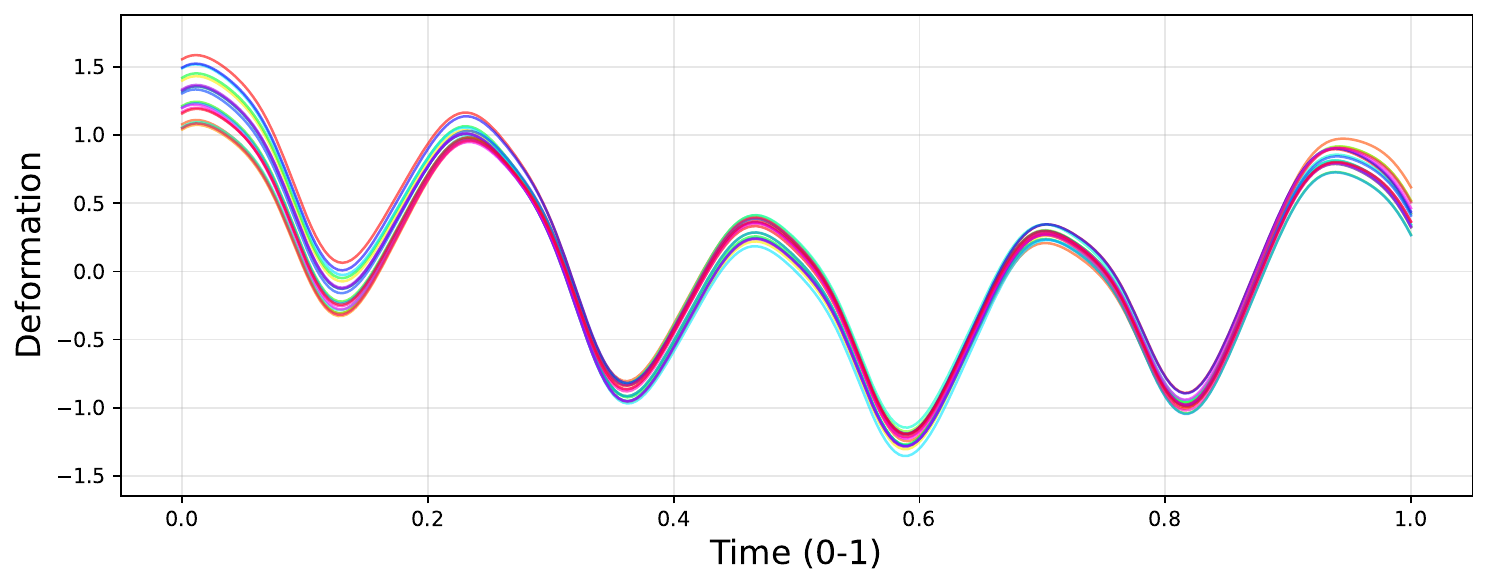}
\caption{6D Rotation Dim. 4}
\vspace{1mm}
\end{subfigure}
\begin{subfigure}[b]{0.49\linewidth}
\centering
\includegraphics[width=0.99\linewidth]{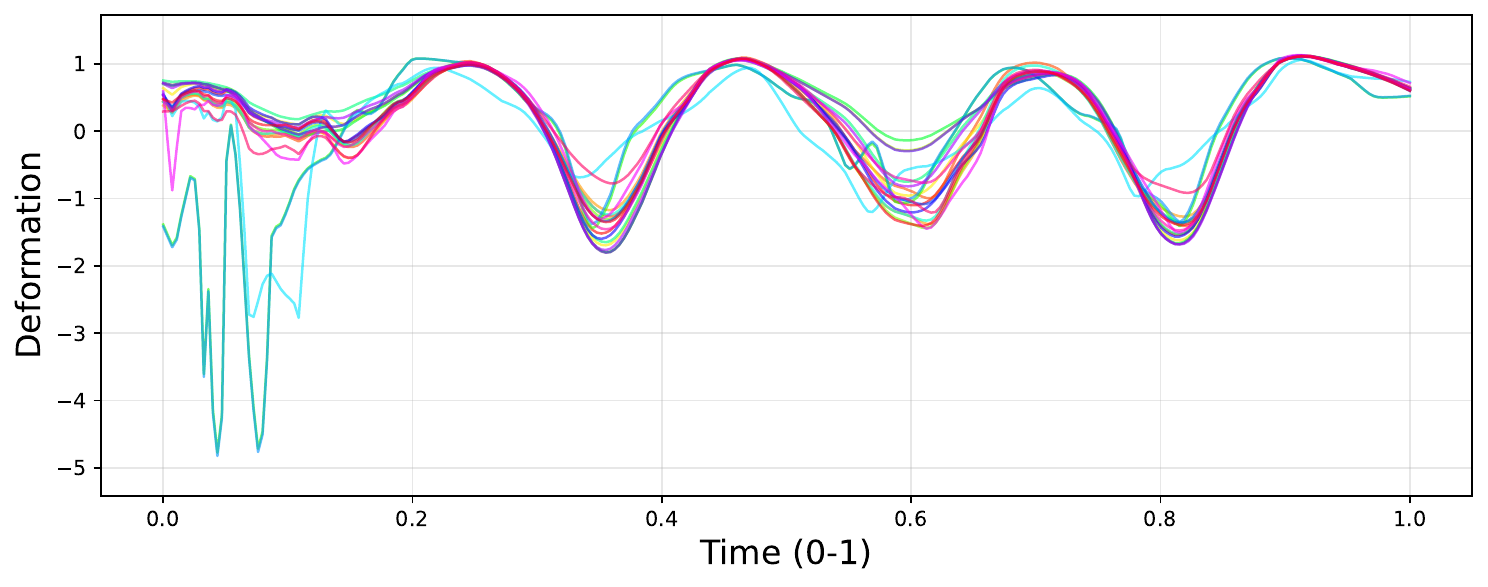}
\includegraphics[width=0.99\linewidth]{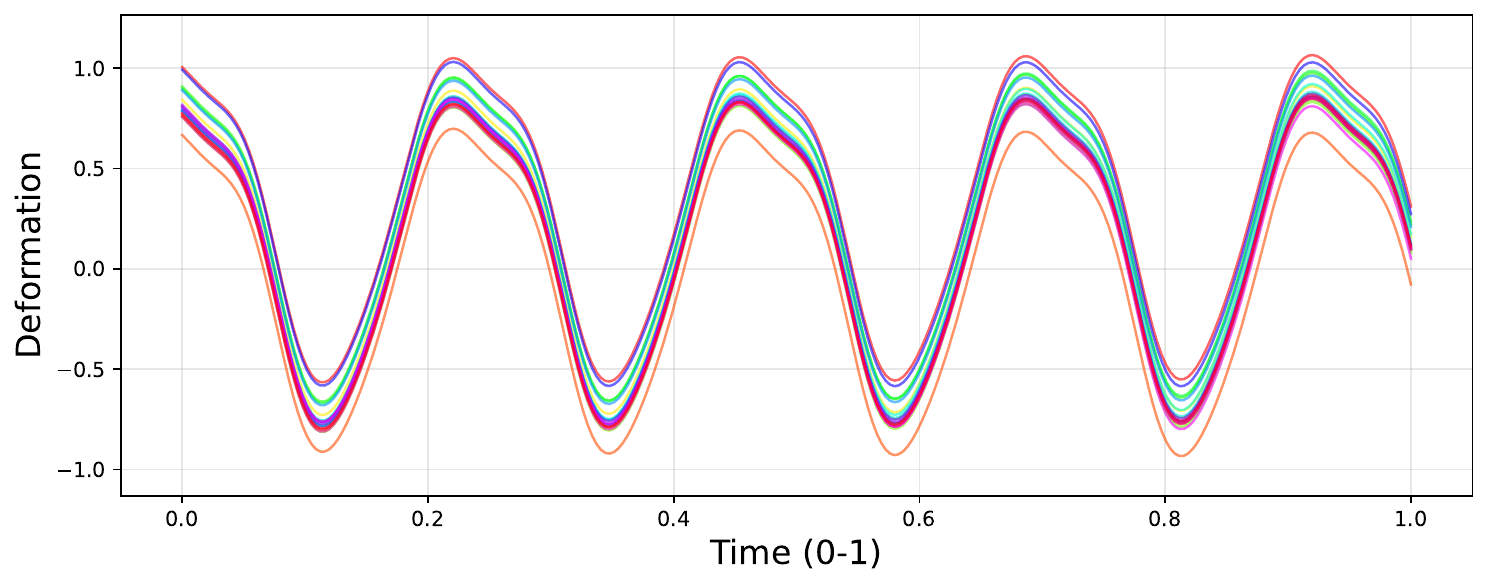}
\caption{6D Rotation Dim. 5}
\vspace{1mm}
\end{subfigure}
\begin{subfigure}[b]{0.49\linewidth}
\centering
\includegraphics[width=0.99\linewidth]{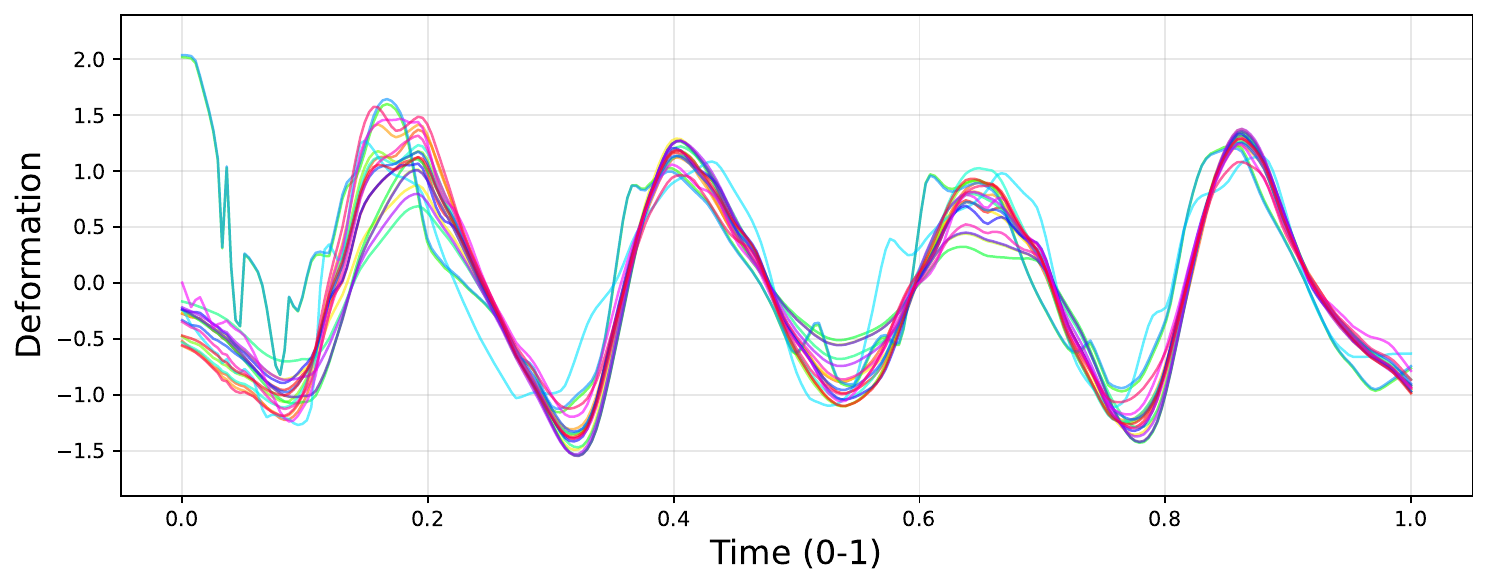}
\includegraphics[width=0.99\linewidth]{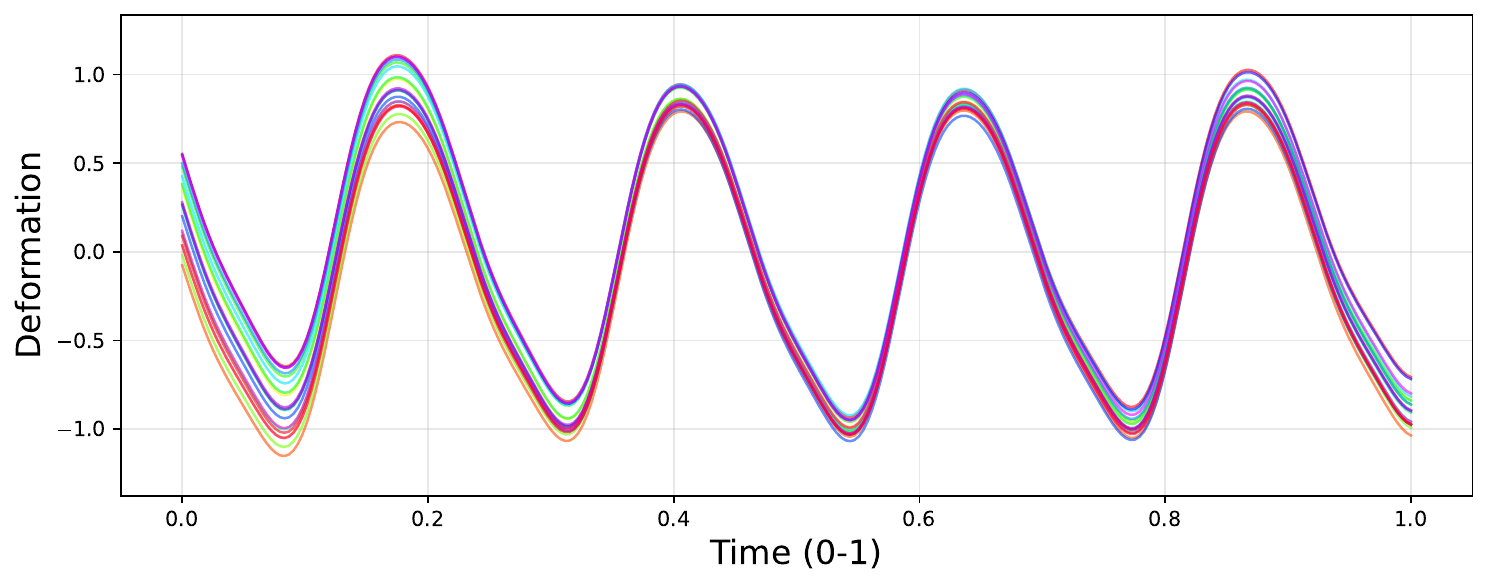}
\caption{6D Rotation Dim. 6}
\vspace{1mm}
\end{subfigure}
\vspace{-2mm}
\label{fig:trajectory_comparison_spin}
\caption{Comprehensive trajectory comparison across all transformation dimensions in the \textit{spin} scene: \textbf{Original (top) vs. GP (bottom)}. Our variational Gaussian process approach successfully captures and denoises both translational and rotational dynamics, properly giving the priors on reconstructing trajectories.}
\vspace{-4mm}
\end{figure}

\begin{figure}[h]
\centering
\label{fig:trajectory_comparison_block}
\begin{subfigure}[b]{0.49\linewidth}
\centering
\includegraphics[width=0.99\linewidth]{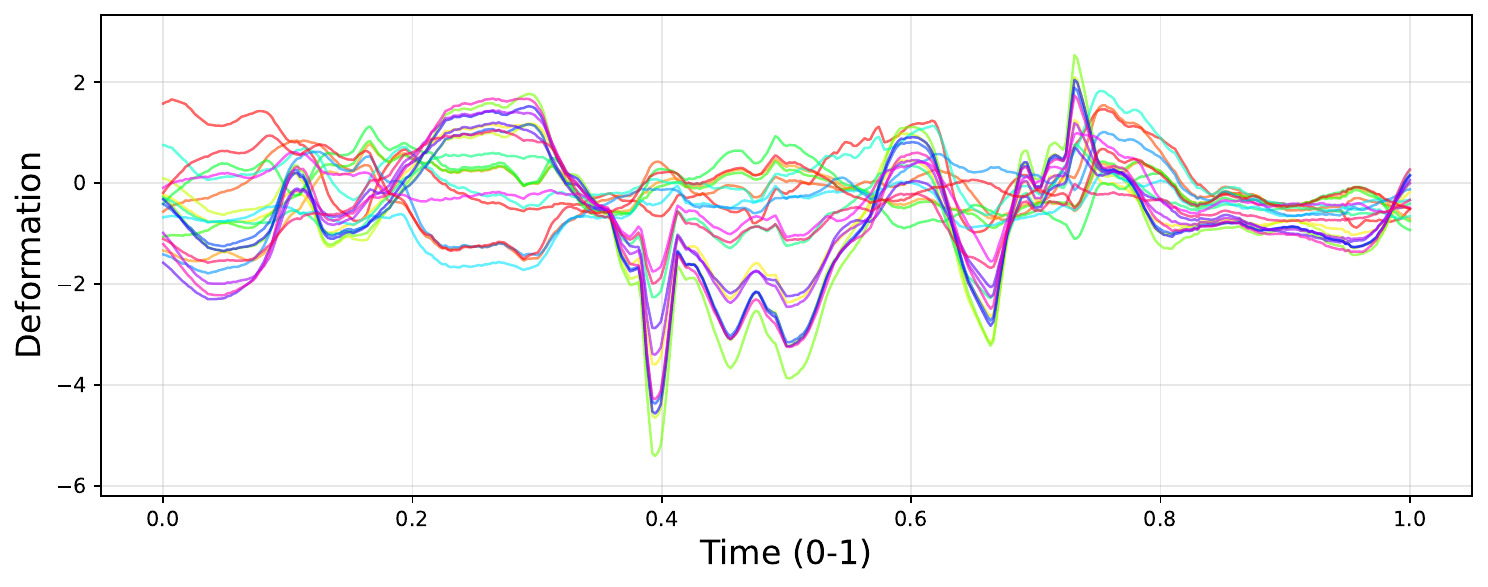}
\includegraphics[width=0.99\linewidth]{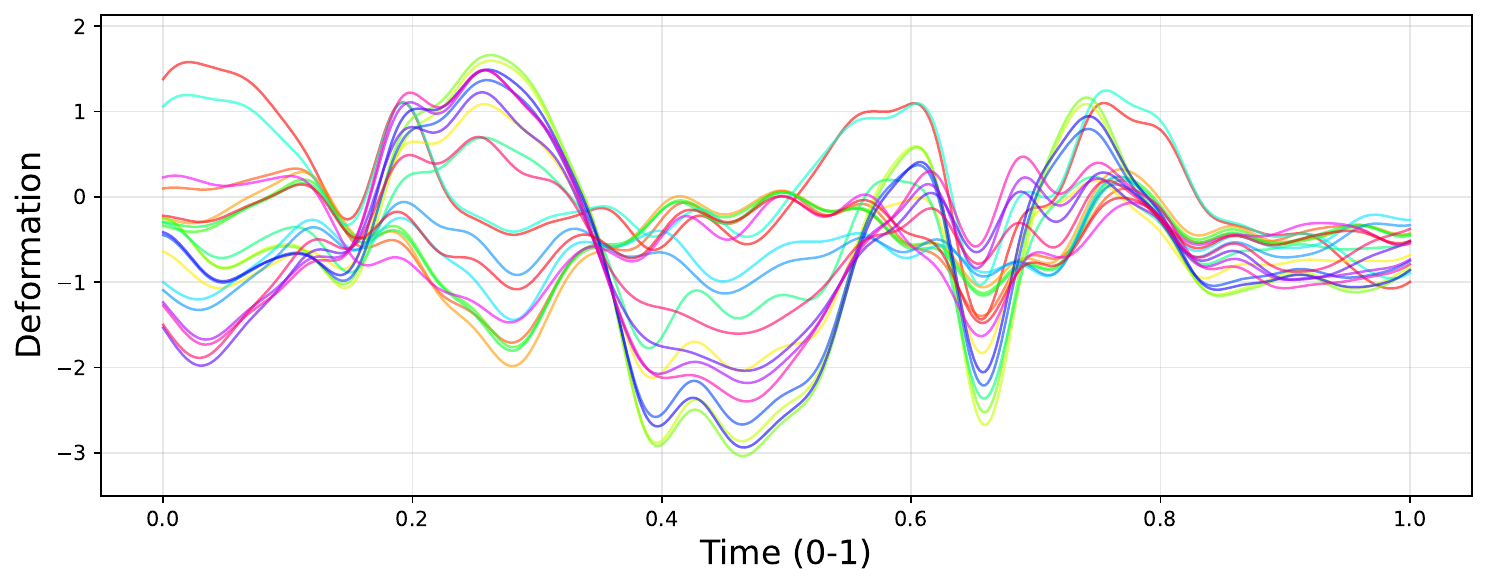}
\caption{X Translation}
\vspace{1mm}
\end{subfigure}
\begin{subfigure}[b]{0.49\linewidth}
\centering
\includegraphics[width=0.99\linewidth]{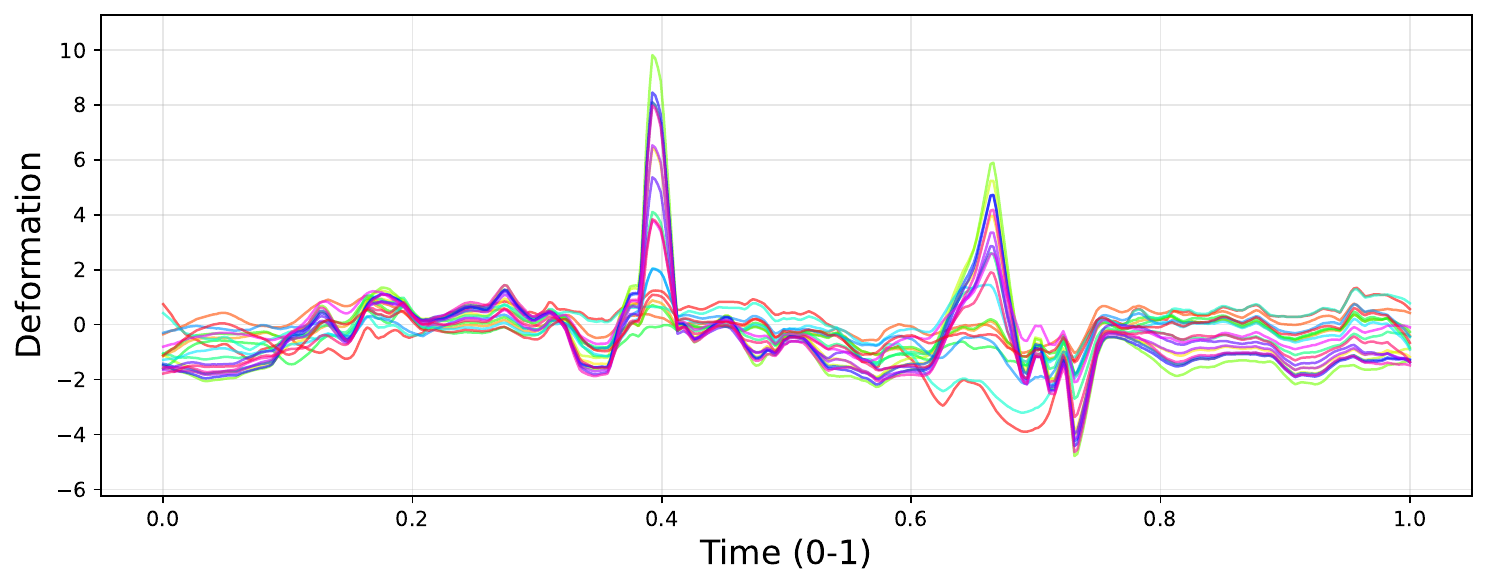}
\includegraphics[width=0.99\linewidth]{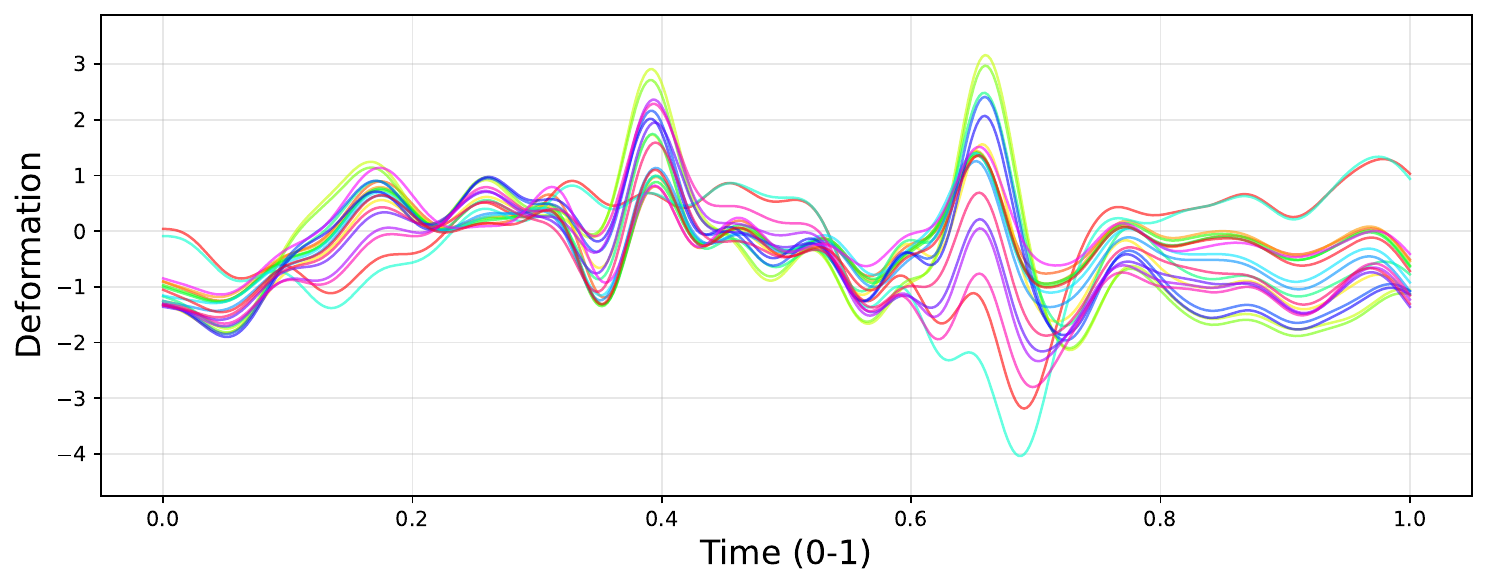}
\caption{Y Translation}
\vspace{1mm}
\end{subfigure}
\begin{subfigure}[b]{0.49\linewidth}
\centering
\includegraphics[width=0.99\linewidth]{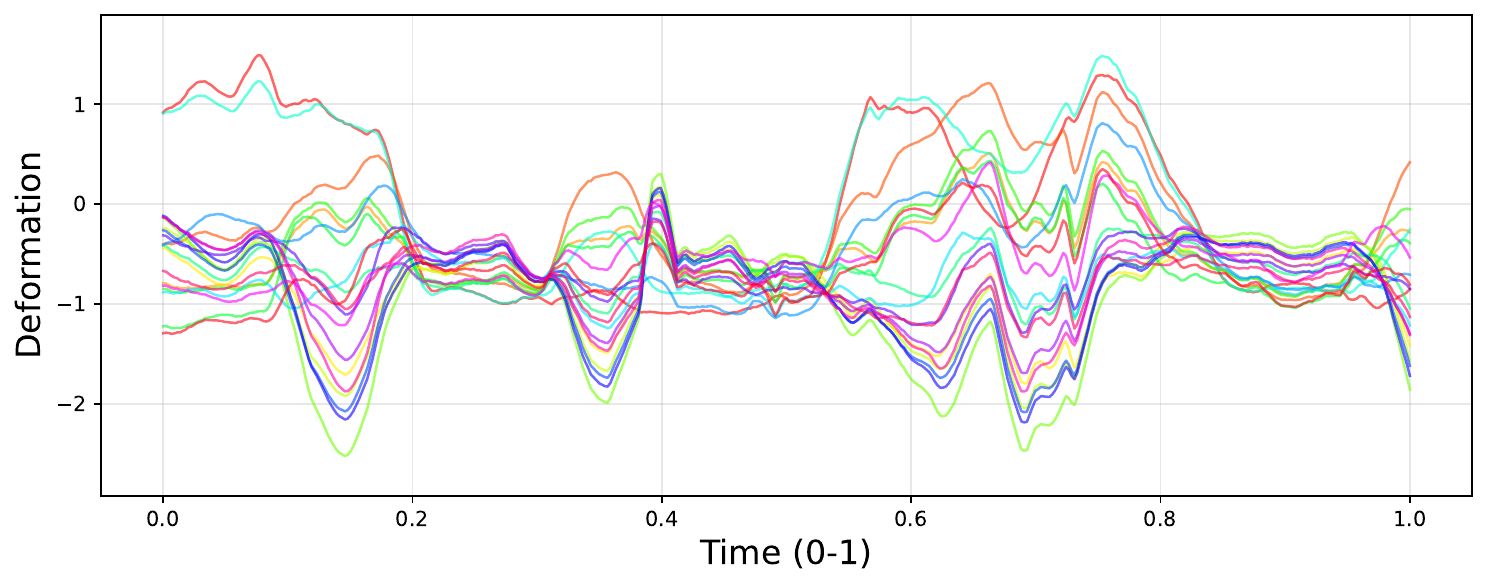}
\includegraphics[width=0.99\linewidth]{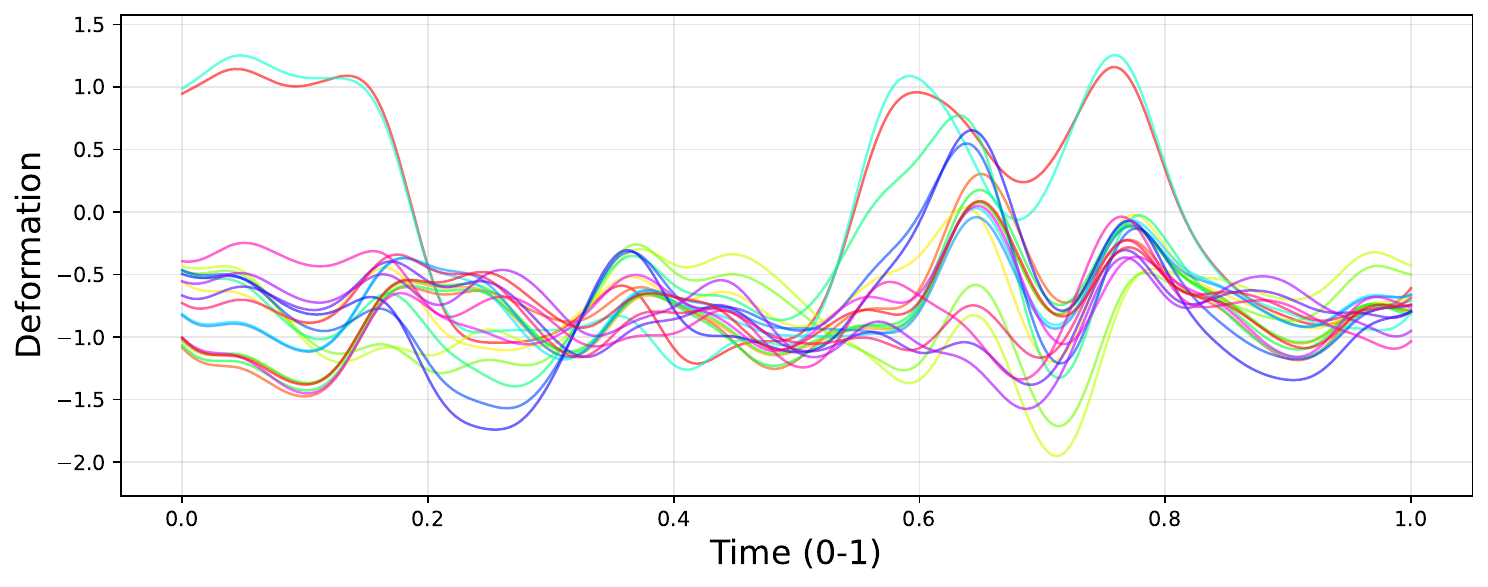}
\caption{Z Translation}
\vspace{1mm}
\end{subfigure}
\begin{subfigure}[b]{0.49\linewidth}
\centering
\includegraphics[width=0.99\linewidth]{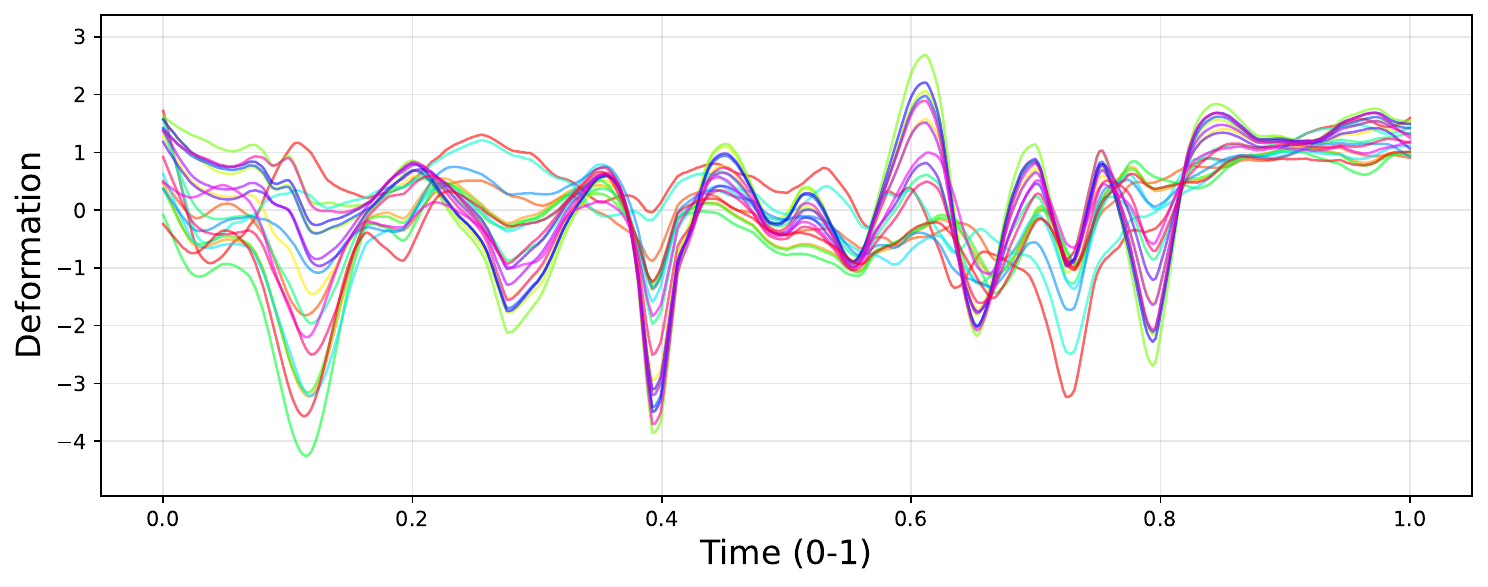}
\includegraphics[width=0.99\linewidth]{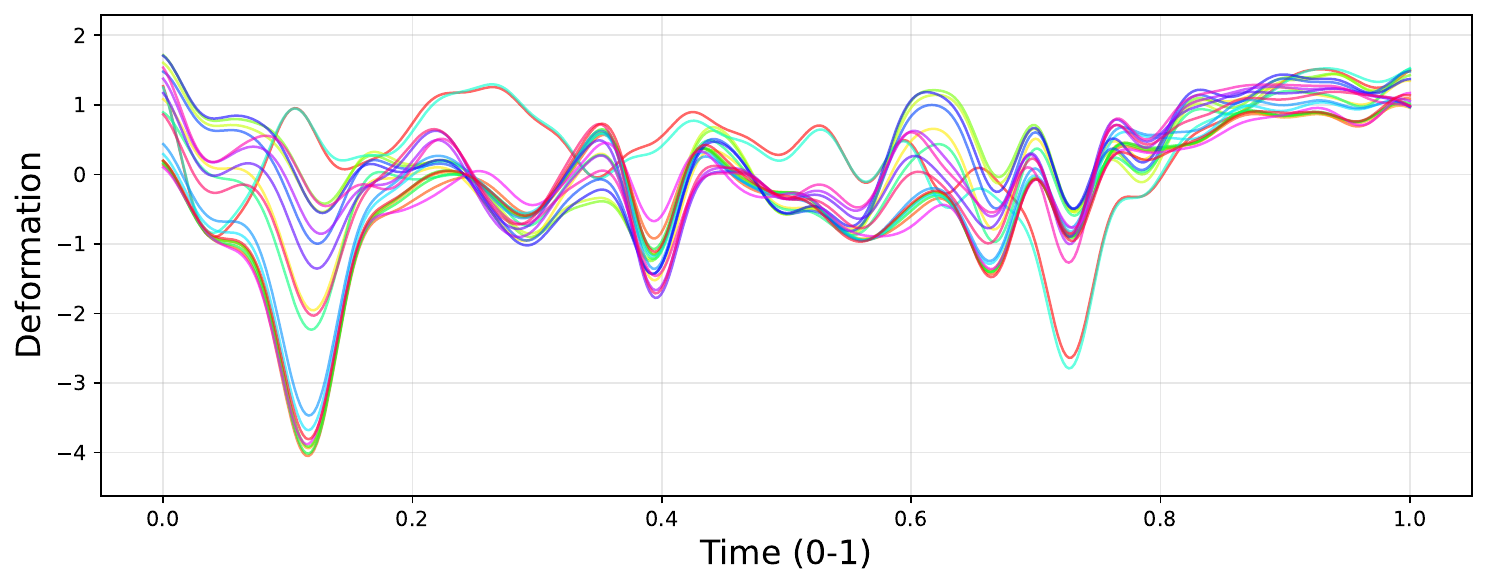}
\caption{6D Rotation Dim. 1}
\vspace{1mm}
\end{subfigure}
\begin{subfigure}[b]{0.49\linewidth}
\centering
\includegraphics[width=0.99\linewidth]{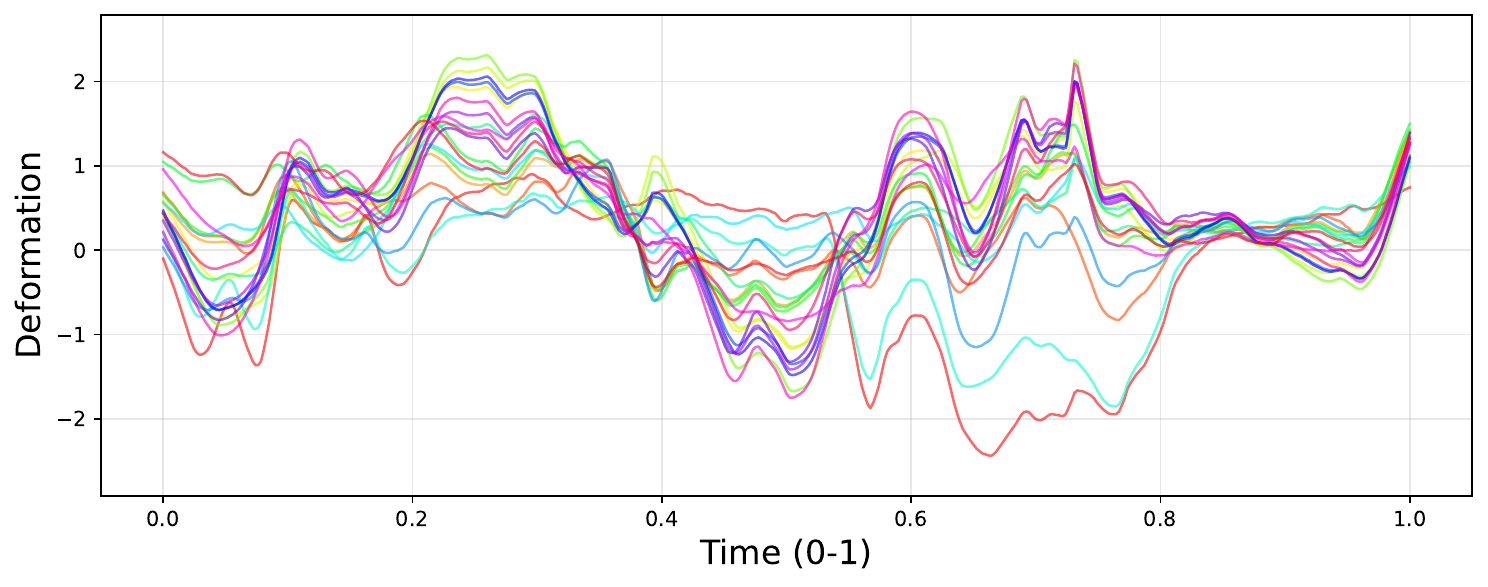}
\includegraphics[width=0.99\linewidth]{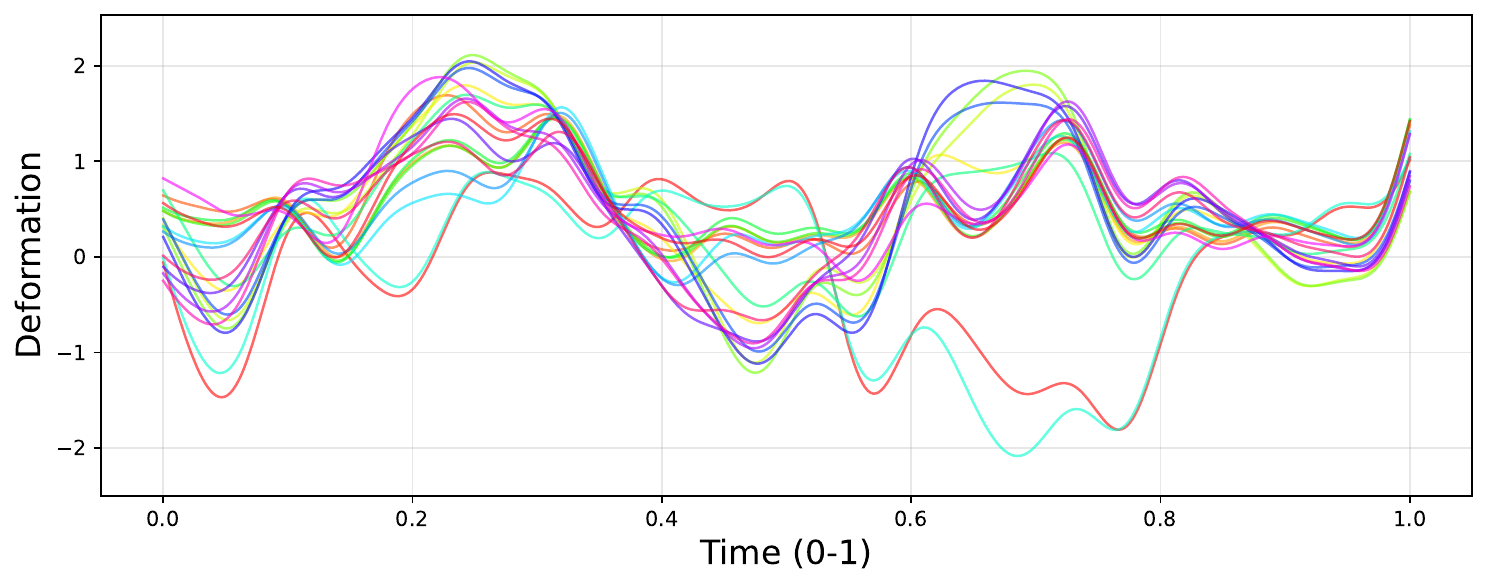}
\caption{6D Rotation Dim. 2}
\vspace{1mm}
\end{subfigure}
\begin{subfigure}[b]{0.49\linewidth}
\centering
\includegraphics[width=0.99\linewidth]{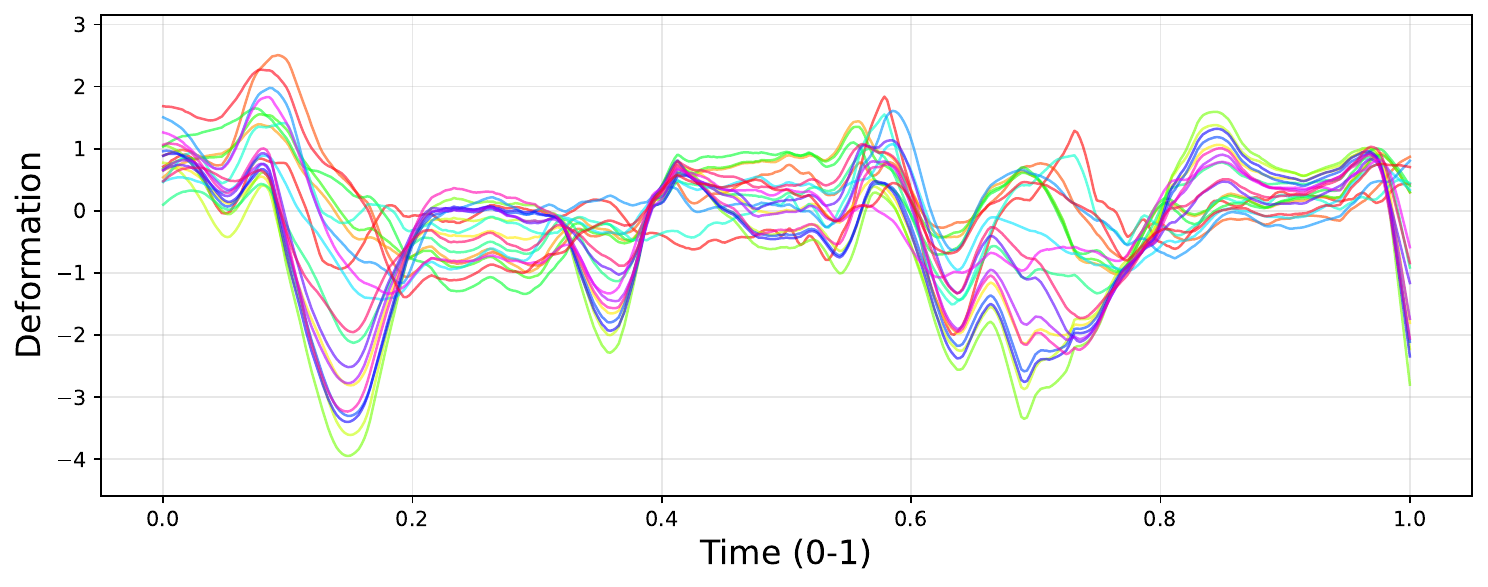}
\includegraphics[width=0.99\linewidth]{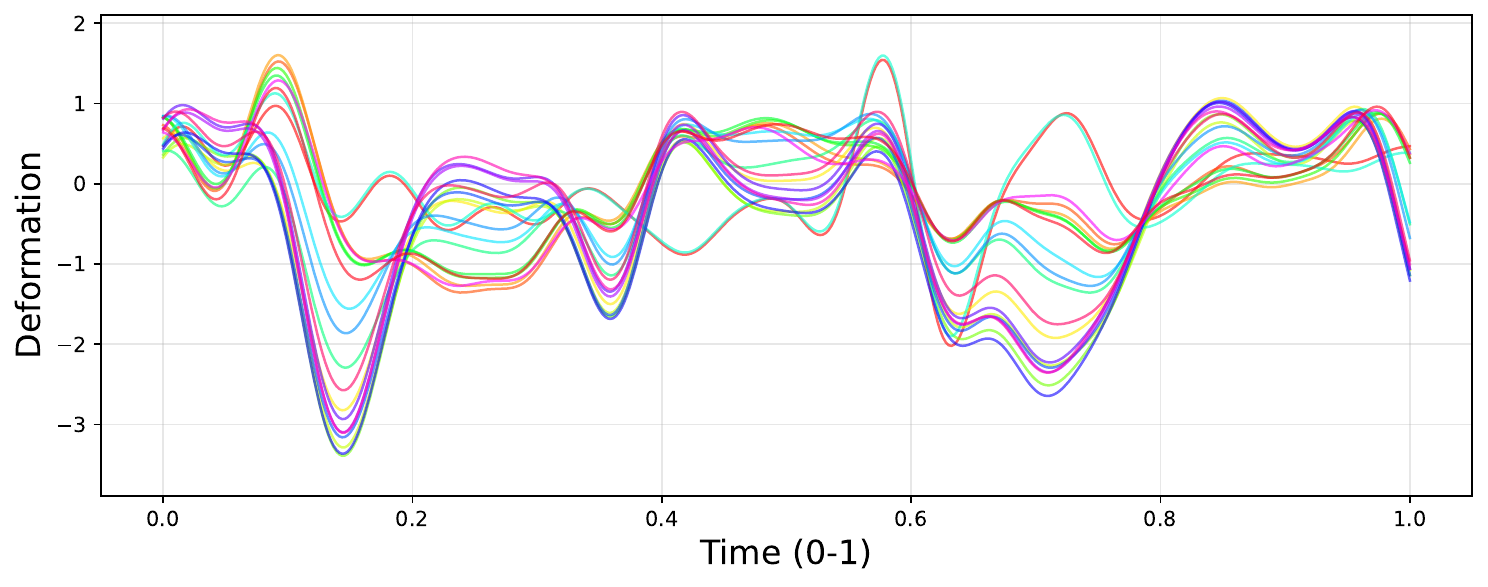}
\caption{6D Rotation Dim. 3}
\vspace{1mm}
\end{subfigure}
\begin{subfigure}[b]{0.49\linewidth}
\centering
\includegraphics[width=0.99\linewidth]{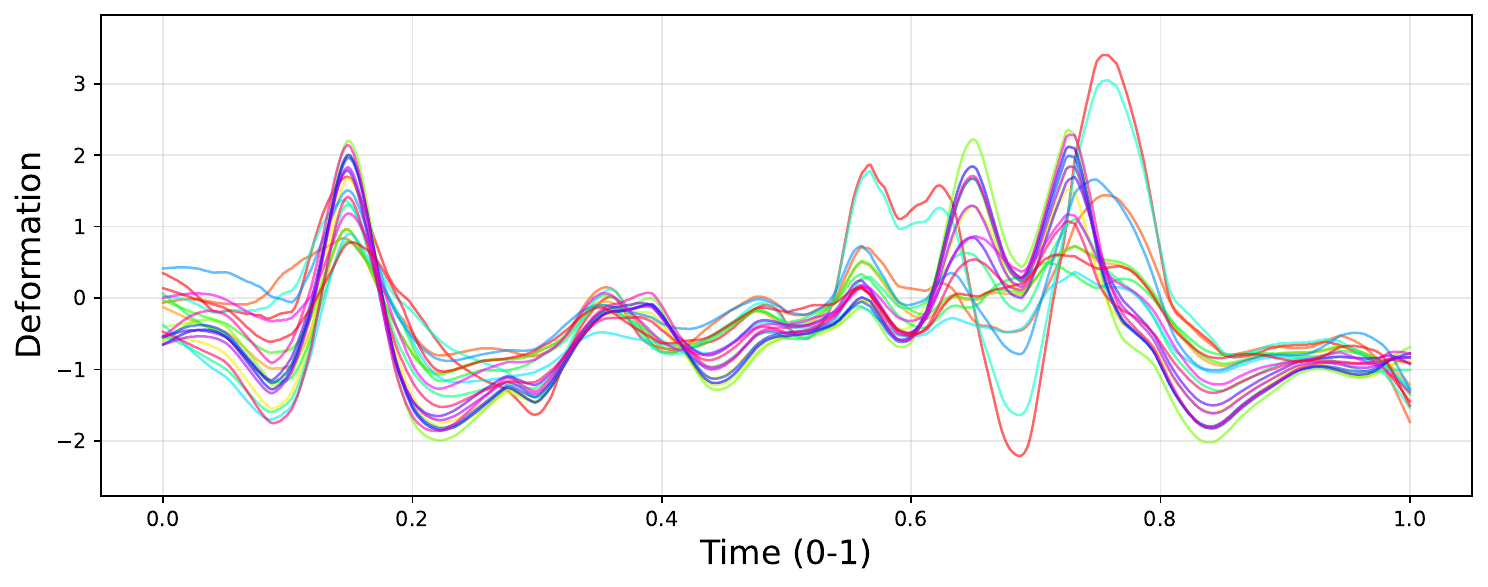}
\includegraphics[width=0.99\linewidth]{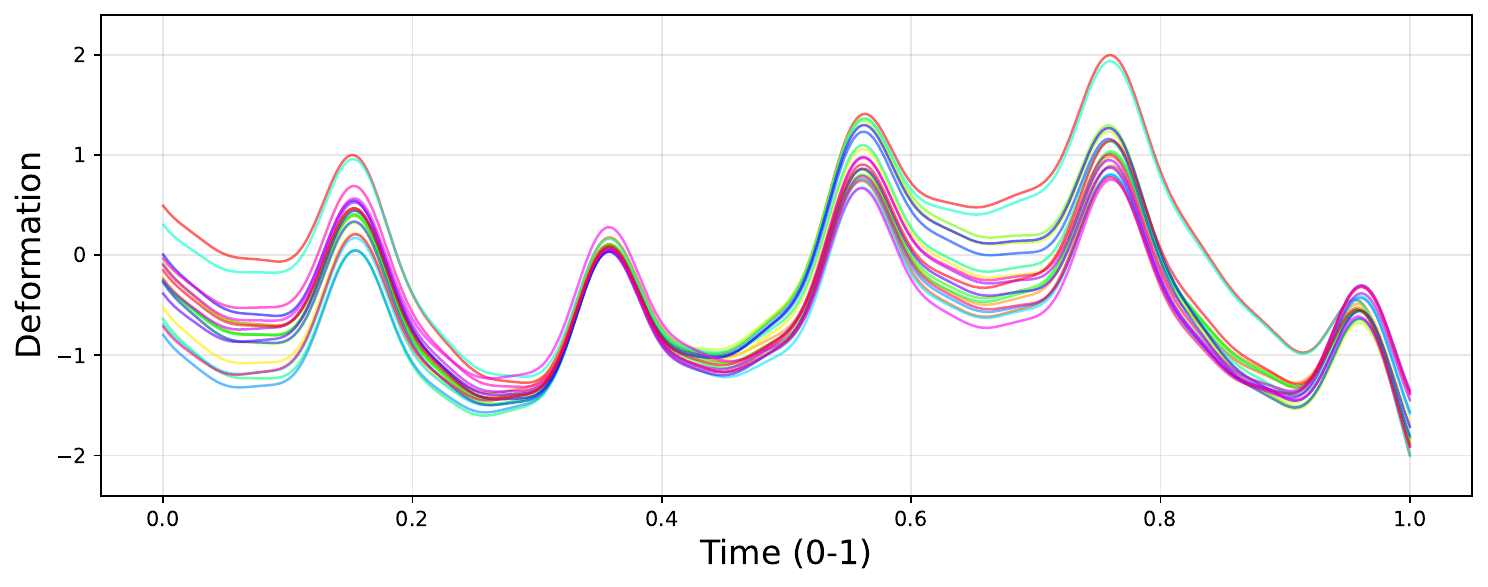}
\caption{6D Rotation Dim. 4}
\vspace{1mm}
\end{subfigure}
\begin{subfigure}[b]{0.49\linewidth}
\centering
\includegraphics[width=0.99\linewidth]{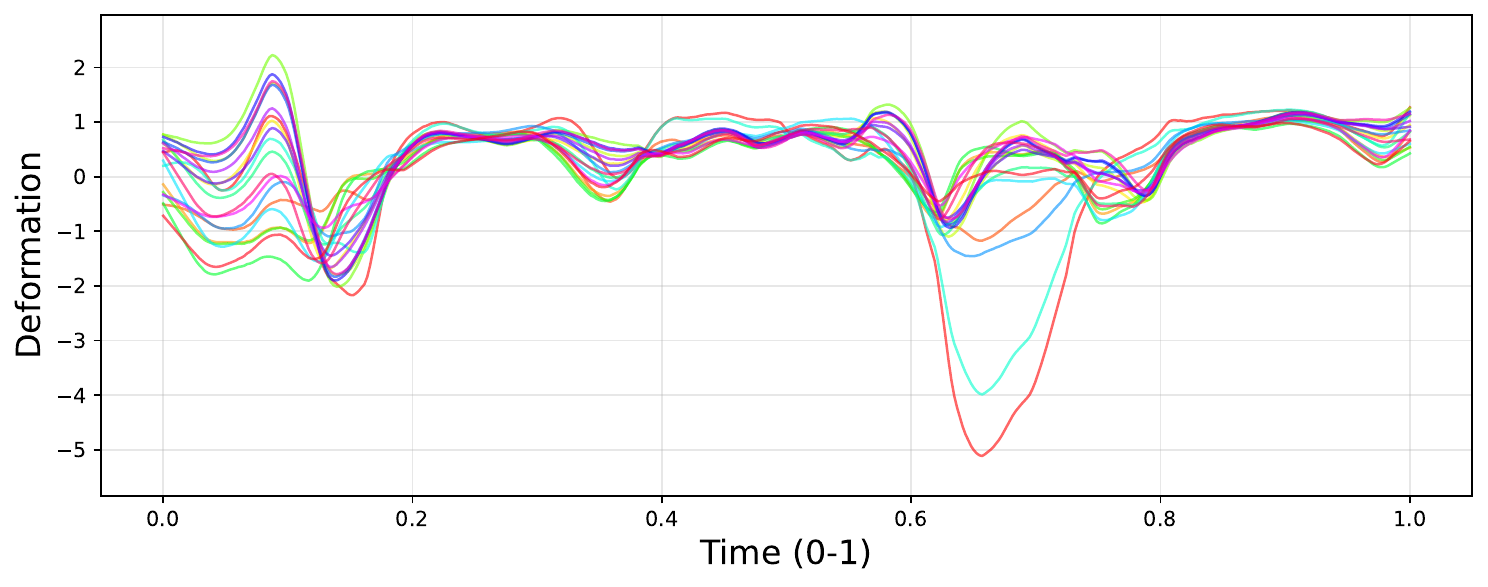}
\includegraphics[width=0.99\linewidth]{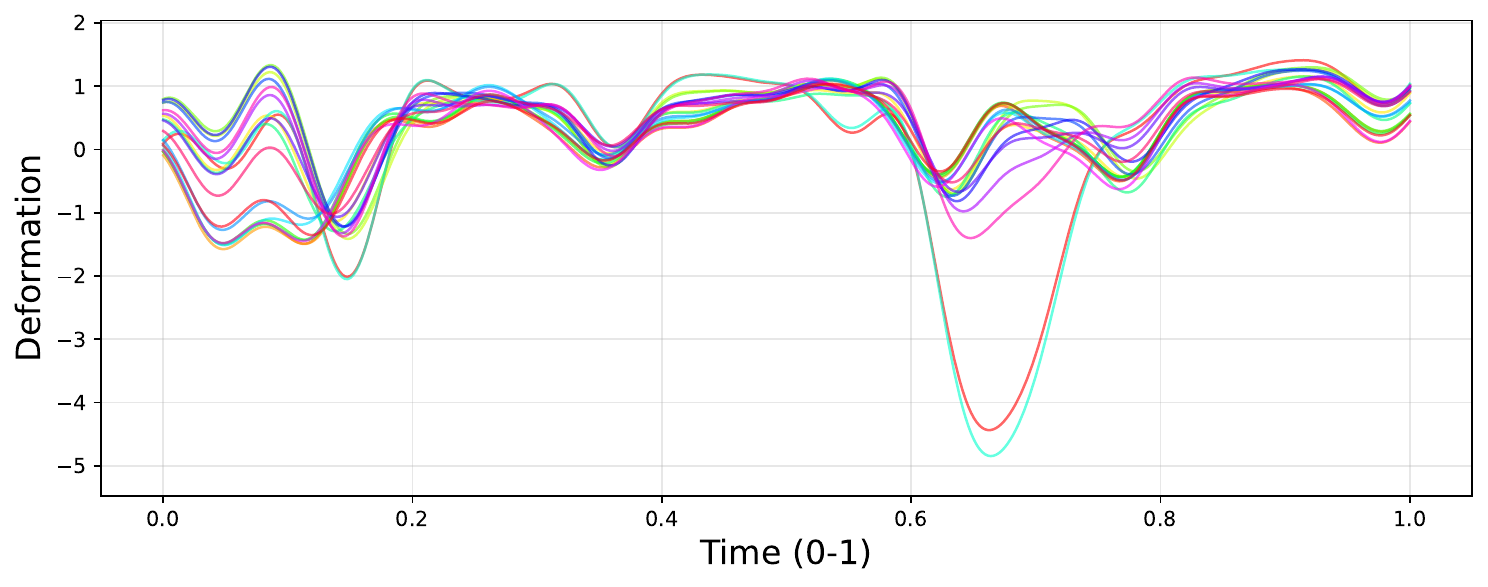}
\caption{6D Rotation Dim. 5}
\vspace{1mm}
\end{subfigure}
\begin{subfigure}[b]{0.49\linewidth}
\centering
\includegraphics[width=0.99\linewidth]{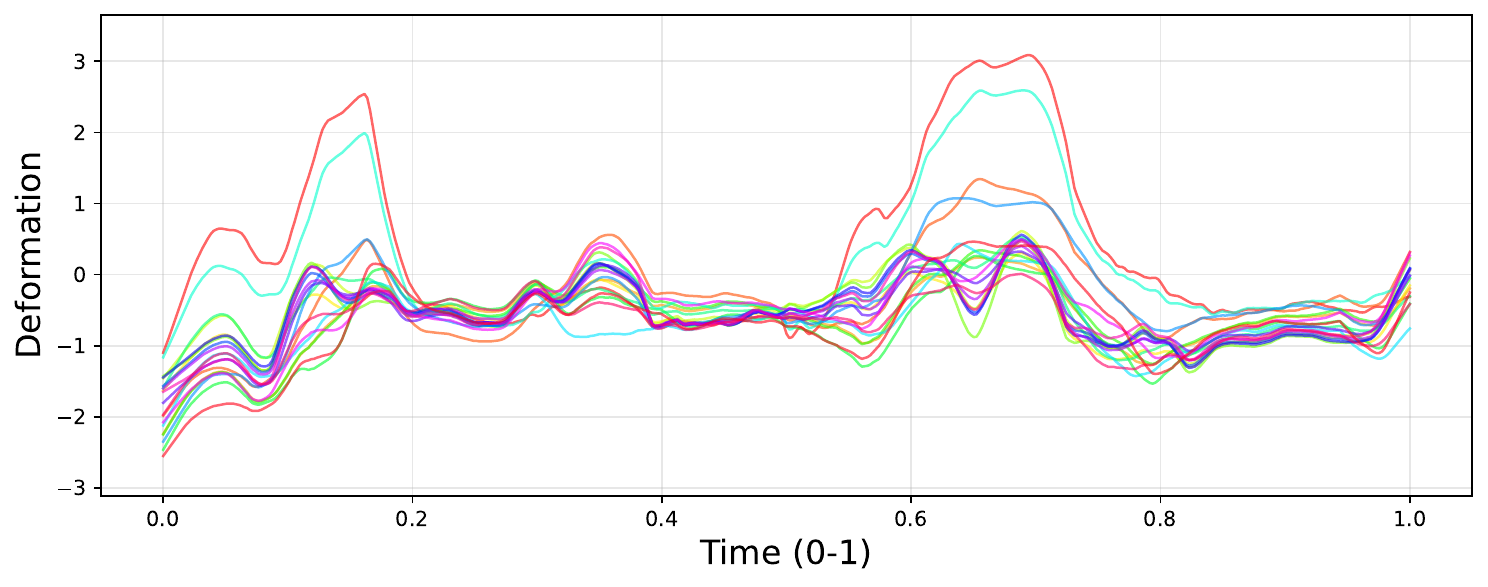}
\includegraphics[width=0.99\linewidth]{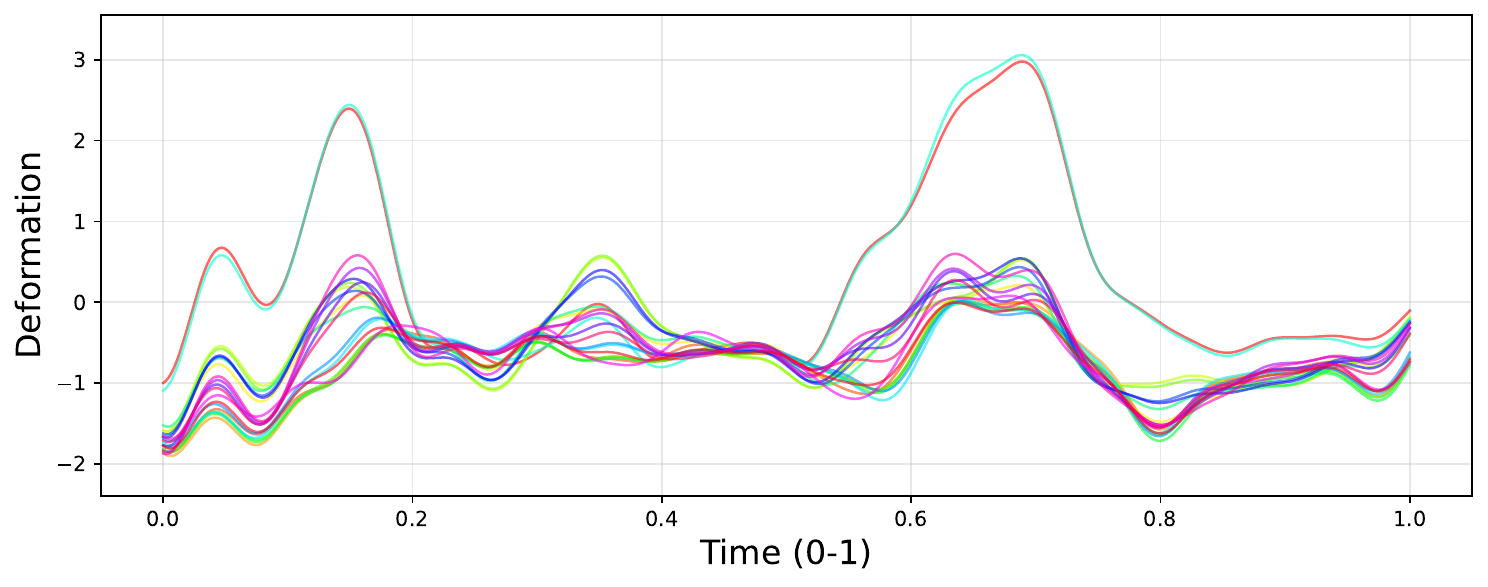}
\caption{6D Rotation Dim. 6}
\vspace{1mm}
\end{subfigure}
\vspace{-2mm}
\caption{Comprehensive trajectory comparison across all transformation dimensions in the \textit{Wheel} scene: \textbf{Original (top) vs. GP (bottom)}. 
Our variational Gaussian process approach successfully captures and denoises both translational and rotational dynamics, properly giving the priors on reconstructing trajectories, where even with non-periodic motion.}
\label{fig:trajectory_comparison_block}
\vspace{-2mm}
\end{figure}

\paragraph{Effect of GP-GS optimization}

Table~\ref{tab:ablation_gp_gs} shows the effect of the GP-GS optimization on the \textit{paper-windmill} scene.
Without joint optimization, the GP fails to learn accurate trajectory priors, resulting in worse performance than baseline methods. 
Our GP-GS optimization enables proper convergence and produces accurate priors that substantially improve reconstruction.

\begin{table}[t]
\caption{Ablation study on GP-GS optimization for the \textit{paper-windmill} scene.  
Our GP-GS optimization enables proper convergence and produces accurate priors that improve reconstruction.
}
\vspace{-2mm}
\centering
\begin{center}
\scalebox{0.80}{
\setlength\tabcolsep{9.5pt} 
\begin{tabular}{ l | c  c  cccccc }
\toprule
Method & mPSNR $\uparrow$ & mSSIM $\uparrow$ & mLPIPS $\downarrow$  
\\
\midrule 
Baseline &19.56 &0.558 &0.21\\
w/o GP-GS optimization &19.22 &0.541 &0.17 \\
w/ GP-GS optimization &19.88 &0.560 &0.19\\
\bottomrule
\end{tabular}
}
\end{center}
\label{tab:ablation_gp_gs}
\vspace{-2mm}
\end{table}

\paragraph{Additional analysis of kernel choice}

We choose Matérn kernel over RBF to better handle discontinuities (Fig.~\ref{fig:kernel_comparison}), which is important for modeling deformation of multi-objects.
As shown in Table~\ref{tab:kernel_ablation}, different temporal kernels achieve similar NVS performance, while periodic kernels perform better for long-horizon extrapolation.
Rather than advocating a single optimal kernel, we emphasize that kernel choice should reflect the desired modeling properties.
Spatial-temporal decomposition is essential; without it, if any single axis value among $(p_x, p_y, p_z, t)$  falls outside the observed range, prediction collapses to prior.

\begin{table}[t]
        	\centering
    	\caption{Ablation study on kernel designs and initialization. }
	\vspace{-2mm}
	\scalebox{0.8}{
   	\setlength\tabcolsep{3pt} 
    	\begin{tabular}{l|cc|cc}
    	\toprule
	\multirow{2}{*}{\textbf{Kernel design}} &   \multicolumn{2}{c}{\textbf{Extrapolation}}   &  \multirow{2}{*}{\textbf{m-PSNR $\uparrow$ }}  \\
	& 5 frames &  15 frames \\
	\midrule 
	w/o decomposition  &12.61	&10.6	&15.0 \\
	\hdashline 
	Temporal kernel $\rightarrow$ RBF & 13.8 	&11.1	&16.4 \\
	Temporal kernel  $\rightarrow$ Mat\'{e}rn & 13.1	&10.4	&17.1\\
	Temporal kernel $\rightarrow$ spectral-mixture&15.7 	&14.1	&\textbf{17.5}\\
	\hdashline
	Ours &\textbf{15.9}	&\textbf{14.2}	&17.3\\
	\bottomrule
    	\end{tabular}
    	}
\vspace{-2mm}
\label{tab:kernel_ablation}
\end{table}

\section{Additional Details}

We use an initial learning rate of $10^{-2}$ for both spatial and temporal inducing points in the GPs. 
The initial length scales are set to 0.001 and 0.002 for the spatial and temporal kernels, respectively. We employ an exponential decay schedule with a decay rate of 0.95 per epoch. 
The stochastic variational inference uses a batch size of 5000.
For GP-GS optimization, we run 1000 inner iterations for GP optimization. 
All GP inputs and outputs are normalized for more stable optimization.
The Gaussian process illustration in Figure~\textcolor{cvprblue}{1} of the main paper is adapted  from this cite\footnote{\url{https://www.lancaster.ac.uk/stor-i-student-sites/thomas-newman/2022/05/05/gaussian-processes-in-regression/}}.

\section{Breakdown Results}

To supplement Table~\cvprb{1} in the main paper, we provide detailed results; Table~\ref{tab:quantitative_dycheck_breakdown} shows the breakdown results on DyCheck.
Our method demonstrates substantial improvements on the \textit{wheel} and \textit{space-out} scenes, which feature high occlusion but simple geometry that can be effectively inferred by our GP algorithm.
The challenging subset uses subsampled frames with reduced overlap to evaluate robustness under sparse observations. 
The sampling for the challenging subset follows the diffusion-based method~\cite{huang2025vivid4d}, which is an orthogonal algorithm, whereas ours focuses on motion modeling, not visual priors.
Our method consistently outperforms baselines, with larger improvements under challenging conditions.

\begin{table}[t]
\caption{Per-scene breakdown results on DyCheck. 
Our method shows large improvements on highly occluded scenes (\textit{wheel}, \textit{space-out}) with simple geometry.}
\vspace{-2mm}
\centering
\begin{center}
\scalebox{0.80}{
\setlength\tabcolsep{4pt} 
\begin{tabular}{ c | l  | c cccccc }
\toprule
Scene & Method & mPSNR $\uparrow$ & mSSIM $\uparrow$ & mLPIPS $\downarrow$  
\\
\midrule 
\multirow{3}{*}{Apple} & Gaussian Marbles~\cite{stearns2024dynamic} &16.84 & 0.702 & 0.68\\ 
    & SoM~\cite{wang2025shape} &16.65 &0.752 &0.56 \\
    & Ours   &16.79 &0.754 &0.55 \\
    \hline
    \multirow{3}{*}{Block} & Gaussian Marbles~\cite{stearns2024dynamic} &16.50 &0.650 &0.51\\ 
    & SoM~\cite{wang2025shape}   &16.69 &0.660 &0.45\\
    & Ours  &17.01 &0.675 &0.44 \\
    \hline
    \multirow{3}{*}{{Paper}} & Gaussian Marbles~\cite{stearns2024dynamic}  &15.96 &0.296 &0.58\\ 
   & SoM~\cite{wang2025shape} &19.56 &0.558 &0.21 \\
    & Ours  &19.88 &0.560 &0.19 \\
    \hline
    \multirow{3}{*}{Spin} &  Gaussian Marbles~\cite{stearns2024dynamic} &17.84 &0.515 &0.48 \\ 
    & SoM~\cite{wang2025shape} &17.31 &0.707 &0.31 \\
    & Ours  &17.56 &0.728 &0.27  \\
    \hline
    \multirow{3}{*}{Teddy} &  Gaussian Marbles~\cite{stearns2024dynamic} &13.01 &0.550 &0.66  \\ 
    & SoM~\cite{wang2025shape} &13.44 &0.546 &0.61 \\
& Ours  &13.55 &0.576 &0.59 \\
\hline
    \multirow{3}{*}{Space-out} &  Gaussian Marbles~\cite{stearns2024dynamic} &15.19 &0.560 &0.54  \\ 
  & SoM~\cite{wang2025shape}     &19.41 &0.617 &0.37 \\
& Ours   &19.83 &0.628 &0.38 \\
\hline
    \multirow{3}{*}{Wheel} &  Gaussian Marbles~\cite{stearns2024dynamic}  &15.55 &0.531 &0.53  \\ 
   & SoM~\cite{wang2025shape}   &16.41 &0.620 &0.37\\
& Ours &17.01 &0.620 &0.35 \\
\bottomrule
\end{tabular}
}
\end{center}
\vspace{-2mm}
\label{tab:quantitative_dycheck_breakdown}
\end{table}

\end{document}